\documentclass[10pt,journal,compsoc]{IEEEtran}
\ifCLASSOPTIONcompsoc
  \usepackage[nocompress]{cite}
\else
  \usepackage{cite}
\fi

\ifCLASSINFOpdf
\else
\fi

\usepackage{algorithm}
\usepackage{times}
\usepackage{url}
\usepackage{epsfig}
\usepackage{graphicx}
\usepackage{subfigure}
\usepackage{amsmath}
\usepackage{amssymb}
\usepackage{mathrsfs}
\usepackage{array}
\usepackage{bm}
\usepackage{algpseudocode}
\usepackage[pagebackref=true,breaklinks=true,letterpaper=true,colorlinks,bookmarks=false]{hyperref}
\usepackage[misc]{ifsym}

\usepackage{tikz}
\usetikzlibrary{matrix}

\def\E{{\rm E}}
\def\N{{\rm N}}
\def\KL{{\rm KL}}

\def\M{{\cal M}}
\def\P{P_{\rm data}}

\def\G{{\cal G}}

\def\tY{\tilde{Y}}
\def\tn{\tilde{n}}

\def\hY{\hat{Y}}
\def\hX{\hat{X}}

\begin{document}
\title{Cooperative Training of \\Descriptor and Generator Networks}

\author{Jianwen Xie,
        Yang Lu,
        Ruiqi Gao,
        Song-Chun Zhu,~\IEEEmembership{Fellow, IEEE,}
        and~Ying Nian Wu% <-this % stops a space
\IEEEcompsocitemizethanks{
\IEEEcompsocthanksitem J. Xie is with Hikvision Research Institute, Santa Clara, USA. Y. Lu is with Facebook, Menlo Park, California. R. Gao, S.-C. Zhu, and Y. N. Wu are with the Department of Statistics, University of California, Los Angeles. 
 \protect
}% <-this % stops an unwanted space
%\thanks{Manuscript received XX, 2017; revised XX, 201X.}
}

\markboth{ }%
{Xie \MakeLowercase{\textit{et al.}}: Bare Demo of IEEEtran.cls for Computer Society Journals}

\IEEEtitleabstractindextext{%
\begin{abstract}
This paper studies the cooperative training of two generative models for image modeling and synthesis. Both models are parametrized by convolutional neural networks (ConvNets). The first model is a deep energy-based model, whose energy function is defined by a bottom-up ConvNet, which maps the observed image to the energy. We call it the descriptor network. The second model is a generator network, which is a non-linear version of factor analysis. It is defined by a top-down ConvNet, which maps the latent factors to the observed image. The maximum likelihood learning algorithms of both models  involve MCMC sampling such as Langevin dynamics.  
%In the training of the descriptor net, the Langevin sampling is used to sample synthesized examples from the model. In the training of the generator net, the Langevin sampling is used to sample the latent factors from the posterior distribution. The Langevin sampling in both algorithms can be time consuming.  
We observe that the two learning algorithms can be  seamlessly interwoven into a cooperative learning algorithm that can train both models simultaneously. Specifically, within each iteration of the cooperative learning algorithm, the generator model generates initial synthesized examples to initialize a finite-step MCMC that samples and trains the energy-based descriptor model.  After that, the generator model learns from how the MCMC changes its synthesized examples. That is, the descriptor model teaches the generator model by MCMC, so that the generator model accumulates the MCMC transitions and reproduces them by direct ancestral sampling.  We call this scheme MCMC teaching. We show that the cooperative algorithm can learn highly realistic generative models. 
%From the Markov chain Monte Carlo (MCMC) perspective, the generator accumulates all the past Markov transitions for sampling the descriptor, and reproduces the current marginal distribution of MCMC in one shot. 
\end{abstract}

}

\maketitle

\IEEEdisplaynontitleabstractindextext

% \fi
\IEEEpeerreviewmaketitle

\IEEEraisesectionheading{\section{Introduction}\label{sec:introduction}}

\IEEEPARstart{L}{earning} generative models of images is a fundamental problem in computer vision and machine learning.  In this article, we propose a cooperative learning algorithm to train two important classes of generative models jointly for image modeling, representation and synthesis.

\subsection{Two generative models}

We begin with an analogy. A student writes up an initial draft of a paper. Her advisor then revises it. After that they submit the revised paper for review. The student then learns from her advisor's revision, while the advisor learns from the outside review. In this analogy, the advisor guides the student, but the student does most of the work. 

This paper is about two generative  models, and they play the roles of teacher and student as mentioned above. Both models are parametrized by  convolutional neural networks (ConvNets or CNNs) \cite{lecun1998gradient, krizhevsky2012imagenet}.  They are of opposite directions. One is bottom-up, and the other is top-down, as illustrated by the following diagram: 
\begin{eqnarray}
\begin{array}[c]{ccc}
\mbox{{ Bottom-up ConvNet}} && \mbox{{ Top-down ConvNet}}\\
\mbox{{\bf energy}} & & \mbox{{\bf latent variables}}\\
 \Uparrow&&\Downarrow\\
{\rm image} & & {\rm image}\\
\mbox{(a) Descriptor Network}  &&\mbox{ (b) Generator Network}\\
\mbox{(teacher)}  &&\mbox{(student)}
\end{array}  \label{eq:diagram0}
\end{eqnarray}

These two nets correspond to two major classes of probabilistic  models. (a) The  energy-based models \cite{Lecun2006, Hinton2002a} or the Markov random field models \cite{zhu1997minimax, roth2005fields}, where the probability distribution is defined by the feature statistics or the energy function computed  from the image by a bottom-up process.   (b) The latent variable models or the directed graphical models, where the image is assumed to be a transformation of the latent factors that follow a known prior distribution. The latent factors generate the image by a top-down process via direct ancestral sampling. A classical example is factor analysis \cite{rubin1982algorithms}. 

The two classes of models have been contrasted by \cite{zhu2003statistical, guo2003modeling, teh2003energy, wu2004information, Ng2011}. Both classes of models can benefit from the high capacity of the ConvNets. (a) In the energy-based model,  the energy function can be defined by a bottom-up ConvNet that maps the image to the energy  \cite{Ng2011, Dai2015ICLR, LuZhuWu2016, XieLuICML, Tu2017}, and the energy function is usually the sum or a linear combination of the features at the top layer. For ease of reference, we call the resulting model a descriptor network  following \cite{zhu2003statistical}, because it is built on descriptive feature statistics. (b) In the latent variable model or the directed graphical model, the transformation from the latent factors to the image can be defined by a top-down ConvNet \cite{zeiler2013visualizing, Alexey2015}, which maps the latent factors to the image. We call the resulting model a generator network  following \cite{goodfellow2014generative}.

The likelihoods of both models involve intractable integrals, and the gradients of both log-likelihoods involve intractable expectations that can be approximated by Markov chain Monte Carlo (MCMC).  We notice that the maximum likelihood algorithms for learning  the two models can be interwoven into a cooperative learning algorithm, where each iteration consists of the following two steps: (1) Modified contrastive divergence for energy-based descriptor model: The learning of the  descriptor model is based on the contrastive divergence  \cite{Hinton2002a}, but the finite-step MCMC sampling of the model is initialized from the synthesized examples generated by the generator model instead of being initialized from the observed examples.  (2) MCMC teaching of the latent variable generator model: The learning of the generator model is based on how the MCMC in (1) changes the initial synthesized examples generated by the generator model.  That is, the descriptor model (teacher) distills its knowledge to the generator model (student) via MCMC, and we call it MCMC teaching. Our experiments show that the cooperative learning algorithm can learn realistic generative models of images. 

\subsection{Motivations and contributions}  The main motivation for our work is that we find it very challenging to learn the two models separately, when the training images are highly varied. We find it much easier for the cooperative algorithm to learn highly realistic models from such data. Another motivation is to develop an alternative system to the generative adversarial networks (GAN) \cite{goodfellow2014generative, denton2015deep, radford2015unsupervised}, where in our system both models are learned generatively.  Our experiments suggest that the cooperative learning is stable and does not encounter mode collapsing issue. 

%The  advantages of the cooperative learning versus separate learning are as follows. (1) The generator (latent variable model) jump-starts the MCMC of the descriptor (energy-based model) by supplying fresh and independent examples via ancestral sampling in each iteration. (2) The generator learns from how the MCMC changes the synthesized examples it generates, where the values of the latent variables are known,  thus they do not need to be inferred, and the learning is much easier than learning from the observed examples.  In MCMC teaching in (2), the generator model seeks to approximate the descriptor model, so that the modified contrastive divergence in (1) is close to maximum likelihood. With repeated teaching by finite-step MCMC in (2), the generator accumulates the MCMC transitions and reproduces them by direct ancestral sampling. 

The contributions of our work are as follows. We propose a cooperative learning algorithm to train the energy-based descriptor model and the latent variable generator model  simultaneously, so that the learned models can synthesize highly realistic images. Our work connects the undirected model (descriptor) and the directed model (generator). It also connects the ancestral sampling (generator) and the MCMC sampling (descriptor).  

In the following subsections, we shall further explain the basic idea of our paper, and review related work. 

\subsection{Two maximum likelihood algorithms} 

Both the energy-based model (the descriptor network) and the latent variable model (the generator network) can be learned from the training examples by maximum likelihood. 

\begin{figure}
\begin{center}
\includegraphics[height=.4\linewidth]{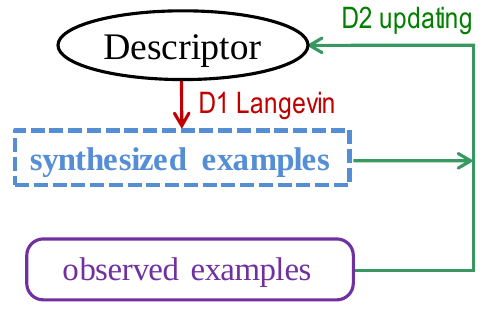}
\caption{The flow chart of Algorithm D for training the descriptor network. The updating in Step D2 is based on the difference between the observed examples and the synthesized examples. The Langevin sampling of the synthesized examples from the current model in Step D1 can be time consuming. }
\label{fig:diagramD}
\end{center}
\end{figure}

The  training algorithm for the descriptor network alternates between the following two steps \cite{XieLuICML}.  We call it Algorithm D. See Figure \ref{fig:diagramD} for an illustration. 

{\bf Step D1 for Langevin revision}: Sampling synthesized examples from the current model by Langevin dynamics \cite{neal2011mcmc, girolami2011riemann, zhu1998grade}.  We call this step  Langevin revision because it keeps revising the current synthesized examples. 

{\bf Step D2 for density shifting}:  Updating the parameters of the descriptor network based on the difference between the observed examples and the synthesized examples obtained in D1 (Langevin revision). This step is to shift the high probability regions of the descriptor from the synthesized examples to the observed examples. 

Both steps can be powered by back-propagation. The algorithm is thus an alternating back-propagation algorithm. 

Intuitively, in Step D1 (Langevin revision), the descriptor network is dreaming by synthesizing examples from the current model. In Step D2 (density shifting), the descriptor updates its parameters to make the dream more realistic.  

\begin{figure}
\begin{center}
\includegraphics[height=.4\linewidth]{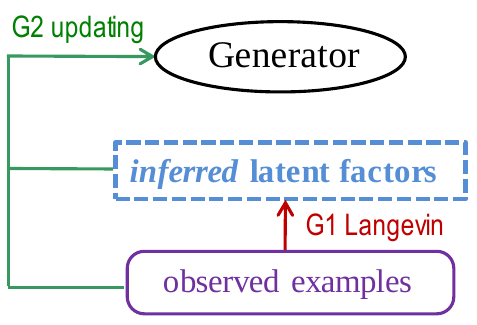}
\caption{The flow chart of Algorithm G for training the generator network. The updating in Step G2 is based on the observed examples and their inferred latent factors. The Langevin sampling of the latent factors from the current posterior distribution in Step G1 can be time consuming. }
\label{fig:diagramG}
\end{center}
\end{figure}

The training algorithm for the generator network alternates between the following two steps \cite{HanLu2016}. We call it Algorithm G. See Figure \ref{fig:diagramG} for an illustration. 

{\bf Step G1 for Langevin inference}: For each observed example, sample the latent factors from the current posterior distribution  by Langevin dynamics. We call this step Langevin inference because it infers the latent factors for each observed example, in order for the inferred latent factors to explain or reconstruct the observed examples. 

{\bf Step G2 for reconstruction}:  Updating the parameters of the generator network based on the observed examples and their inferred latent factors obtained in G1 (Langevin inference), so that the inferred latent factors can better reconstruct the observed examples. 

Again, both steps can be powered by back-propagation, and the algorithm is thus an alternating back-propagation algorithm. 

The training algorithm for generator network is similar to the EM algorithm \cite{dempster1977maximum}, where step G1 (Langevin inference) can be mapped to the E-step, and step G2 (reconstruction) can be mapped to the M-step. It is an unsupervised learning algorithm because the latent factors are unobserved. 

Intuitively, in Step G1 (Langevin inference), the generator network is thinking about each observed example by inferring the latent factors that can reconstruct it. The thinking involves explaining-away reasoning: the latent factors compete with each other in the Langevin inference process to explain the example. In Step G2 (reconstruction), the generator network updates its parameters to make the thinking more accurate. 

Compared to Step D2 (density shifting), Step G2 (reconstruction) is also a form of density shifting from the current reconstructed examples towards the observed examples, but it requires inferring the latent factors for each observed example. 

\subsection{Cooperative training via MCMC teaching} 

The two algorithms can operate separately on their own \cite{XieLuICML, HanLu2016}.  But just like the case with the student and the advisor, they benefit from cooperating with each other, where the generator network plays the role of the student, and the descriptor network plays the role of the teacher. 

The needs for cooperation stem from the fact that the Langevin sampling in Step D1 (Langevin revision) and Step G1 (Langevin inference) can be time consuming. If the two algorithms cooperate, they can jumpstart each other's Langevin sampling in D1 (Langevin revision) and G1 (Langevin inference). The resulting algorithm seamlessly interweaves the steps in the two algorithms with minimal modifications. 

Intuitively, while the descriptor needs to dream hard in Step D1 (Langevin revision) for synthesis, the generator needs to think hard in Step G1 (Langevin inference) for explaining-away reasoning. On the other hand, the generator is actually a much better dreamer because it can generate images by direct ancestral sampling without MCMC, while the descriptor does not need to think in order to learn. 

Specifically, we have the following steps in the cooperative learning algorithm: 

{\bf Step G0 for initial generation}: Generate the initial synthesized examples using the generator network. These initial examples can be obtained by direct ancestral sampling via a top-down process. 

{\bf Step D1 for Langevin revision}:  Starting from the initial synthesized examples produced in Step G0 (initial generation), run Langevin revision dynamics for a finite number of steps to obtain  the revised synthesized examples.  

{\bf Step D2 for density shifting}: The same as before, except that we use the revised synthesized examples produced by Step D1 (Langevin revision) to shift the density of the descriptor towards the observed examples. 

{\bf Step G1 for Langevin inference}: The generator network can learn from the revised synthesized examples produced by Step D1 (Langevin revision). For each revised synthesized example, we know the values of the latent factors that generate the corresponding initial synthesized example in Step G0 (initial generation), therefore we may simply infer the latent factors to be their known values given in Step G0 (initial generation), or initialize  the Langevin inference dynamics in Step G1 (Langevin inference) from the known values.  

{\bf Step G2 for reconstruction}: The same as before, except that we use the revised synthesized examples and the inferred latent factors obtained in Step G1 (Langevin inference) to update the generator.  The generator in Step G0 (initial generation) generates and thus reconstructs the initial synthesized examples. Step G2 (reconstruction) updates the generator to reconstruct the revised synthesized examples. The revision of the generator accounts for the revisions made by the Langevin revision dynamics in Step D1 (Langevin revision).  

\begin{figure}
\begin{center}
\includegraphics[width=.7\linewidth]{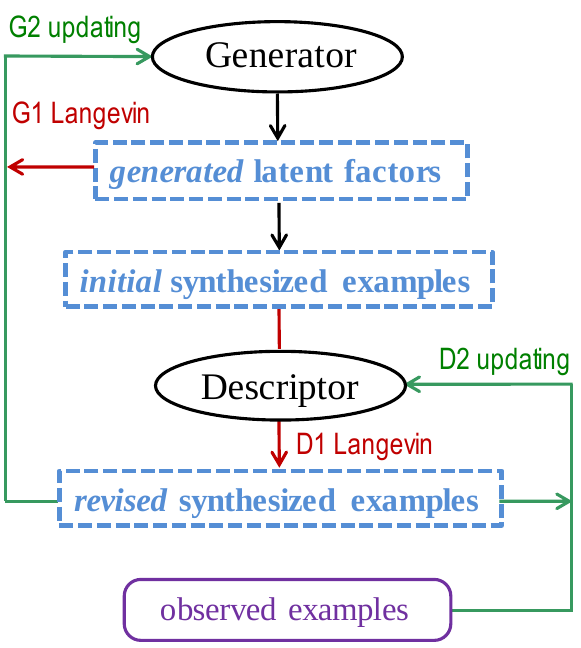}
\caption{The flow chart of the cooperative algorithm. The part of the flow chart for training the descriptor is similar to Algorithm D in Figure \ref{fig:diagramD}, except that the D1 Langevin sampling is initialized from the initial synthesized examples supplied by the generator. The part of the flow chart for training the generator can also be mapped to Algorithm G in Figure \ref{fig:diagramG}, except that the revised synthesized examples play the role of the observed examples, and the {known} generated latent factors can be used as inferred latent factors (or be used to initialize the G1 Langevin sampling of the latent factors). }
\label{fig:diagram}
\end{center}
\end{figure}

Figure \ref{fig:diagram}  shows  the flow chart of the cooperative algorithm. The generator is like the student. It generates the initial draft of the synthesized examples. The descriptor is like the teacher. It revises the initial draft  by running a number of Langevin revisions. The descriptor learns from the outside review, which is in the form of the difference between the observed examples and the revised synthesized examples. This is a modified version of contrastive divergence, where the MCMC is initialized by the generator instead of being initialized from the observed examples. The generator learns from how the descriptor's MCMC revises the initial draft  by reconstructing the revised draft. This is MCMC teaching. 

The reason we let the generator network learn from the revised synthesized examples instead of the observed examples is that the generator does not know the latent factors that generate the observed examples, and it has to think hard to infer them by explaining-away reasoning. However, the generator knows the values of the latent factors that generate each initial synthesized example, and thus it essentially knows the latent factors when learning from the revised synthesized examples. By reconstructing the revised synthesized examples, the generator traces and accumulates the Langevin revisions made by  the descriptor.   This cooperation is thus beneficial to the generator by relieving it the burden of inferring the latent factors.  

While the generator may find it hard to learn from the observed examples directly, the descriptor has no problem learning from the observed examples  because it only needs to compute the bottom-up features deterministically. However, it needs synthesized examples to find its way to shift its density, and they do not come by easily. The generator can provide unlimited number of examples, and in each learning iteration, the generator supplies a completely {new batch} of independent examples on demand.  The descriptor only needs to revise the new batch of examples instead of generating a new batch by itself from scratch. The generator has  memorized the cumulative effect of {all} the past Langevin revisions, so that it can produce new samples in one shot.  This cooperation is thus  beneficial to the descriptor by relieving it the burden of synthesizing examples from scratch. 

In terms of density shifting, the descriptor shifts its density from the revised synthesized examples towards the observed examples in Step D2, and the generator shifts its density from the initial synthesized examples towards the revised synthesized examples in Step G2 by reconstructing the latter. 

In terms of energy function,   the descriptor shifts its low energy regions towards the observed examples, and it induces the generator to map the latent factors to its low energy regions. It achieves that by stochastically relaxing the synthesized examples towards  low energy regions, and let the generator track the synthesized examples. 

%True to the essence of learning, the two nets still learn from the data. The descriptor learns from the observed data and teaches the generator through the synthesized data. Specifically, they collaborate and communicate with each other through synthesized data in the following three versions (S stands for synthesis). (S1) The generator generates the initial draft. (S2) The descriptor revises the draft. (S3) The generator reconstructs the revised draft. 

\subsection{Related work}

Our work is related to contrastive divergence \cite{Hinton2002a} for training the energy-based model, such as deep Boltzmann machine \cite{salakhutdinov2009deep} and deep belief network \cite{Hinton06}.   The contrastive divergence initializes the finite-step MCMC sampling from the observed examples. Our method initializes the MCMC sampling from  the generator that seeks to approximate the descriptor, so that the learning is closer to maximum likelihood. 

Our work  is similar to the recent  work of \cite{Bengio2016}.  In \cite{Bengio2016}, the generator learns from the energy-based model by minimizing the Kullback-Leibler divergence from the generator model to the energy-based model, which can be decomposed into an energy term and an entropy term. In our work, the energy-based descriptor model teaches the generator via MCMC teaching. Our method does not need to approximate the intractable entropy term. 

Another method for training the generator network is variational auto-encoder (VAE) \cite{KingmaCoRR13,RezendeICML2014,MnihGregor2014}, which learns an inferential or recognition network. The MCMC teaching in our work avoids the challenging problem of inferring the latent variables from the observed examples. 

%Our work is somewhat related to the generative stochastic network of \cite{thibodeau2014deep}, which learns the Markov transition probability directly. In our work, the generator serves to absorb the {cumulative} effect of all the past Markov transitions (i.e., the Langevin revisions) of the descriptor and reproduce it in one shot. 

Our work is related to knowledge distilling \cite{hinton2015distilling}. In our work, the descriptor distills its MCMC algorithm to the generator through MCMC teaching. 

Building on the pioneering work of \cite{tu2007learning}, recently \cite{lazarow2017introspective, jin2017introspective, lee2017wasserstein} have developed an introspective learning method to learn the energy-based model, where the energy function is discriminatively learned, and the learned energy function is used to generate synthesized examples via Langevin dynamics. It can be interesting to combine introspective learning with the proposed cooperative learning method that recruits a generator to jumpstart the Langevin sampling. 

Another recently proposed model is pixelCNN \cite{oord2016conditional}, which learns an autoregressive model. Unlike pixelCNN, the descriptor and generator networks model the joint distribution of the image pixels directly without factorizing it into a sequence of conditional distributions. 

This paper is an expansion of our conference paper \cite{xie2018CoopNets}. It is also related to our recent papers on 3D descriptor model \cite{xie2018learning} and the spatial-temporal descriptor model \cite{xie2017synthesizing}. 

\section{Two models and two algorithms} 

{\bf $(p_\theta, q_\alpha)$ notation.}  Let $Y$ be the $D$-dimensional signal, such as an image. We use $p(Y; \theta)$ or $p_\theta$ to denote the probability distribution of the descriptor network (energy-based model), where $\theta$ denotes the parameters of the bottom-up ConvNet. We use $q(Y; \alpha)$ or $q_\alpha$ to denote the probability distribution of the generator network (latent variable model), where $\alpha$ denotes the parameters of the top-down ConvNet.

\subsection{Energy-based model and maximum likelihood learning} 

As an energy-based model, the descriptor network is in the form of exponential tilting of a reference distribution  \cite{XieLuICML}: 
\begin{eqnarray} 
   p(Y; \theta) = \frac{1}{Z(\theta)} \exp\left[ f(Y; \theta)\right] p_0(Y). 
\end{eqnarray}
 $p_0(Y)$ is the reference distribution such as Gaussian white noise 
\begin{eqnarray} 
 p_0(Y) = \frac{1}{(2\pi s^2)^{D/2}} \exp \left[ - \frac{\|Y\|^2}{2 s^2}\right], 
 \end{eqnarray}
 where $D$ is the dimensionality of  the image $Y$, and $Y \sim \N(0, s^2 I_D)$ under $p_0(Y)$  ($I_D$ denotes the $D$-dimensional identity matrix). 
 $f(Y; \theta)$ is a ConvNet whose parameters are denoted by $\theta$. This ConvNet is bottom-up because it maps the image $Y$ to the energy. See the diagram in (\ref{eq:diagram0}). 
$
Z(\theta) = \int  \exp\left[ f(Y; \theta)\right] p_0(Y) dY 
$ is the normalizing constant, and this integral is analytically intractable. The energy function of the model is 
\begin{eqnarray}
{\cal E}(Y; \theta) = \frac{1}{2s^2} \|Y\|^2 - f(Y; \theta).
\end{eqnarray}
 $p_0$ can also be taken to be the uniform measure, and in that case, ${\cal E}(Y; \theta) = - f(Y; \theta)$. 
  
 Suppose we observe training examples $\{Y_i, i = 1, ..., n\}$ from an unknown data distribution $\P(Y)$. The maximum likelihood learning maximizes the log-likelihood function 
 \begin{eqnarray} 
   L_{p}(\theta) = \frac{1}{n} \sum_{i=1}^{n} \log p(Y_i; \theta). 
\end{eqnarray} 
If the sample size $n$ is large, the maximum likelihood estimator minimizes $\KL(\P|p_\theta)$, the Kullback-Leibler divergence from the data distribution $\P$ to the model distribution $p_\theta$. The log-likelihood is analytically intractable because of the intractable integral $Z(\theta)$.

The gradient of the log-likelihood $L_{p}(\theta)$ is 
  \begin{eqnarray} 
  L_{p}'(\theta)& =& \frac{1}{n} \sum_{i=1}^{n} \frac{\partial}{\partial \theta} f(Y_i; \theta) -  \E_{\theta} \left[\frac{\partial}{\partial \theta} f(Y; \theta)\right],  \label{eq:lD}
\end{eqnarray} 
where  $\E_{\theta}$ denotes the expectation with respect to $p(Y; \theta)$. Equation (\ref{eq:lD}) follows from $\frac{\partial}{\partial \theta}  \log Z(\theta) = \E_{\theta}[\frac{\partial}{\partial \theta}  f(Y; \theta)]$. 

The expectation in equation (\ref{eq:lD}) is analytically intractable and has to be approximated by MCMC, such as Langevin revision dynamics, which iterates the following steps: 
\begin{eqnarray}
   Y_{\tau+1} &=& Y_\tau -  \frac{\delta^2}{2}  \frac{\partial}{\partial Y} {\cal E}(Y_\tau; \theta) + \delta U_{\tau} \nonumber \\
   &=& Y_\tau - \frac{\delta^2}{2} \left[ \frac{Y_\tau}{s^2} - \frac{\partial}{\partial Y} f(Y_\tau; \theta) \right] + \delta U_{\tau},  \label{eq:LangevinD}
\end{eqnarray}
where $\tau$ indexes the time steps of  Langevin dynamics, $\delta$ is the step size, and $U_{\tau} \sim \N(0, I_D)$ is the Gaussian white noise term.  
%The Langevin dynamics can be justified by stochastic differential equation. A more intuitive understanding of the one-dimensional case is to discretize the real line into equally spaced grid points, and consider a Metropolis algorithm whose base chain is a nearest neighbor random walk on this grid. The mean and variance of the one-step displacement of the Metropolis algorithm corresponds to the gradient term and the Gaussian noise term of the Langevin equation (\ref{eq:LangevinD}). 
The Langevin dynamics is a process of stochastic relaxation. The gradient term seeks to reduce the energy function  ${\cal E}(Y; \theta)$ while the noise term provides the randomness to increase the entropy.  A Metropolis-Hastings step can be added to correct for the finite step size. 

We can run $\tn$ parallel chains of Langevin dynamics according to (\ref{eq:LangevinD}) to obtain the synthesized examples  $\{\tY_i, i = 1, ..., \tn\}$. The Monte Carlo approximation to $L_{p}'(\theta)$ is 
\begin{eqnarray} 
  \label{eq:learningD}
  L_{p}'(\theta) &\approx&  \frac{1}{n} \sum_{i=1}^{n} \frac{\partial}{\partial \theta} f(Y_i; \theta)- \frac{1}{\tn} \sum_{i=1}^{\tn} \frac{\partial}{\partial \theta} f(\tY_i; \theta)  \\
  %&=& \frac{\partial}{\partial \theta} \left[\frac{1}{n} \sum_{i=1}^{n}  f(Y_i; \theta)- \frac{1}{\tn} \sum_{i=1}^{\tn}  f(\tY_i; \theta)\right]\\
    &=& \frac{\partial}{\partial \theta} \left[\frac{1}{\tn} \sum_{i=1}^{\tn}  {\cal E}(\tY_i; \theta) - \frac{1}{n} \sum_{i=1}^{n}  {\cal E}(Y_i; \theta)\right],  \nonumber
\end{eqnarray} 
which is the difference between the observed examples and the synthesized examples. See \cite{LuZhuWu2016} for details. 

\begin{algorithm}
\caption{Algorithm D}
\label{code:1}
\begin{algorithmic}[1]
\Require
\Statex (1) training examples $\{Y_i, i=1,...,n\}$
\Statex (2) number of Langevin steps $l_p$
\Statex (3) number of learning iterations $T$
\Ensure
\Statex (1) estimated parameters $\theta$
\Statex (2) synthesized examples $\{\tY_i, i= 1, ..., \tilde{n}\}$ 
\item[]
\State  Let $t\leftarrow 0$, initialize $\theta$.
\State Initialize $\tY_i$,  $i = 1, ..., \tilde{n}$. 
\Repeat 
\State {\bf Step D1 Langevin revision}:  For each $i$, run $l_p$ steps of Langevin dynamics to update $\tY_i$, i.e., starting from the current $\tY_i$, each step 
follows equation (\ref{eq:LangevinD}). 
\State {\bf Step D2 density shifting}: Update $\theta^{(t+1)} = \theta^{(t)} + \gamma_t L_{p}'(\theta^{(t)})$,   with learning rate $\gamma_t$, where $L_{p}'(\theta^{(t)})$ is computed according to (\ref{eq:learningD}). 
\State Let $t \leftarrow t+1$
\Until $t = T$
\end{algorithmic}
\end{algorithm}

Algorithm   \ref{code:1}  \cite{XieLuICML} describes the training algorithm.  See Figure \ref{fig:diagramD} for an illustration. Step D1 (Langevin revision) tends to settle the synthesized examples at the low energy (high density) regions. Step D2 (density shifting) tends to shift low energy (high density) regions from the synthesized examples to the observed examples. 
 Step D1 needs to compute $\frac{\partial}{\partial Y} f(Y; \theta)$. Step D2 needs to compute $\frac{\partial}{\partial \theta} f(Y; \theta)$. The computations of both derivatives can be powered by back-propagation. Because of the ConvNet structure of $f(Y; \theta)$, the computations of the two derivatives share most of their steps in the chain rule computations. 
   
Because the parameter $\theta$ keeps changing in the learning process, the energy landscape and the local energy minima also keep changing.  This may help the Langevin revision dynamics avoid being trapped by the local energy minima.

Algorithm D is a stochastic approximation algorithm \cite{robbins1951stochastic},  except that the synthesized examples are obtained by a finite number of Langevin steps in each learning iteration. The convergence of an algorithm of this type to the maximum likelihood estimate has been studied by \cite{younes1999convergence}. 

 {\bf Contrastive divergence.} If we initialize the synthesized examples $\{\tY_i\}$ from the observed examples $\{Y_i\}$,  the learning algorithm becomes persistent contrastive divergence \cite{tieleman2008training}. 

{\bf Zero temperature limit}. We can write the model $p(Y; \theta) = \exp(-{\cal E}(Y; \theta)/T)/Z_T$ where $T$ is the temperature term ($T = 1$ in our model). At the zero-temperature limit $T \rightarrow 0$, we can disable the noise term in the Langevin dynamics, so that Step D1 (Langevin revision) becomes gradient descent towards the local minima of the energy function. Define the value function 
\begin{eqnarray}
   V(\theta, \tY_i, i = 1, ..., \tn)  = \frac{1}{\tn} \sum_{i=1}^{\tn}  {\cal E}(\tY_i; \theta) - \frac{1}{n} \sum_{i=1}^{n}  {\cal E}(Y_i; \theta), 
\end{eqnarray} 
then the learning algorithm solves the following minimax game 
\begin{eqnarray}
   \max_\theta \min_{\{\tY_i\}} V(\theta, \tY_i, i = 1, ..., \tn), 
   \end{eqnarray} 
where Step D1  (Langevin revision) seeks to decrease $V$, while Step D2  (density shifting) seeks to increase $V$. 

The energy-based model can be used for inverse optimal control \cite{ziebart2008maximum, abbeel2004apprenticeship}, where the Langevin dynamics or the gradient descent algorithm in Step D1 can be considered an optimal control algorithm, and the energy function can be considered the cost function or critic, which is updated in Step D2. 

The energy-based model is also related to the Hopfield model for content addressable memory \cite{hopfield1982neural}. The Langevin dynamics in Step D1 can be considered an attractor dynamics towards the local modes for memory recall \cite{seung1998learning}, while Step D2 shifts the local modes towards the observed examples. 

\subsection{Latent variable model and maximum likelihood learning} 

As a latent variable model, the generator network seeks to explain the image $Y$ of dimension $D$ by a vector of latent factors $X$ of dimension $d$, and usually $d \ll D$.  The model is of the following form: 
\begin{eqnarray} 
  &&X \sim \N(0, I_d),  \nonumber\\ 
  &&Y = g(X; \alpha) + \epsilon,  \; \epsilon \sim \N(0, \sigma^2  I_D).  \label{eq:FA}
 \end{eqnarray}
$g(X; \alpha)$  is a top-down ConvNet defined by the parameters $\alpha$. The ConvNet $g$  maps the latent factors $X$ to the image $Y$. See the diagram in (\ref{eq:diagram0}). 
Model (\ref{eq:FA}) is a directed graphical model, where $Y$ can be readily generated by ancestral sampling: first sampling $X$ from its known prior distribution $\N(0, I_d)$ and then transforming $X$ to $Y$ via $g$. 

The joint density  is $q(X, Y; \alpha) = q(X) q(Y|X; \alpha)$, and 
\begin{eqnarray} 
  \log q(X, Y; \alpha) &=& - \frac{1}{2\sigma^2} \|Y - g(X; \alpha)\|^2 \nonumber \\
  &-& \frac{1}{2} \|X\|^2 + {\rm constant}, \label{eq:complete}
\end{eqnarray} 
where the constant term is independent of $X$, $Y$ and $\alpha$. 
The marginal density is obtained by integrating out the latent factors $X$, i.e., $q(Y; \alpha) = \int q(X, Y; \alpha) dX$. This integral is analytically intractable. The inference of $X$ given $Y$ is based on the posterior density $q(X|Y; \alpha) = q(X, Y; \alpha)/q(Y; \alpha) \propto q(X, Y; \alpha)$ as a function of $X$. 

For the training data $\{Y_i, i = 1, ..., n\}$, the generator network can  be trained by maximizing the log-likelihood 
\begin{eqnarray} 
  L_{q}(\alpha) = \frac{1}{n}  \sum_{i=1}^{n} \log q(Y_i; \alpha). 
\end{eqnarray} 
The log-likelihood is intractable because the marginal distribution $q(Y; \alpha)$ is an intractable integral. 
If the sample size $n$ is large, the maximum likelihood estimator minimizes the Kullback-Leibler divergence $\KL(\P|q_\alpha)$ from the data distribution $\P$ to the model distribution $q_\alpha$.

The gradient of $L_{q}(\alpha)$ is obtained according to the following identity 
\begin{eqnarray} 
    \frac{\partial}{\partial \alpha} \log q(Y; \alpha)    &=&  \frac{1}{q(Y; \alpha)}  \frac{\partial}{\partial \alpha}  \int q(Y, X; \alpha) dX \nonumber \\
%&=& \frac{1}{q(Y; \alpha)} \int \left[\frac{\partial}{\partial \alpha} \log q(Y, X; \alpha) \right] q(Y, X; \alpha) dX \nonumber  \\
&=& \int \left[\frac{\partial}{\partial \alpha} \log q(Y, X; \alpha) \right] \frac{q(Y, X; \alpha)}{q(Y; \alpha)} dX  \nonumber \\
&=& \E_{q(X|Y; \alpha)} \left[ \frac{\partial}{\partial \alpha} \log q(X, Y;  \alpha)\right].  \label{eq:EM}
 \end{eqnarray}
The above identity underlies the EM algorithm, 
where $\E_{q(X|Y; \alpha)}$ is the expectation with respect to the posterior distribution of the latent factors $q(X|Y; \alpha)$, and is computed in the E-step. 
  The usefulness of identify (\ref{eq:EM}) lies in the fact that the derivative of the complete-data log-likelihood $ \log q(X, Y;  \alpha)$ on the right hand side can be obtained in closed form.  
 % In the EM algorithm, the M-step maximizes the expectation of $ \log q(X, Y;  \alpha)$  with respect to the current posterior distribution of the latent factors. 
 
In general, the expectation in (\ref{eq:EM}) is analytically intractable even though the term inside the expectation can be easily computed, and the expectation has to be approximated by MCMC that samples from the posterior $q(X|Y; \alpha)$, such as the Langevin inference dynamics, which iterates 
 \begin{eqnarray}
   X_{\tau + 1} = X_\tau + \frac{\delta^2}{2}  \frac{\partial}{\partial X}  \log q(X_\tau, Y;  \alpha) + \delta U_\tau,  \label{eq:LangevinG}
      \end{eqnarray}
 where  $\tau$ indexes the time step, $\delta$ is the step size, and for notational simplicity, we continue to use $U_\tau$ to denote the noise term, but here $U_\tau \sim \N(0, I_d)$.  
 We take the derivative of $\log q(X, Y; \alpha)$ in (\ref{eq:LangevinG}) because this derivative is the same as the derivative of the log-posterior  $\log q(X|Y; \alpha)$, since $q(X|Y; \alpha) $ is proportional to $q(X, Y; \alpha)$ as a function of $X$.  
 The Langevin inference solves a $\ell_2$ penalized non-linear least squares problem so that $X_i$ can reconstruct $Y_i$ given the current $\alpha$. The Langevin inference process performs explaining-away reasoning, where the latent factors in $X$ compete with each other to explain the current residual $Y - g(X; \alpha)$. 
 
With  $X_i$ sampled from  $q(X_i \mid Y_i; \alpha)$ for each observation $Y_i$ by the Langevin inference process,  the Monte Carlo approximation to $L_{q}'(\alpha)$ is 
 \begin{eqnarray} 
  L_{q}'(\alpha) &\approx& \frac{1}{n}  \sum_{i=1}^{n} \frac{\partial}{\partial \alpha} \log q(X_i, Y_i;  \alpha) \nonumber \\
  &=& \frac{1}{n}  \sum_{i=1}^{n} \frac{1}{\sigma^2} \left(Y_i - g(X_i; \alpha)\right)\frac{\partial}{\partial \alpha} g(X_i; \alpha). \label{eq:learningG} 
\end{eqnarray} 
The updating of $\alpha$ solves a non-linear regression problem, so that the learned $\alpha$ enables  better  reconstruction of $Y_i$ by the inferred $X_i$. Given the inferred $X_i$, the learning of $\alpha$ is a supervised learning problem \cite{Alexey2015}.

\begin{algorithm}
\caption{Algorithm G}
\label{code:2}
\begin{algorithmic}[1]

\Require
\Statex(1)  training examples $\{Y_i, i=1,...,n\}$ 
\Statex(2) number of Langevin steps $l_q$
\Statex(3) number of learning iterations $T$

\Ensure
\Statex(1) estimated parameters $\alpha$
\Statex(2) inferred latent factors $\{X_i, i =  1, ..., n\}$ 

\item[]
\State Let $t\leftarrow 0$, initialize $\alpha$.
\State Initialize $X_i$,  $i =  1, ..., {n}$. 
\Repeat 
\State {\bf Step G1 Langevin inference}: For each $i$, run $l_q$ steps of Langevin dynamics to update $X_i$, i.e., starting from the current $X_i$, each step 
follows equation (\ref{eq:LangevinG}). 
\State {\bf Step G2 reconstruction}: Update $\alpha^{(t+1)} = \alpha^{(t)} + \gamma_t {L_{q}}'(\alpha^{(t)}) $,  with learning rate  $\gamma_t$, where ${L_{q}}'(\alpha^{(t)})$ is computed according to equation (\ref{eq:learningG}). 
\State Let $t \leftarrow t+1$
\Until $t = T$
\end{algorithmic}
\end{algorithm}

 Algorithm   \ref{code:2}   \cite{HanLu2016} describes the training algorithm. See Figure \ref{fig:diagramG} for an illustration. 
  Step G1 (Langevin inference) needs to compute $\frac{\partial}{\partial X} g(X; \alpha)$. Step G2 (reconstruction) needs to compute $\frac{\partial}{\partial \alpha} g(X; \alpha)$. The computations of both derivatives can be powered by back-propagation, and the computations of the two derivatives share most of their steps in the chain rule computations. 
  
 Algorithm G  is a stochastic approximation or stochastic gradient algorithm that converges to the maximum likelihood estimate  \cite{younes1999convergence}. 
 
 The generator network can be considered a generalization of the factor analysis model. In factor analysis,  the mapping from the latent factors to the signal is linear. The mapping becomes non-linear in generator network. 
 
 As shown in \cite{HanLu2016}, the generator network can be learned directly from the incomplete data. It can also be used for pattern completion in the testing data, which may be considered as an alternative approach to content addressable memory \cite{hopfield1982neural}. 

\section{Cooperative training} \label{sect:CoopNets}

\subsection{Cooperation of two algorithms}

In Algorithm D and Algorithm G, both steps D1 (Langevin revision) and G1 (Langevin inference) are Langevin dynamics. They may be slow to converge and may become bottlenecks in their respective algorithms. An interesting observation is that the two algorithms can cooperate with each other by jumpstarting each other's Langevin sampling.

Specifically, in Step D1 (Langevin revision), we can initialize the synthesized examples by generating examples from the generator network, which does not require MCMC, because the generator network is a directed graphical model. More specifically, we first generate $\hX_i \sim \N(0, I_d)$, and then generate $\hY_i = g(\hX_i; \alpha) + \epsilon_i$, for $i = 1, ..., \tn$. If the current generator $q$ is close to the current descriptor $p$,  then the generated $\{\hY_i\}$ should be a good initialization for sampling from the descriptor network, i.e., starting from the $\{\hY_i, i = 1, ..., \tn\}$ supplied by the generator network, we run Langevin dynamics in Step D1 (Langevin revision) for $l_p$ steps to get  $\{\tY_i, i = 1, ..., \tn\}$, which are revised versions of $\{\hY_i\}$.  These $\{\tY_i\}$ can be used as the synthesized examples from the descriptor network.  We can then update $\theta$ according to Step D2 (density shifting) of Algorithm D. This is modified contrastive divergence. 

In order to update $\alpha$ of the generator network, we treat the $\{\tY_i, i = 1, ..., \tn\}$ produced by the above D1 (Langevin revision) step as the training data for the generator. Since these $\{\tY_i\}$ are obtained by the Langevin revision dynamics initialized from the $\{\hY_i, i = 1, ..., \tn\}$ produced by the generator network with known latent factors $\{\hX_i, i = 1, ..., \tn\}$, we can update $\alpha$ by learning from $\{(\tY_i, \hX_i), i = 1, ..., \tn\}$, which is a supervised learning problem, or more specifically, a non-linear regression of $\tY_i$ on $\hX_i$. At $\alpha^{(t)}$,  the latent factors $\hX_i$ generates and thus reconstructs the initial example  $\hY_i$. After updating $\alpha$, we want $\hX_i$ to reconstruct the revised example $\tY_i$. That is, we revise $\alpha$ to absorb the revision from $\hY_i$ to $\tY_i$, so that the generator shifts its density from $\{\hY_i\}$ to $\{\tY_i\}$. This is MCMC teaching. 
The reconstruction error can tell us whether the generator has caught up with the descriptor by fully absorbing the revision. 

The diagrams in (\ref{eq:m}) illustrate the basic idea of MCMC teaching:
\begin{eqnarray}
\begin{tikzpicture}
  \matrix (m) [matrix of math nodes,row sep=3em,column sep=4em,minimum width=2em]
  {
     \hX_i & \\
      \hY_i & \tY_i \\};
  \path[-stealth]
    (m-1-1) edge [double] node [left] {$\alpha^{(t)}$} (m-2-1)
        (m-1-1) edge [double] node [right] {$\alpha^{(t+1)}$} (m-2-2)      
            (m-2-1) edge [dashed]  node [below] {$\theta^{(t)}$} (m-2-2);
                     \end{tikzpicture}  
            \begin{tikzpicture}
  \matrix (m) [matrix of math nodes,row sep=3em,column sep=4em,minimum width=2em]
  {
     \hX_i & X_i \\
      \hY_i & \tY_i \\};
  \path[-stealth]
      (m-1-1) edge  [dashed]  node [above] {$\alpha^{(t)}$} (m-1-2)
    (m-1-1) edge [double] node [left] {$\alpha^{(t)}$} (m-2-1)
        (m-1-2) edge [double]  node [right] {$\alpha^{(t+1)}$} (m-2-2)
          (m-2-1) edge  [dashed]   node [below] {$\theta^{(t)}$} (m-2-2);
            \end{tikzpicture}      
            \label{eq:m} 
            \end{eqnarray}
In the two diagrams in (\ref{eq:m}),  the double line arrows indicate generation and reconstruction in the generator network, while the dashed line arrows indicate Langevin dynamics for revision and inference in the two nets. 
The diagram on the right in (\ref{eq:m}) illustrates a more rigorous method, where we initialize the Langevin inference  of $\{X_i, i = 1, ..., \tn\}$ in Step G1 (Langevin inference) from $\{\hX_i\}$, and  then update $\alpha$ in Step G2 (reconstruction) based on $\{(\tY_i, X_i), i = 1, ..., \tn\}$.  The diagram on the right shows how the two nets jumpstart each other's MCMC.

\begin{algorithm}[h]
\caption{CoopNets Algorithm}
\label{code:3}
\begin{algorithmic}[1]

\Require
\Statex (1)  training examples $\{Y_i, i=1,...,n\}$ 
\Statex (2) numbers of Langevin steps $l_p$ and $l_q$
\Statex (3) number of learning iterations $T$

\Ensure
\Statex (1) estimated parameters $\theta$ and $\alpha$
\Statex (2) synthesized examples $\{\hY_i, \tY_i, i= 1, ..., \tilde{n}\}$ 

\item[]
\State Let $t\leftarrow 0$, initialize $\theta$ and $\alpha$.
\Repeat 
\State {\bf Step G0 Initial generation}: For $i = 1, ..., \tn$, generate $\hX_i \sim \N(0, I_d)$, and generate $\hY_i = g(\hX_i; \alpha^{(t)}) + \epsilon_i$. 
\State {\bf Step D1 Langevin revision}: For $i = 1, ..., \tn$,  starting from $\hY_i$, run $l_p$ steps of Langevin revision dynamics to obtain $\tY_i$,  each step 
following equation (\ref{eq:LangevinD}). 
\State {\bf Step G1 Langevin inference}: Treat the current $\{\tY_i, i = 1, ..., \tn\}$ as the training data, for each $i$, 
infer $X_i = \hX_i$. Or more rigorously, 
starting from $X_i = \hX_i$, run $l_q$ steps of Langevin inference dynamics to update $X_i$, each step 
following equation (\ref{eq:LangevinG}). 
\State {\bf Step D2 Density shifting}: Update $\theta^{(t+1)} = \theta^{(t)} + \gamma_t L_{p}'(\theta^{(t)})$,  where $L_{p}'(\theta^{(t)})$ is computed according to (\ref{eq:learningD}). 
\State {\bf Step G2 Reconstruction}: Update $\alpha^{(t+1)} = \alpha^{(t)} + \gamma_t {L_{q}}'(\alpha^{(t)}) $,  where ${L_{q}}'(\alpha^{(t)})$ is computed according to equation (\ref{eq:learningG}), except that $Y_i$ is replaced by $\tY_i$, and $n$ by $\tn$. We can run multiple iterations of Step G2 to learn from and reconstruct $\{\tY_i\}$,  and  to allow the generator to catch up with the descriptor. 
\State Let $t \leftarrow t+1$
\Until $t = T$
\end{algorithmic}
\end{algorithm}

Algorithm  \ref{code:3} describes the cooperative training that interweaves Algorithm D and Algorithm G.  For ease of reference, we call the algorithm the CoopNets algorithm. See Figure \ref{fig:diagram}  for  the flow chart of the CoopNets algorithm. 

The following are the special cases of the above algorithm. 

{\bf Special case (1):} In Step G1, let $l_q = 0$, i.e., disable the Langevin inference process, i.e., the left diagram of (\ref{eq:m}).  We use this algorithm in our experiments and it works well. 

{\bf Special case (2):}  In addition to (1), in Step D1, let $l_p = 1$, i.e., we only run one step of Langevin revision. This algorithm is very efficient, and works well according to our experience. 

The learning of both the descriptor and the generator follows the ``analysis by synthesis'' principle \cite{grenander2007pattern}. There are three sets of synthesized examples (S stands for synthesis). 
Data (S1): Initial synthesized examples $\{\hY_i\}$ generated by Step G0 (initial generation). 
Data (S2): Revised synthesized examples $\{\tY_i\}$ produced by Step D1 (Langevin revision). 
Data (S3): Reconstructed synthesized examples $\{g(X_i; \alpha^{(t+1)})\}$ produced by Step G2 (reconstruction). 
The descriptor shifts its density from (S2) towards the observed data, while the generator shifts its density from (S1) towards (S2). 

The evolution from (S1) to (S2) is the work of the descriptor. It is a process of stochastic relaxation that settles the synthesized examples in the low energy regions of the descriptor. The descriptor works as an associative memory, with (S1) being the cue, and (S2) being the recalled memory. It serves as a feedback to the generator. The reconstruction of  (S2) by (S3) is the work of the generator that seeks to absorb the feedback conveyed by the evolution from (S1) to (S2). The descriptor can test whether the generator learns well by checking whether (S3) is close to (S2). The two nets collaborate and communicate with each other via synthesized data. 

The general idea of the interaction between MCMC and the generator can be illustrated by the following diagram, 
\begin{eqnarray}
\begin{array}[c]{cccc}
{\rm MCMC:} &P^{(t)}&\stackrel{\rm Markov \; transition}{\xrightarrow{\hspace*{2cm}}}&P^{(t+1)}\\
& \Updownarrow&&\Updownarrow\\
{\rm Generator:} & \alpha^{(t)}&\stackrel{\rm Parameter\; updating}{\xrightarrow{\hspace*{2cm}}}&\alpha^{(t+1)}  
\end{array}  \label{eq:diagram}
\end{eqnarray}
where $P^{(t)}$ is the marginal distribution of MCMC and $\alpha^{(t)}$ is the parameter of the generator which traces the evolution of the marginal distribution in MCMC by absorbing  the cumulative effect of all the past Markov transitions.

In traditional MCMC, we only have access to Monte Carlo samples, instead of their marginal distributions, which exist only theoretically but are analytically intractable. However, with the generator net, we can actually implement MCMC at the level of the whole distributions, instead of a number of Monte Carlo samples, in the sense that after learning $\alpha^{(t)}$ from the existing samples of $P^{(t)}$, we can replace the existing samples by fresh new samples to rejuvenate the Markov chains, by sampling from the generator defined by the learned $\alpha^{(t)}$, which is illustrated by the two-way arrow between $P^{(t)}$ and $\alpha^{(t)}$. Effectively, the generator powers MCMC by implicitly running an infinite number of parallel chains. Conversely, the MCMC does not only drive the evolution of the samples, it also drives the evolution of a generator model. 

\subsection{Theoretical understanding} 

In the CoopNets algorithm, Steps G0 (initial generation), D1 (Langevin revision), and D2 (density shifting) are modified contrastive divergence, and Steps G0 (initial generation), G1 (Langevin inference), and G2 (reconstruction) are MCMC teaching. 

(1) Modified contrastive divergence for descriptor (energy-based model). In the traditional contrastive divergence \cite{Hinton2002a}, $\hY_i$ in Step D1 (Langevin revision) is taken to be the observed $Y_i$. In cooperative learning, $\hY_i$ is generated by $q(Y; \alpha^{(t)})$. Let $\M_\theta$ be the Markov transition kernel of $l_p$ steps of Langevin dynamics that samples $p_\theta$. For any distribution $p$ and any Markov transition kernel $\M$, let $\M p$ be the marginal distribution obtained by running the Markov transition $\M$ from $p$.  Then similar to the traditional contrastive divergence, the learning gradient of the descriptor $\theta$ at iteration $t$ is the gradient of 
\begin{eqnarray}
\KL(\P|p_\theta) - \KL(\M_{\theta^{(t)}} q_{\alpha^{(t)}}|p_\theta) \label{eq:MCD}
\end{eqnarray}
 with respect to $\theta$. In the traditional contrastive divergence, $\P$ takes the place of $q_{\alpha^{(t)}}$ in the second KL-divergence.

\begin{figure}
\begin{center}
\includegraphics[height=.3\linewidth]{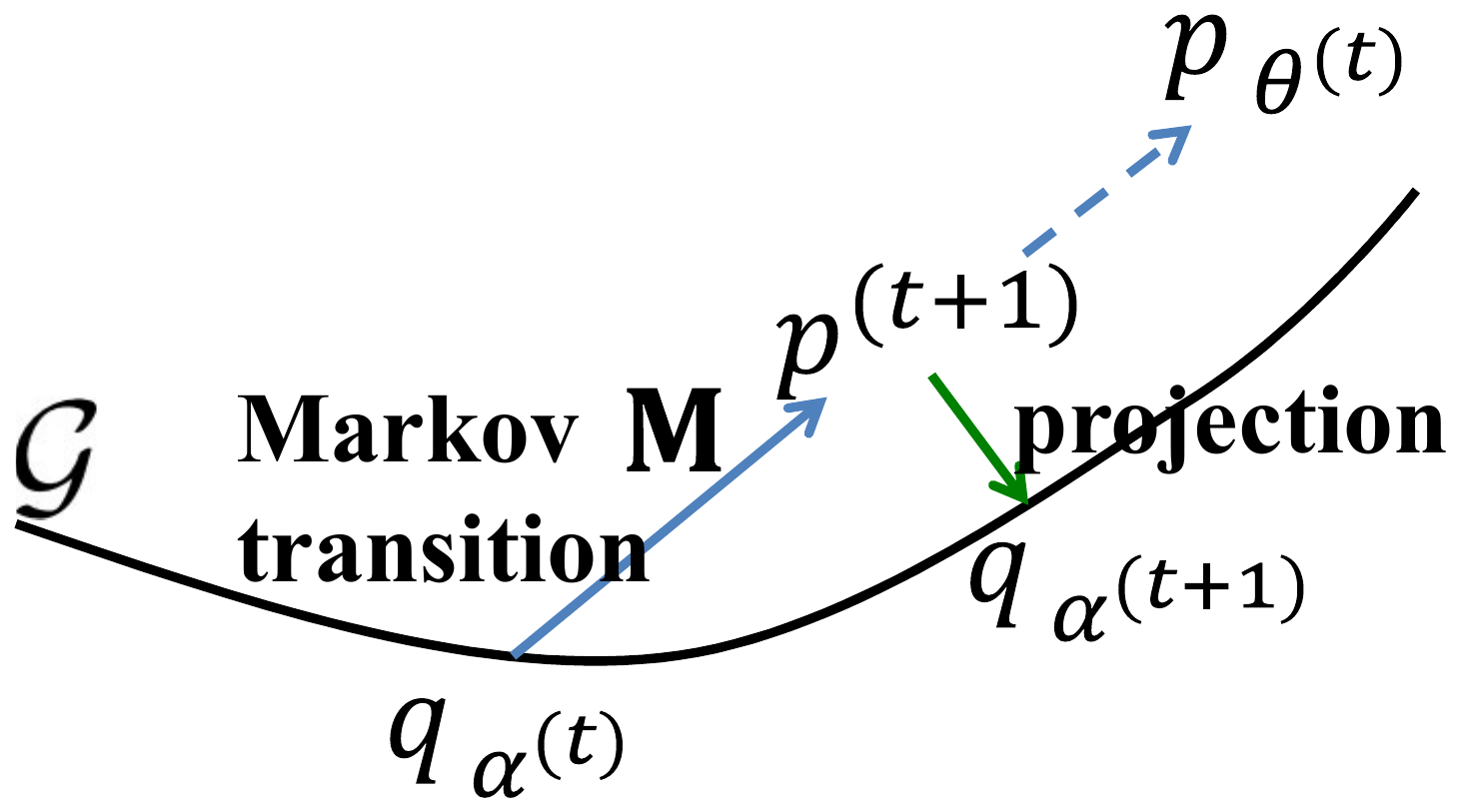}
\caption{The MCMC teaching of the generator alternates between Markov transition and projection. The family of the generator models $\G$ is illustrated by the black curve. Each distribution is illustrated by a point. }
\label{fig:LP}
\end{center}
\end{figure}

(2) MCMC teaching of the generator model (latent variable model). The learning gradient of the generator $\alpha$ in the right diagram of (\ref{eq:m})  is the gradient of  
\begin{eqnarray}
\KL(\M_{\theta^{(t)}} q_{\alpha^{(t)}}|q_\alpha) \label{eq:MT}
\end{eqnarray}
with respect to $\alpha$. Here  $p^{(t+1)} = \M_{\theta^{(t)}} q_{\alpha^{(t)}}$ takes the place of $\P$ as the data to train the generator model. It is much easier to minimize  $\KL(\M_{\theta^{(t)}} q_{\alpha^{(t)}}|q_\alpha)$ than minimizing $\KL(\P|q_\alpha)$ because the latent variables are essentially known in the former, so that the learning is supervised. The MCMC teaching alternates between Markov transition from $q_{\alpha^{(t)}}$ to $p^{(t+1)}$,  and projection from $p^{(t+1)}$ to $q_{\alpha^{(t+1)}}$, as illustrated by Figure \ref{fig:LP}.

Assume the learning algorithm converges to a fixed point $(\hat{\theta}, \hat{\alpha})$, then 
\begin{eqnarray} 
\hat{\theta} &=& \arg\min_\theta \left[\KL(\P|p_\theta) - \KL(\M_{\hat{\theta}} q_{\hat{\alpha}}|p_\theta)\right], \label{eq:T1}\\
\hat{\alpha} &=& \arg\min_\alpha \KL(\M_{\hat{\theta}} q_{\hat{\alpha}} | q_\alpha). \label{eq:T2}
\end{eqnarray}
(assuming $\hat{\theta}$ and  $\hat{\alpha}$ are local minima). 
Equation (\ref{eq:T2}) tells us that $q_{\hat{\alpha}}$ seeks to be the stationary distribution of $\M_{\hat{\theta}}$, and the stationary distribution is nothing but $p_{\hat{\theta}}$. In the idealized scenario where the generator  $q_\alpha$ has infinite capacity, so that $\min_\alpha \KL(\M_{\hat{\theta}} q_{\hat{\alpha}} | q_\alpha) = 0$, then $q_{\hat{\alpha}} = \M_{\hat{\theta}} q_{\hat{\alpha}}$, so that $q_{\hat{\alpha}}$ is the stationary distribution of $\M_{\hat{\theta}}$, which is $p_{\hat{\theta}}$, thus $q_{\hat{\alpha}} = p_{\hat{\theta}}$. As a consequence, the second divergence in (\ref{eq:T1}) vanishes, i.e., $\KL(\M_{\hat{\theta}} q_{\hat{\alpha}}|p_{\hat{\theta}}) = 0$, so that $\hat{\theta}$ becomes maximum likelihood estimate that minimizes the first KL-divergence $\KL(\P|p_\theta)$. 

To further understand the dynamics of the MCMC teaching in this idealized scenario, suppose the descriptor $p_\theta$ is fixed, and it teaches the generator $q_\alpha$ by the MCMC teaching such that $\alpha^{(t+1)} = \arg\min_\alpha \KL(\M_\theta q_{\alpha^{(t)}}|q_\alpha)$, then $q_{\alpha^{(t+1)}} = \M_\theta q_{\alpha^{(t)}}$, so that $q_{\alpha^{(t)}} = \M_\theta^t q_{\alpha^{(0)}} \rightarrow p_\theta$, i.e., $q_{\alpha}$ accumulates the MCMC transitions and convergences to the stationary distribution $p_\theta$. 

As is the case with the traditional contrastive divergence, the analysis of the finite capacity situation can be rather involved. We leave it to future investigation, while relying on empirical evaluations in this paper. 

\cite{Bengio2016}  learned the generator  model by gradient descent on $\KL(q_{\alpha}|p_{\theta^{(t)}})$ over $\alpha$. In fact their learning objective is 
\begin{eqnarray}
\min_\theta \max_\alpha [\KL(\P|p_\theta) - \KL(q_\alpha|p_\theta)]. \label{ACD}
\end{eqnarray} 
The objective function for $\alpha$  is 
\begin{eqnarray}
\KL(q_\alpha|p_{\theta^{(t)}}) = \E_{q_\alpha}[\log q(Y; \alpha)] - \E_{q_\alpha}[\log p(Y; \theta^{(t)})],
\end{eqnarray}
 where the first term is the negative entropy that is intractable, and the second term is the expected energy that is tractable. Our MCMC teaching of the generator is consistent with the learning objective $\KL(q_\alpha|p_{\theta^{(t)}})$,  because 
\begin{eqnarray}
 \KL(p^{(t+1)} |p_{\theta^{(t)}}) \leq  \KL(q_{\alpha^{(t)}} |p_{\theta^{(t)}}). \label{eq:mono}
\end{eqnarray}
In fact, $\KL(p^{(t+1)} |p_{\theta^{(t)}}) \rightarrow 0$ monotonically as $l_p \rightarrow \infty$ due to  the second law of thermodynamics \cite{cover2012elements}.  The MCMC teaching of the generator  in the cooperative learning algorithm alternates between Markov transition from $q_{\alpha^{(t)}}$ to $p^{(t+1)}$,  and projection from $p^{(t+1)}$ to $q_{\alpha^{(t+1)}}$, as illustrated by Figure \ref{fig:LP}. The reduction of the Kullback-Leibler divergence  in (\ref{eq:mono}) and the projection in the MCMC teaching   are consistent with the learning objective of reducing $\KL(q_\alpha|p_{\theta^{(t)}})$ in \cite{Bengio2016}. But the Monte Carlo implementation of ${\cal M}$ in our work avoids the need to approximate the intractable entropy term. As to the updating of the descriptor model, \cite{Bengio2016} uses the initial synthesized examples generated by the generator model, while our method uses the revised synthesized examples obtained by finite-step MCMC towards the descriptor model, which are closer to the fair samples from the descriptor model. 

{\bf Special case (2)} as noise-injected back-propagation. The special case (2) with one step Langevin revision in the previous subsection amounts to a noise-injected back-propagation for minimizing ${\rm KL}(q_\alpha|p_\theta)$ with respect to $\alpha$. Specifically, consider gradient ascent on $\E[f(g(X; \alpha); \theta)]$ with respect to $\alpha$, where the expectation is with respect to $X \sim {\rm N}(0, I_d)$. The chain rule computation involves $\partial f/\partial Y \times \partial Y/\partial \alpha$, with $Y = g(X; \alpha)$. Special case (2) amounts to adding noise to $\partial f/\partial Y$ according to the Langevin dynamics, and then back-propagate to $\alpha$. The added noise increases the entropy of $q_\alpha$. 

\begin{figure}
	\centering
	\setlength{\fboxrule}{1pt}
	\setlength{\fboxsep}{0cm}	
	\includegraphics[width=.23\linewidth, height=.23\linewidth]{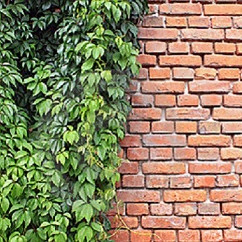}
	\includegraphics[width=.23\linewidth, height=.23\linewidth]{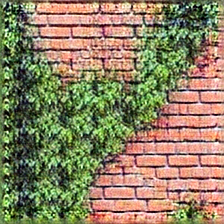}	
	\includegraphics[width=.23\linewidth, height=.23\linewidth]{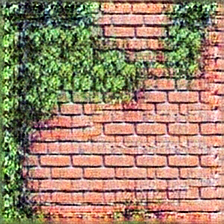}	
	\includegraphics[width=.23\linewidth, height=.23\linewidth]{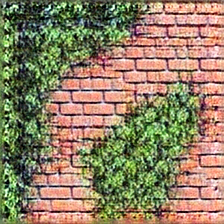} \\ \vspace{0.6mm}	
	\includegraphics[width=.23\linewidth, height=.23\linewidth]{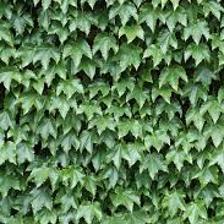}
	\includegraphics[width=.23\linewidth, height=.23\linewidth]{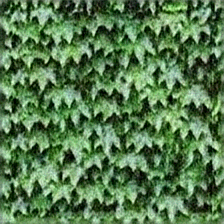}	
	\includegraphics[width=.23\linewidth, height=.23\linewidth]{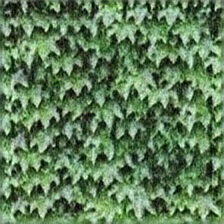}	
	\includegraphics[width=.23\linewidth, height=.23\linewidth]{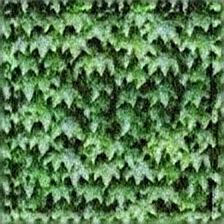} \\ \vspace{0.6mm}
	\includegraphics[width=.23\linewidth, height=.23\linewidth]{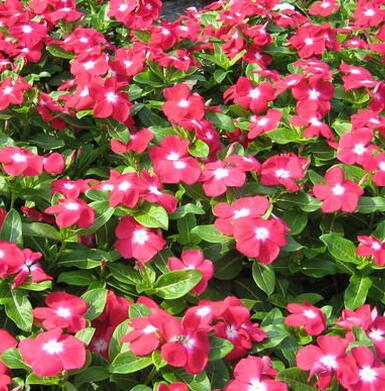}
	\includegraphics[width=.23\linewidth, height=.23\linewidth]{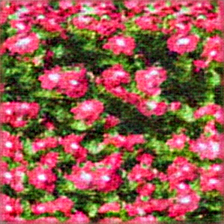}	
	\includegraphics[width=.23\linewidth, height=.23\linewidth]{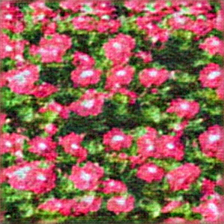}	
	\includegraphics[width=.23\linewidth, height=.23\linewidth]{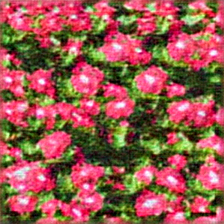} \\ \vspace{0.6mm}
	\includegraphics[width=.23\linewidth, height=.23\linewidth]{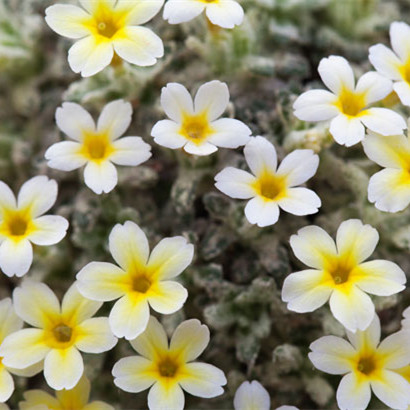}
	\includegraphics[width=.23\linewidth, height=.23\linewidth]{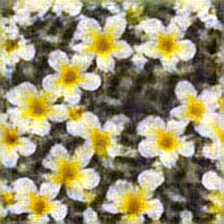}	
	\includegraphics[width=.23\linewidth, height=.23\linewidth]{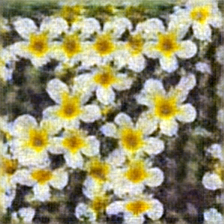}	
	\includegraphics[width=.23\linewidth, height=.23\linewidth]{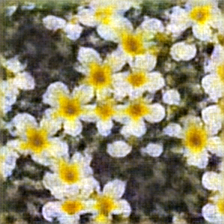} 	 \\ \vspace{0.6mm}

	\includegraphics[width=.23\linewidth, height=.23\linewidth]{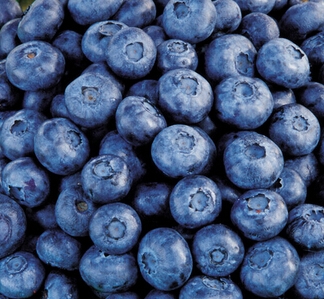}
	\includegraphics[width=.23\linewidth, height=.23\linewidth]{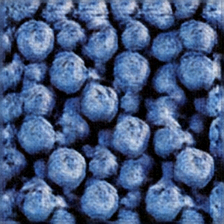}	
	\includegraphics[width=.23\linewidth, height=.23\linewidth]{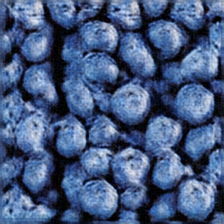}	
	\includegraphics[width=.23\linewidth, height=.23\linewidth]{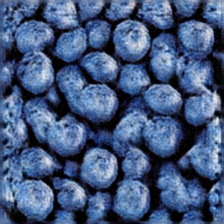} 	\\ \vspace{0.6mm}
    	\includegraphics[width=.23\linewidth, height=.23\linewidth]{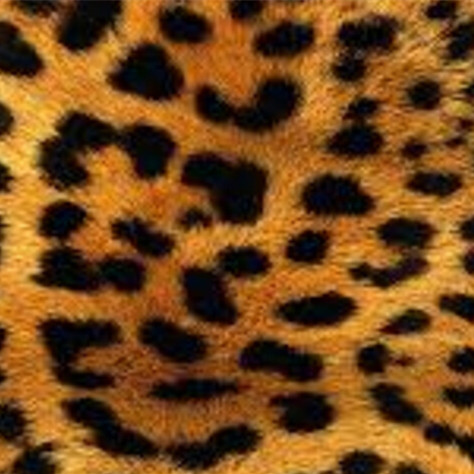}
	\includegraphics[width=.23\linewidth, height=.23\linewidth]{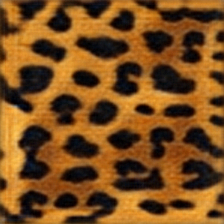}	
	\includegraphics[width=.23\linewidth, height=.23\linewidth]{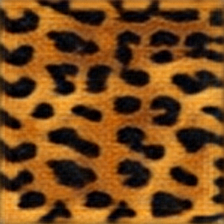}	
	\includegraphics[width=.23\linewidth, height=.23\linewidth]{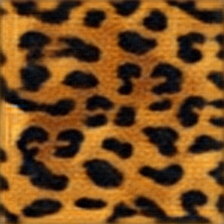} 	\\ \vspace{0.6mm}

	\includegraphics[width=.23\linewidth, height=.23\linewidth]{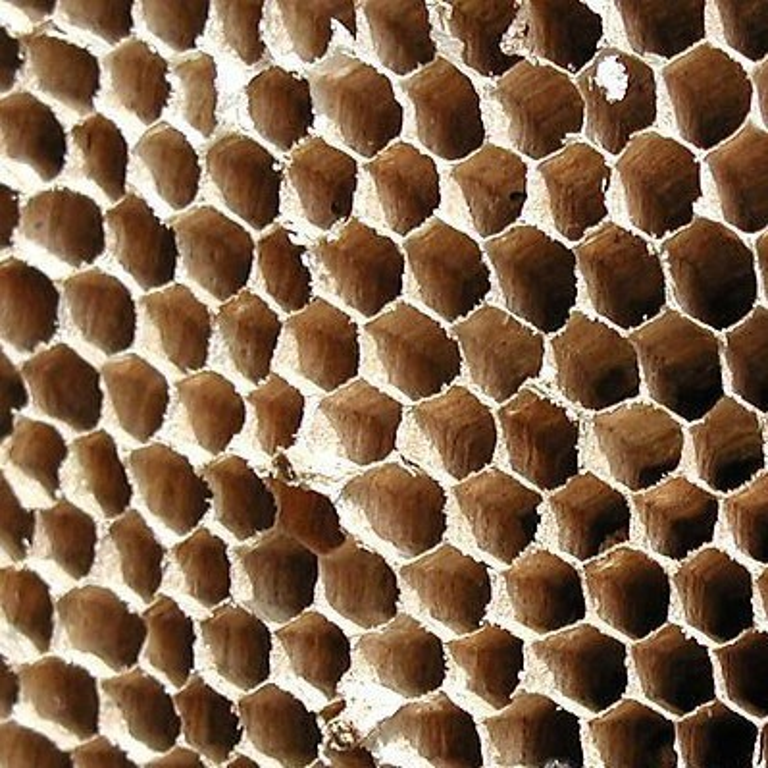}
	\includegraphics[width=.23\linewidth, height=.23\linewidth]{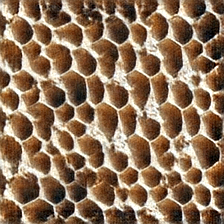}	
	\includegraphics[width=.23\linewidth, height=.23\linewidth]{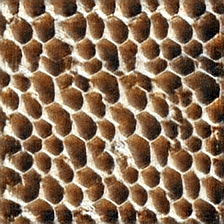}	
	\includegraphics[width=.23\linewidth, height=.23\linewidth]{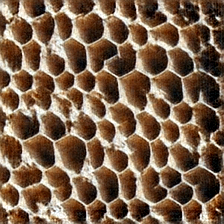} 	\\ \vspace{0.6mm}
	
	\includegraphics[width=.23\linewidth, height=.23\linewidth]{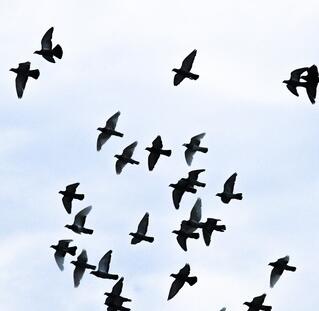}
	\includegraphics[width=.23\linewidth, height=.23\linewidth]{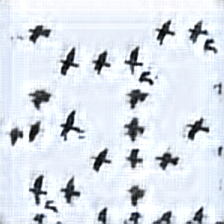}	
	\includegraphics[width=.23\linewidth, height=.23\linewidth]{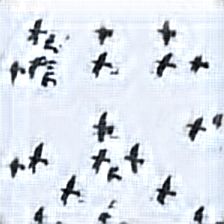}	
	\includegraphics[width=.23\linewidth, height=.23\linewidth]{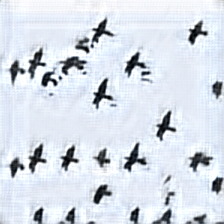} 			
	\caption{Generating texture patterns. Each row displays one texture experiment, where the first image is the training image, and the rest are 3 of the images generated by the CoopNets algorithm. The observed and synthesized images are of size 224 $\times$ 224 pixels. }
	\label{fig:texture}
\end{figure}

{\bf Special case (1)} as variational learning of MCMC teaching. For MCMC teaching, the right diagram in (\ref{eq:m}) leads to the update $\alpha^{(t+1)} = \arg\min_\alpha \KL(\M_\theta q_{\alpha^{(t)}}|q_\alpha)$, where $\{\tilde{Y}_i\} \sim \M_\theta q_{\alpha^{(t)}}$ serve as the training data, and the latent vector $X_i$ is inferred by Langevin dynamics initialized from $\hat{X}_i$. The left diagram in (\ref{eq:m}), i.e., the special case (1) in the previous subsection, can be viewed as a simplified approximation to the right diagram by fixing the latent factors at $\hat{X_i}$. It minimizes a variational upper bound  
\begin{eqnarray}
\KL(\M_\theta q_{\alpha^{(t)}}|q_\alpha) + \KL(q(\hat{X}_i | \tilde{Y}_i, \alpha^{(t)})|q({X}_i|\tilde{Y}_i, \alpha)).
\end{eqnarray}
 Our experiments suggest that the variational learning step in the left diagram works as well as the maximum likelihood learning step in the right diagram.

\section{Experiments}
We scale the training data to the range of the tanh activation function $[-1, 1]$. All learning parameters are initialized from a zero-centered Normal distribution with standard deviation 0.001. For the descriptor network, we adopt the structure of  \cite{XieLuICML}, where the bottom-up network consists of multiple layers of convolution by linear filtering, ReLU non-linearity, and down-sampling. We adopt the structure of the generator network of  \cite{radford2015unsupervised, Alexey2015}, where the top-down network consists of multiple layers of deconvolution by linear superposition with up-sampling, batch normalization \cite{ioffe2015batch}, and ReLU non-linearity, with tanh non-linearity at the bottom-layer \cite{radford2015unsupervised} to make the signals fall within $[-1, 1]$.    In our experiments, we set $l_{q} = 0$ and infer $X_i = \hX_i$,  i.e., we follow the left diagram in (\ref{eq:m}).  We have also experimented with $l_{q} >0$, i.e., the right diagram, but did not observe significant improvement.  

\subsection{Experiment 1: Generating texture patterns} 

We conduct experiments on generating texture patterns. We learn a separate model from each texture image. The training images are collected from the Internet, and then resized to 224 $\times$ 224 pixels. The synthesized images are of  the same size as the training images. We use a 3-layer descriptor network, where the first layer has 100 15 $\times$ 15 filters with sub-sampling rate of 3 pixels, the second layer has 70 9 $\times$ 9 filters with sub-sampling of 1, and the third layer has 30 7 $\times$ 7 filters with sub-sampling of 1. We set the standard deviation of the reference distribution of the descriptor network to be $s = 0.012$. We use $l_p=20$ or 30 steps of Langevin revision dynamics within each learning iteration, and the Langevin step size is set at 0.003. The learning rate is 0.01. Starting from 7 $\times$ 7 latent factors, the generator network has 5 layers of deconvolution with 5 $\times$ 5 kernels (basis functions), with an up-sampling factor of 2 at each layer (i.e., the basis functions are 2 pixels apart). The standard deviation of the noise vector is $\sigma =  0.3$. The learning rate is $10^{-6}$.  %The number of generator learning steps is 1 at each cooperative learning iteration.     
We run $10^4$ cooperative learning iterations to train the models. 

Figure \ref{fig:texture} displays the results of generating texture patterns. The synthesis results of the CoopNets algorithm shown in this paper are those generated by the descriptor (i.e. $\tilde{Y_i}$) unless otherwise specified. For each category, the first image is the training image, and the rest are 3 of the images generated by the learning algorithm. We run $\tn = 6$  parallel chains for the first example, where images from 3 of them are presented. We run a single chain for the rest of the examples, where the synthesized images are generated at different iterations.  Even though we run a single chain, it is as if we run an infinite number of chains, because in each iteration, we run Langevin revision dynamics from a new image sampled from the generator. 

\begin{figure}
\centering
\includegraphics[width=.1\linewidth, height=.1\linewidth]{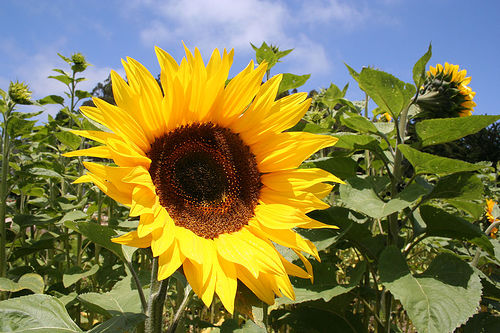}
\includegraphics[width=.1\linewidth,height=.1\linewidth]{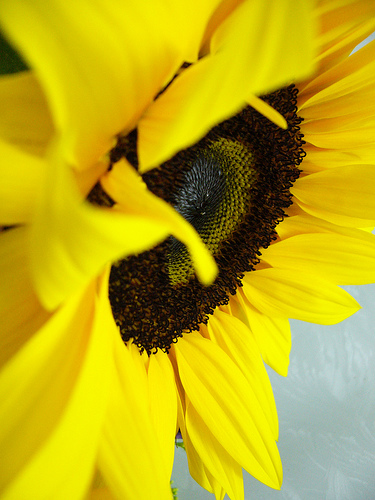}
\includegraphics[width=.1\linewidth, height=.1\linewidth]{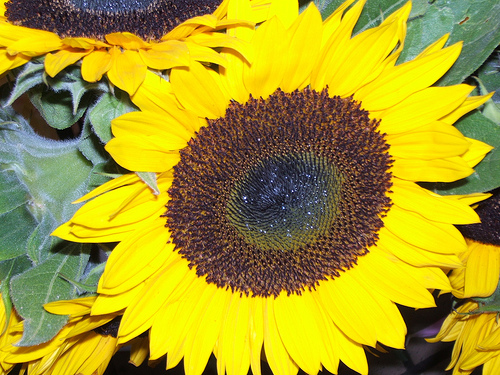} \hspace{1mm}
\includegraphics[width=.1\linewidth]{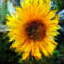}
\includegraphics[width=.1\linewidth]{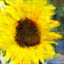}
\includegraphics[width=.1\linewidth]{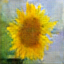}
\includegraphics[width=.1\linewidth]{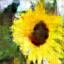} 
\includegraphics[width=.1\linewidth]{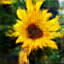}  
\includegraphics[width=.1\linewidth]{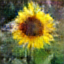}   \\ \vspace{1mm}
\includegraphics[width=.1\linewidth, height=.1\linewidth]{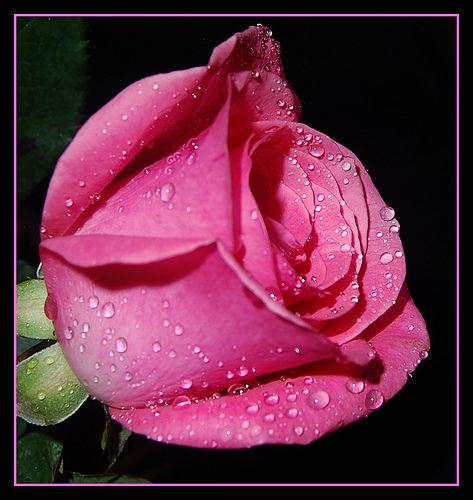}
\includegraphics[width=.1\linewidth,height=.1\linewidth]{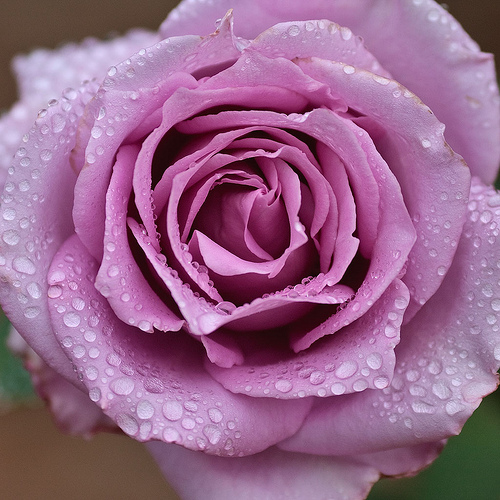}
\includegraphics[width=.1\linewidth, height=.1\linewidth]{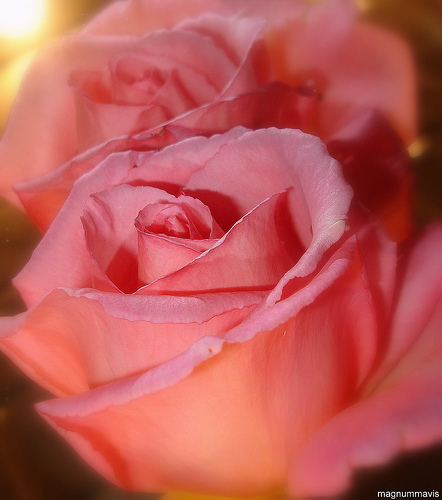} \hspace{1mm}
\includegraphics[width=.1\linewidth]{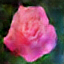}
\includegraphics[width=.1\linewidth]{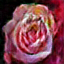}
\includegraphics[width=.1\linewidth]{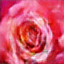}
\includegraphics[width=.1\linewidth]{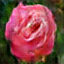} 
\includegraphics[width=.1\linewidth]{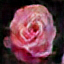}  
\includegraphics[width=.1\linewidth]{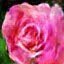}   \\ \vspace{1mm}

\includegraphics[width=.1\linewidth, height=.1\linewidth]{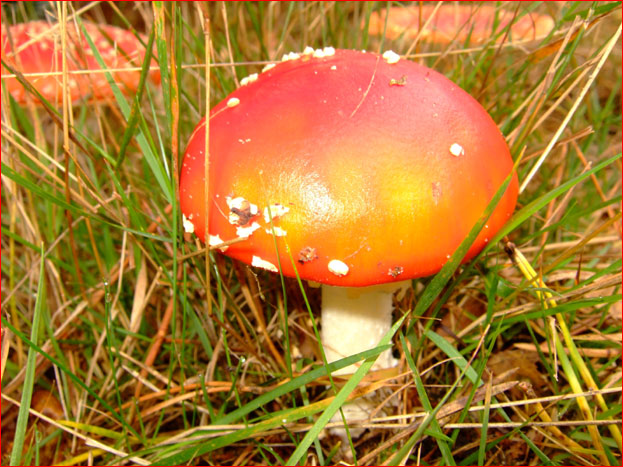}
\includegraphics[width=.1\linewidth,height=.1\linewidth]{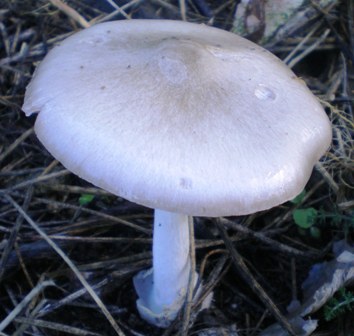}
\includegraphics[width=.1\linewidth, height=.1\linewidth]{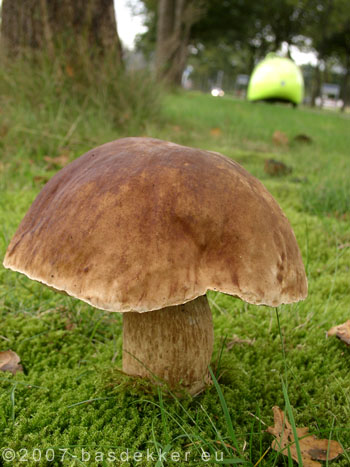} \hspace{1mm}
\includegraphics[width=.1\linewidth]{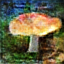}
\includegraphics[width=.1\linewidth]{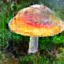}
\includegraphics[width=.1\linewidth]{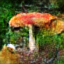}
\includegraphics[width=.1\linewidth]{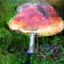} 
\includegraphics[width=.1\linewidth]{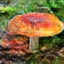}  
\includegraphics[width=.1\linewidth]{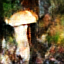} \\ \vspace{1mm}

\includegraphics[width=.1\linewidth, height=.1\linewidth]{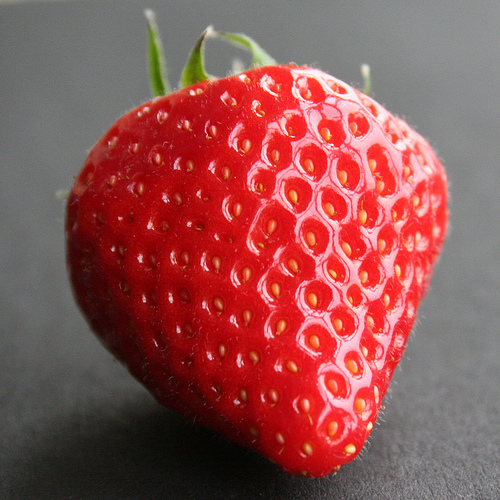}
\includegraphics[width=.1\linewidth,height=.1\linewidth]{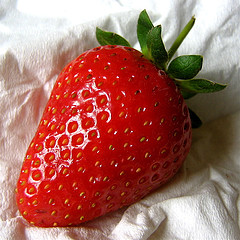}
\includegraphics[width=.1\linewidth, height=.1\linewidth]{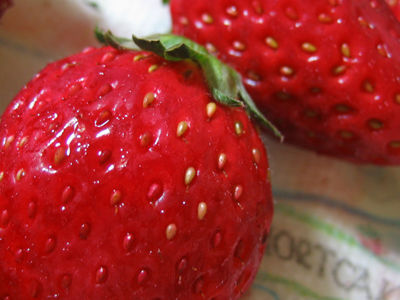} \hspace{1mm}
\includegraphics[width=.1\linewidth]{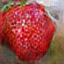}
\includegraphics[width=.1\linewidth]{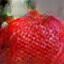}
\includegraphics[width=.1\linewidth]{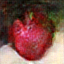}
\includegraphics[width=.1\linewidth]{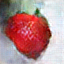} 
\includegraphics[width=.1\linewidth]{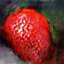}  
\includegraphics[width=.1\linewidth]{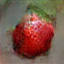} \\ \vspace{1mm}

\includegraphics[width=.1\linewidth, height=.1\linewidth]{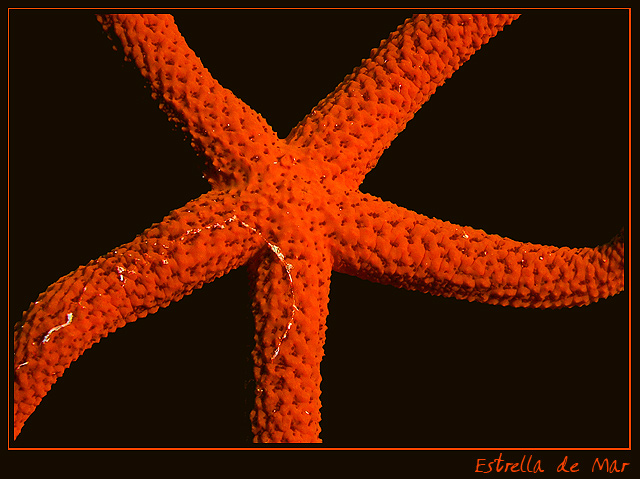}
\includegraphics[width=.1\linewidth,height=.1\linewidth]{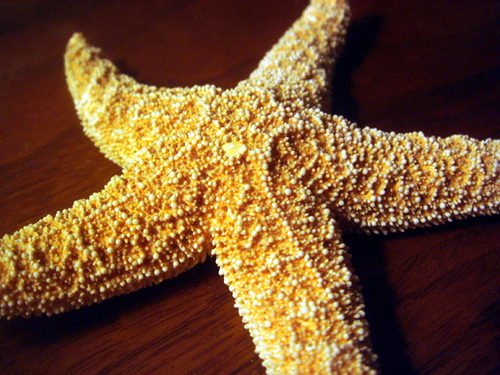}
\includegraphics[width=.1\linewidth, height=.1\linewidth]{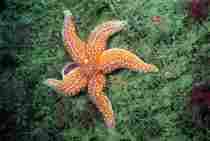} \hspace{1mm}
\includegraphics[width=.1\linewidth]{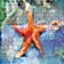}
\includegraphics[width=.1\linewidth]{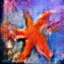}
\includegraphics[width=.1\linewidth]{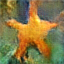}
\includegraphics[width=.1\linewidth]{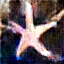} 
\includegraphics[width=.1\linewidth]{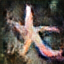}  
\includegraphics[width=.1\linewidth]{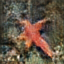} \\ \vspace{1mm}

\includegraphics[width=.1\linewidth, height=.1\linewidth]{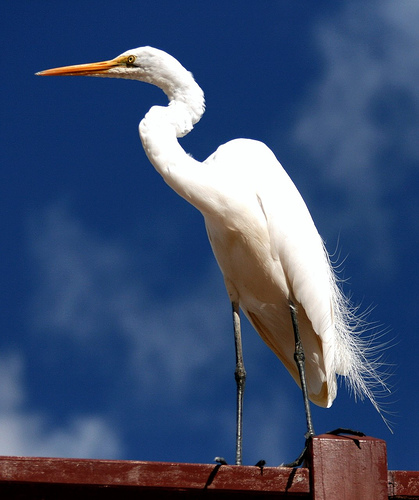}
\includegraphics[width=.1\linewidth,height=.1\linewidth]{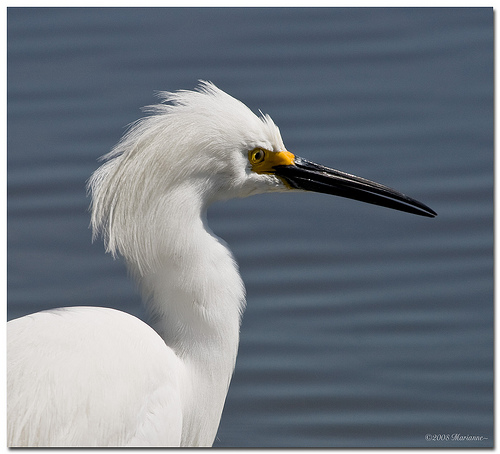}
\includegraphics[width=.1\linewidth, height=.1\linewidth]{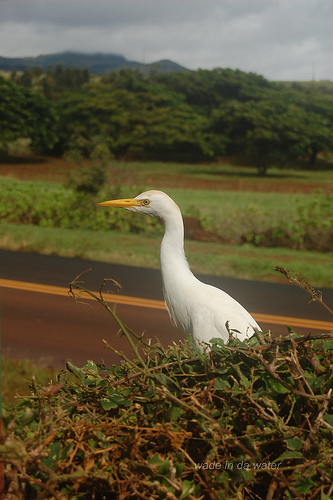} \hspace{1mm}
\includegraphics[width=.1\linewidth]{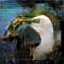}
\includegraphics[width=.1\linewidth]{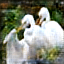}
\includegraphics[width=.1\linewidth]{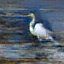}
\includegraphics[width=.1\linewidth]{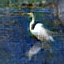} 
\includegraphics[width=.1\linewidth]{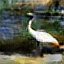}  
\includegraphics[width=.1\linewidth]{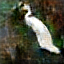} \\ \vspace{1mm}

\includegraphics[width=.1\linewidth, height=.1\linewidth]{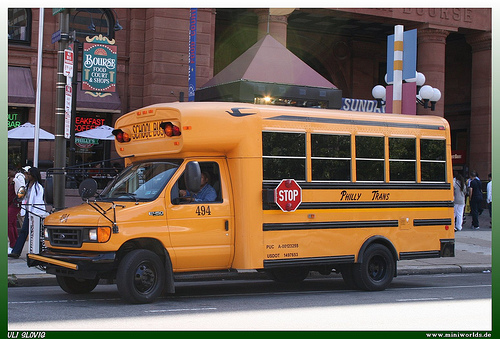}
\includegraphics[width=.1\linewidth,height=.1\linewidth]{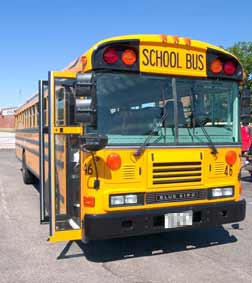}
\includegraphics[width=.1\linewidth, height=.1\linewidth]{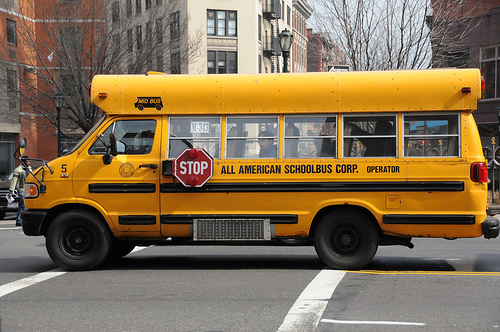} \hspace{1mm}
\includegraphics[width=.1\linewidth]{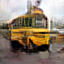}
\includegraphics[width=.1\linewidth]{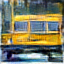}
\includegraphics[width=.1\linewidth]{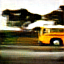}
\includegraphics[width=.1\linewidth]{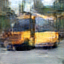} 
\includegraphics[width=.1\linewidth]{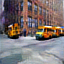}  
\includegraphics[width=.1\linewidth]{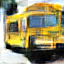} 
\caption{Generating object patterns. Each row displays one object experiment, where the first 3 images are 3 of the training images, and the rest are 6 of the synthesized images generated by the CoopNets algorithm. The observed and synthesized images are of size $64 \times 64 $ pixels. }	
\label{fig:object}
\end{figure}

\subsection{Experiment 2: Generating object patterns} 
\label{subsec:Exp2}
We study generating object patterns via the CoopNets algorithm. We use object categories selected from Imagenet-1k dataset \cite{deng2009imagenet}. Each category contains roughly 1,200+ training images, each of which is resized to $64 \times 64$ pixels.

We adopt a 4-layer descriptor network, where the first layer has 64 $5\times5$ filters with sub-sampling of $2$ pixels, the second layer has 128 $3\times3$ filters with sub-sampling of $2$, the third layer has 256 $3\times3$ filters with sub-sampling of $1$, and the final layer is a fully connected layer with 100 channels as output. We set the number of Langevin dynamics steps in each learning iteration to $l_p=10$ and the step size to 0.002. The standard deviation for reference distribution is $s=0.016$. The learning rate is 0.007.

Starting from a 100-dimensional latent factor, the generator network has one fully connected layer with $4 \times 4$ kernels under the latent factors, which is followed by 4 layers of deconvolution with kernels of size $5 \times 5$, and up-sampling factor of 2. The numbers of channels from top layer to bottom layer are 512, 256, 128, 64, and 3 respectively. The output size is $64 \times 64$. The learning rate for updating the parameters of generator is 0.0001. We set $\sigma=0.3$ for the standard deviation of the noise vector $\epsilon$.

The CoopNets algorithm is trained by Adam optimizer \cite{kingma2014adam} with a mini-batch size of 100. We run $\tilde{n}=144$ parallel chains for synthesis. The number of cooperative learning iterations is $1,000$. After learning the models, we synthesize images using the learned models. As in the CoopNets algorithm, we sample from the learned descriptor network by running 10 to 50 steps of Langevin dynamics initialized from the examples generated by the learned generator network. 

For qualitative experiment, we learn a separate model for each of 7 selected object categories (i.e., sunflower, rose, mushroom, strawberry, sea star, egret, and school bus). Figure \ref{fig:object} shows the synthesis results. Each row displays an experiment, where the first 3 images are 3 typical examples of the training images, and the rest are 6 of the synthesized images generated by the CoopNets algorithm.

\begin{figure*}
	\centering	
	\includegraphics[height=.21\linewidth, width=.3\linewidth]{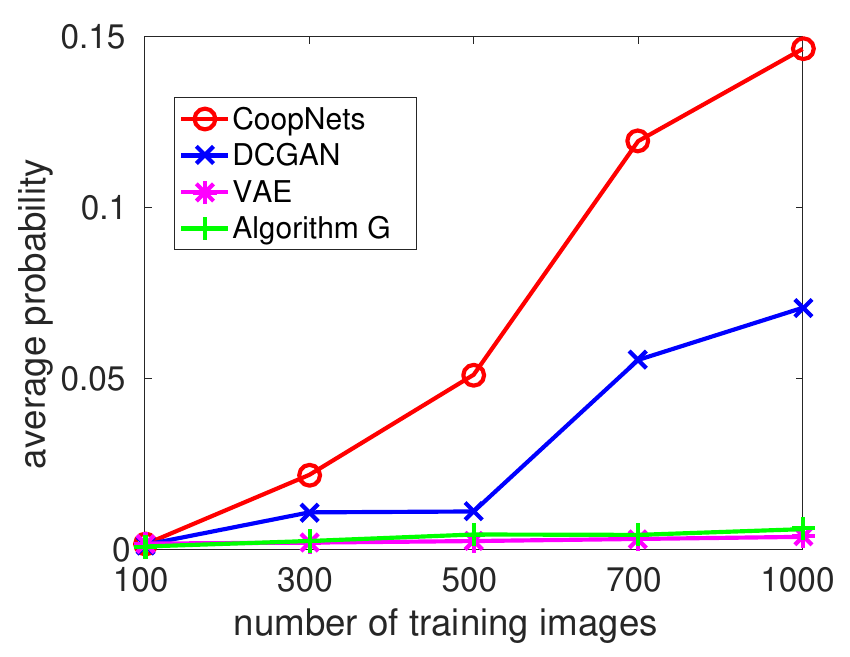}		
	\includegraphics[height=.21\linewidth, width=.3\linewidth]{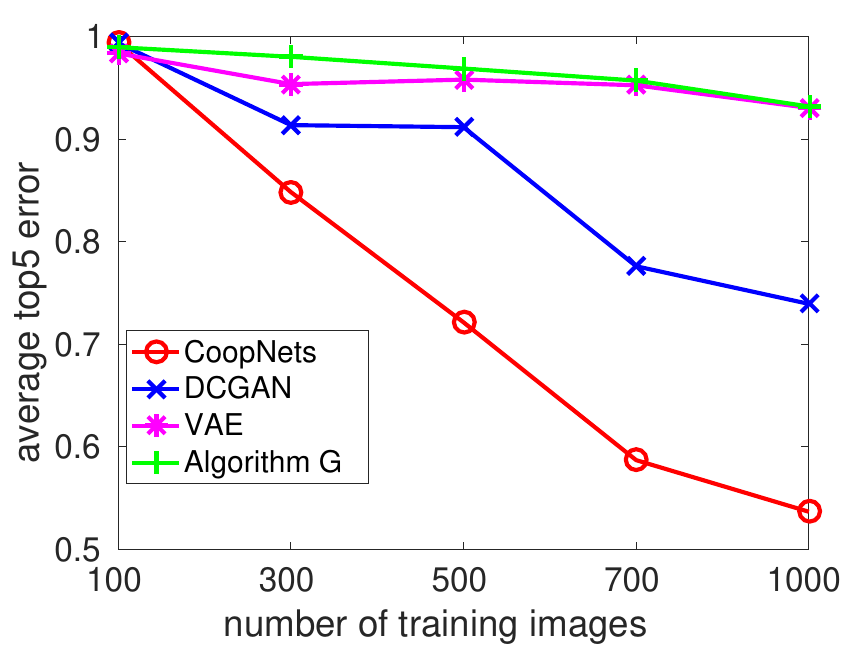}		
	\includegraphics[height=.21\linewidth, width=.3\linewidth]{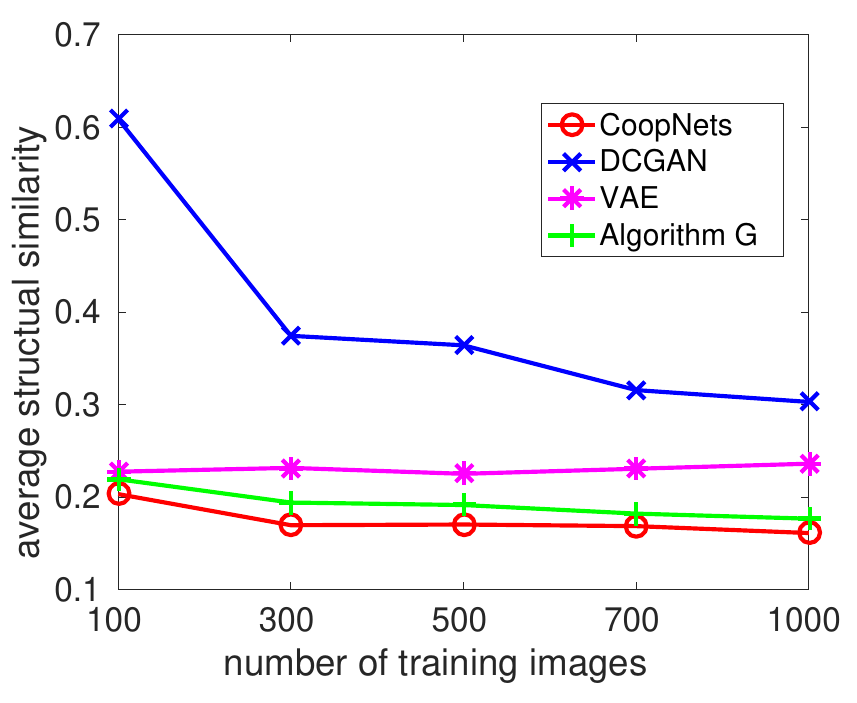}\\		
	\caption{Left: Average softmax class probability on single Imagenet-1k category versus the number of training images. Middle: Top 5 classification error. Right: Average pairwise structural similarity.}
\label{fig:curves}
\end{figure*}

For quantitative experiment, we use five object categories (i,e., lemon, lifeboat, strawberry, school bus and zebra), and train models on different numbers of randomly sampled training images for each category. The synthesis quality is quantitatively evaluated using three criteria: (1) average softmax class probability that Inception network \cite{szegedy2016rethinking} assigns to the synthesized images for the underlying category. (2) top-5 classification error by Inception network, i.e., the probability that the underlying category does not belong to the categories with the top 5 softmax probabilities. (3) Average pairwise structural similarity \cite{wang2004image} between two randomly selected synthesized images. While (1) and (2) examine the realism of the synthesized images, (3) examines the variabilities of the synthesized images. The lack of variabilities may be caused by mode collapsing.  

Figure \ref{fig:curves} displays the average results over the 5 categories versus the number of training examples. It can be seen that CoopNets generates images with higher softmax class probabilities, lower classification errors, and higher variabilities than DCGAN \cite{radford2015unsupervised}, VAE \cite{KingmaCoRR13} and separate training by Algorithm G. The advantage can be due to the fact that both models in CoopNets are learned generatively by maximum likelihood. Our experiments suggest that the CoopNets learning method is stable and does not encounter mode collapsing issue.

\subsection{Experiment 3: Generating scene patterns} 

\begin{figure*}
\centering
\includegraphics[width=.22\linewidth]{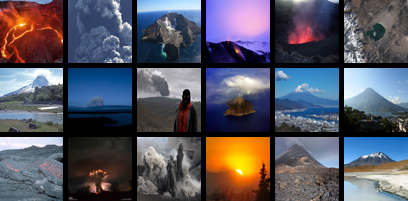}	\hspace{1.6mm}	
\includegraphics[width=.22\linewidth]{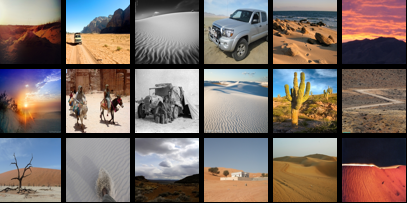} \hspace{1.6mm}	
\includegraphics[width=.22\linewidth]{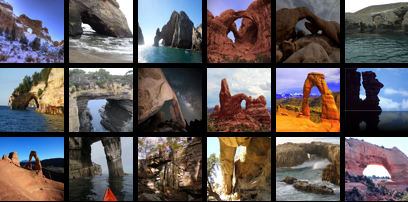} \hspace{1.6mm}	
\includegraphics[width=.22\linewidth]{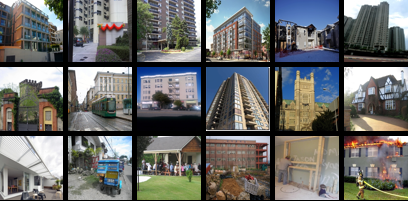}\\
 observed images  \hspace{19mm} observed images \hspace{19mm} observed images \hspace{19mm}  observed images  \\ \vspace{2mm}
\includegraphics[width=.22\linewidth]{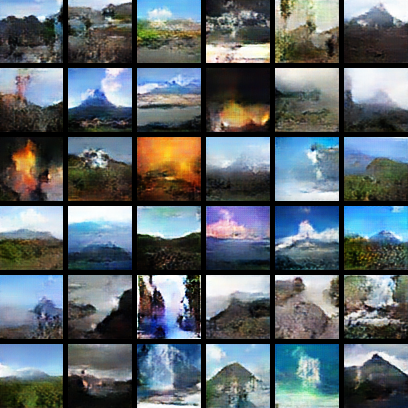} \hspace{1.6mm}
\includegraphics[width=.22\linewidth]{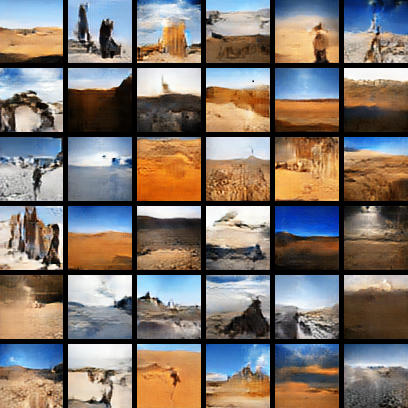} \hspace{1.6mm}
\includegraphics[width=.22\linewidth]{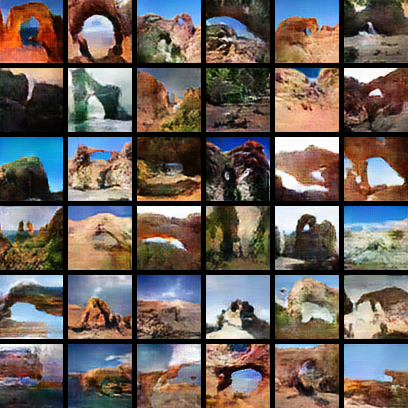} \hspace{1.6mm}
\includegraphics[width=.22\linewidth]{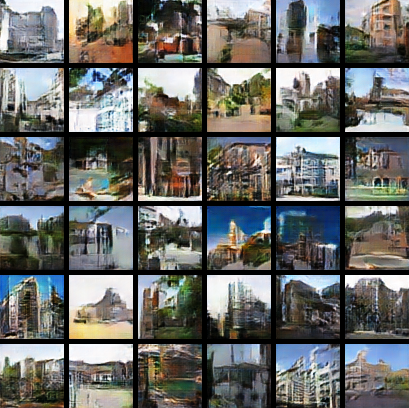}
 synthesized images  \hspace{15mm} synthesized images \hspace{15mm} synthesized images \hspace{15mm}  synthesized images  \\
\hspace{7mm}  (a) volcano \hspace{27mm} (b) desert \hspace{28mm} (c) rock \hspace{21mm}  (d) apartment building 
\caption{Generating scene patterns. Both observed and synthesized scene images are shown for each category. The image size is $64 \times 64$ pixels. The categories are from MIT places205 dataset. (a) volcano. (b) desert. (c) rock. (d) apartment building.}	
\label{fig:ImageNet}
\end{figure*}

We conduct experiments on synthesizing scene patterns. We learn a separate model for each of the 4 scene categories (i.e., volcano, desert, rock, and apartment building) selected from MIT place205 dataset  \cite{zhou2014learning}. Each category has 10,000+ training images. The images are resized to 64 $\times$ 64 pixels. We adopt the same architecture of CoopNets as the one for object patterns in Section \ref{subsec:Exp2}. Figure \ref{fig:ImageNet} displays 36 synthesized scene images generated by the learned model, along with 18 randomly sampled training images for each experiment.

\begin{figure}[t]
\centering
\subfigure[observed images]{
\includegraphics[width=.93\linewidth]{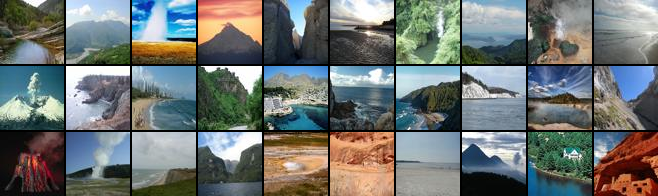}}
\subfigure[synthesized images]{
\includegraphics[width=.93\linewidth]{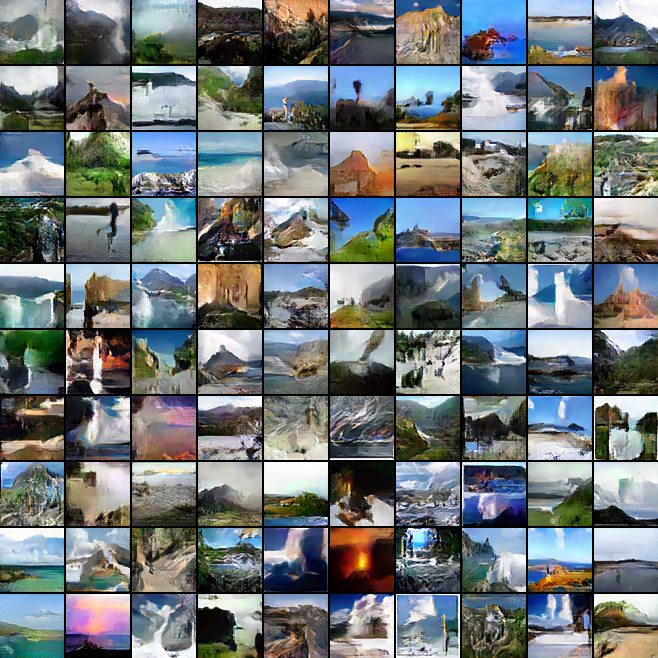}}
\caption{Generating scene patterns. (a) observed images randomly selected from 10 Imagenet-1k scene categories. (b) synthesized images generated by CoopNets learned from 10 Imagenet-1k scene categories. The training set consists of 1,100 images randomly sampled from each category. The observed and synthesized images are of size 64 $\times$ 64 pixels.}	
\label{fig:syn_10category1}
\end{figure}

We then learn from mixed images that are randomly sampled from 10 different scene categories (i.e., alp, cliff drop, cliff dwelling, geyser, lakeside, promontory, sandbar, seashore, valley, and volcano) selected from Imagenet-1k  dataset. We conduct 7 runs. The numbers of images sampled from each category are 50, 100, 300, 500, 700, 900, and 1,100 respectively in these 7 runs. Figure \ref{fig:syn_10category1} displays the observed examples randomly sampled from the training set, and the synthesized examples generated by the CoopNets, where the number of training images from each category is 1,100. The synthesized examples are randomly sampled from the learned models without cheery picking. We evaluate the synthesis quality by the Inception score \cite{salimans2016improved}. Table \ref{table1} displays the Inception scores of the CoopNets, DCGAN, EBGAN \cite{zhao2016energy}, Wasserstein GAN \cite{arjovsky2017wasserstein}, InfoGAN \cite{chen2016infogan}, VAE, the method of \cite{Bengio2016}, and separate training by Algorithm G and Algorithm D. For Algorithm D, we initialize the synthesized examples from the observed examples, so it is persistent contrastive divergence \cite{tieleman2008training}. 

\begin{figure}
	\centering	
	\includegraphics[width=0.9\linewidth]{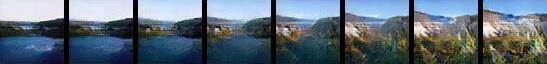} \\ \vspace{1mm}
	\includegraphics[width=0.9\linewidth]{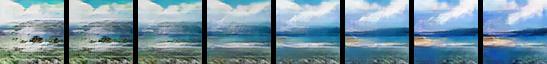} \\ \vspace{1mm}	 
	\includegraphics[width=0.9\linewidth]{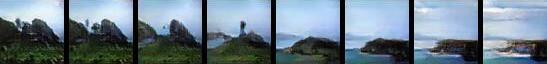} \\ \vspace{1mm}
	\includegraphics[width=0.9\linewidth]{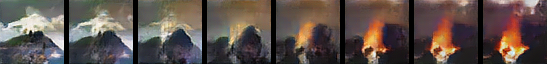}	\\ \vspace{1mm}
	\includegraphics[width=0.9\linewidth]{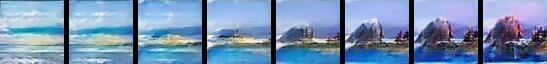}	\\ %\vspace{1mm}
	\caption{Interpolation between latent vectors of the scene images on the two ends.}	
	\label{fig:iterpolation}
\end{figure}

Figure \ref{fig:iterpolation} shows 5 examples of interpolation between two latent vectors of $X$. For each row, the images at the two ends are generated from $X$ vectors randomly sampled from ${\rm N}(0,I_d)$. Each image in the middle is obtained by first interpolating the $X$ vectors of the two end images, and then generating the image using the generator, followed by 10 steps of Langevin dynamics.  This experiment shows that we learn smooth generator model that traces the manifold of the data distribution. 

\begin{table*}
\centering
\caption{Inception scores of different methods on learning from 10 Imagenet-1k scene categories. $n$ is the number of training images randomly sampled from each category. }
\label{tablee}
{\footnotesize %\scriptsize %
\begin{tabular}{|c|c|c|c|c|c|c|c|}
\hline 
            & $n$ = 50            & $n$ = 100           & $n$ = 300           & $n$ = 500           & $n$ = 700 & $n$ = 900 & $n$ = 1100          \\ \hline \hline
DCGAN \cite{radford2015unsupervised}         & 2.26$\pm$.16 & 2.50$\pm$.15  & 3.16$\pm$.15  & 3.05$\pm$.12  & 3.13$\pm$.09 & 3.34$\pm$.05 & 3.47$\pm$.06 \\
EBGAN \cite{zhao2016energy} & 2.23$\pm$.17 & 2.40$\pm$.14 & 2.62$\pm$.08 & 2.46$\pm$.09 & 2.65$\pm$.04 & 2.64$\pm$.04 & 2.75$\pm$.08 \\
W-GAN \cite{arjovsky2017wasserstein} & 1.80$\pm$.09 & 2.19$\pm$.12 & 2.34$\pm$.06 & 2.62$\pm$.08 & 2.86$\pm$.10 & 2.88$\pm$.07 & 3.14$\pm$.06 \\
VAE \cite{KingmaCoRR13}       & 1.62$\pm$.09 & 1.63$\pm$.06 & 1.65$\pm$.05 & 1.73$\pm$.04 & 1.67$\pm$.03 & 1.72$\pm$.02 & 1.73$\pm$.02 \\ 
InfoGAN \cite{chen2016infogan} & 2.21$\pm$.04 &1.73$\pm$.01 &2.15$\pm$.03 & 2.42$\pm$.05 &2.47$\pm$.05 & 2.29$\pm$.03 & 2.08$\pm$.04 \\
Method of \cite{Bengio2016} & 2.44$\pm$.27 & 2.38$\pm$.13 & 2.42$\pm$.09 & 2.94$\pm$.11 & 3.02$\pm$.06 & 3.08$\pm$.08 & 3.15$\pm$.06 \\ 
Algorithm G \cite{HanLu2016} &  1.72$\pm$.07 & 1.94$\pm$.09 & 2.32$\pm$.09 & 2.40$\pm$.06 & 2.45$\pm$.05 & 2.54$\pm$.05 & 2.61$\pm$.06 \\
Persistent CD \cite{tieleman2008training} &  1.30$\pm$.08 & 1.94$\pm$.03 & 1.80$\pm$.02 & 1.53$\pm$.02 & 1.45$\pm$.04 & 1.35$\pm$.02 & 1.51$\pm$.02 \\
CoopNets (ours)    & {\bf 2.66$\pm$.13} & {\bf 3.04$\pm$.13}  & {\bf 3.41$\pm$.13}  & {\bf 3.48$\pm$.08} & {\bf 3.59$\pm$.11} & {\bf 3.65$\pm$.07} & {\bf 3.79$\pm$.15} \\ 
\hline
\end{tabular}
}
\label{table1}
\end{table*}

\subsection{Experiment 4: Generating handwritten digits}

\begin{figure}
\begin{center}
\includegraphics[height=.56\linewidth]{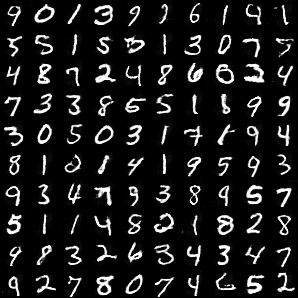}
\caption{Generating handwritten digits. The synthesized images are generated by the CoopNets algorithm that learns from MNIST dataset with 55,000 training images. The observed and synthesized images are of size 28 $\times$ 28 pixels.}
\label{fig:MNIST}
\end{center}
\end{figure}

\begin{figure}[h]
\begin{center}
\includegraphics[height=.55\linewidth]{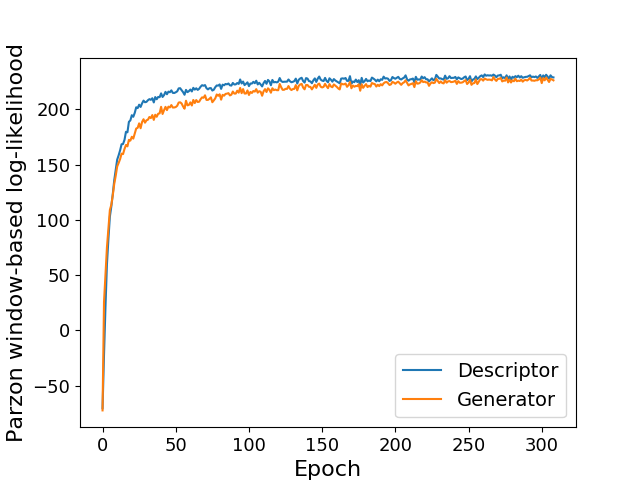}
\caption{Parzen window-based log-likelihood estimates of the descriptor network and the generator network in the CoopNets algorithm.}
\label{fig:MNIST_curve}
\end{center}
\end{figure}

We learn CoopNets from MNIST dataset \cite{lecun1998gradient} of handwritten digits. The training images are grey-scale with size of 28 $\times$ 28 pixels. We adopt a 4-layer descriptor network, where the numbers of filters at different layers are 64, 128, 256, and 100 from bottom to top. The filter size of each layer is $4 \times 4$, and the sub-sampling rate is 2. The final layer is a fully connected layer. Taking a 100-dimensional latent vector as input, the generator network consists of one fully connected layer and 3 deconvolutional layers with kernels of size $4 \times 4$ and up-sampling factor of 2. The numbers of channels from top layer to bottom layer are 512, 256, 128, and 1 respectively. The output size is 28 $\times$ 28 pixels. Except network architectures, we follow the same hyper-parameter setting as used in Section \ref{subsec:Exp2}. Figure \ref{fig:MNIST} displays some synthesized examples generated by the learned models after training. 

To quantitatively evaluate the learned model, we first synthesize 10,000 samples from the CoopNets learned on the training set of 55,000 examples. We then fit a Gaussian Parzen window to the synthesized samples, and estimate the log-likelihood of the testing set using the Parzen window distribution. The standard deviation of the Gaussian is obtained by cross validation on the validation set. This evaluation method was proposed by \cite{breuleux2011quickly} and has been used by \cite{bengio2014deep, bengio2013representation, goodfellow2014generative} for evaluating generative models with non-tractable likelihoods. Figure \ref{fig:MNIST_curve} displays the Parzen window-based log-likelihood estimates of the MNIST testing set for both the descriptor network and the generator network trained by the CoopNets algorithm with different numbers of training epochs. It can be seen that in the proposed cooperative training scheme, the descriptor network improves the generator network, and the generator network eventually gets very close to the descriptor network. We also compare our model against other baselines, e.g., DBN \cite{bengio2013representation}, Stacked CAE \cite{bengio2013representation}, Deep GSN \cite{bengio2014deep}, and GAN \cite{goodfellow2014generative} in Table \ref{tab:Parzon}, where both descriptor and generator networks outperform other baseline models.

\begin{table}
\centering
\caption{A comparison of Parzen window-based log-likelihood estimates for MNIST dataset. The mean log-likelihood of testing samples, with the standard error of the mean computed across examples, are reported. }
\label{tab:Parzon}
\begin{tabular}{|c|c|}
\hline
   Model   & Log-likelihood  \\
 \hline \hline
   DBN \cite{bengio2013representation}    &  138 $\pm$ 2.0  \\
   Stacked CAE \cite{bengio2013representation} & 121 $\pm$ 1.6 \\
   Deep GSN \cite{bengio2014deep}   & 214 $\pm$ 1.1\\
   GAN \cite{goodfellow2014generative} & 225 $\pm$ 2.0\\ \hline
   Generator in CoopNets (ours)  & 226 $\pm$ 2.1\\
   Descriptor in CoopNets (ours)  & \textbf{228 $\pm$ 2.1}\\
   \hline
\end{tabular}
\end{table}

\subsection{Experiment 5: Evaluation on large-scale benchmark datasets}

To demonstrate how the CoopNets scales with more complex data, we evaluate the model on three challenging large-scale benchmark datasets, i.e., LSUN bedrooms \cite{yu15lsun}, CelebA human faces \cite{liu2015deep}, and Cifar-10 objects \cite{krizhevsky2009learning} datasets. 

\begin{figure}
\centering
\subfigure[training images]{
\includegraphics[width=.95\linewidth]{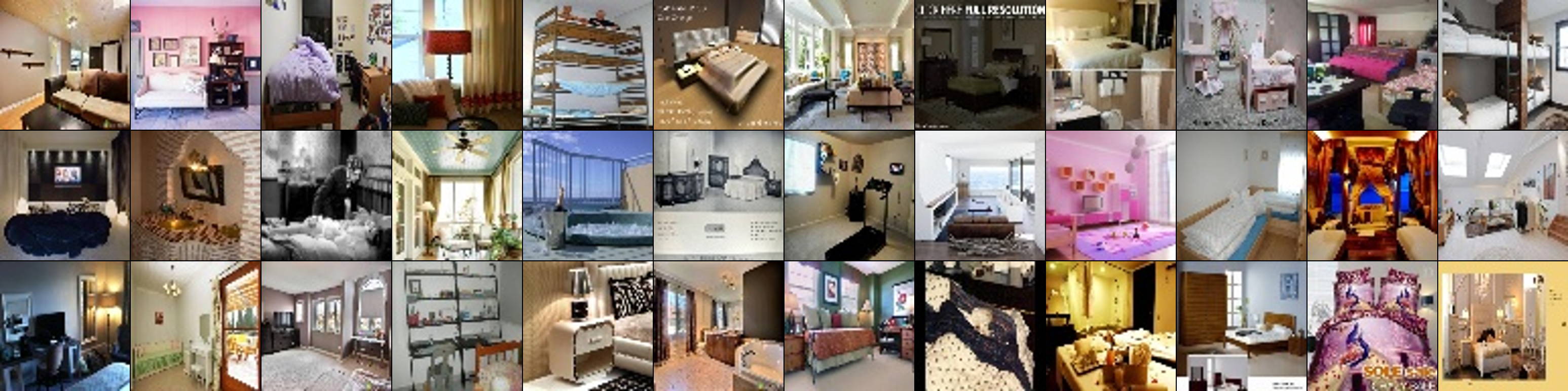}}
\subfigure[synthesized images]{
\includegraphics[width=.95\linewidth]{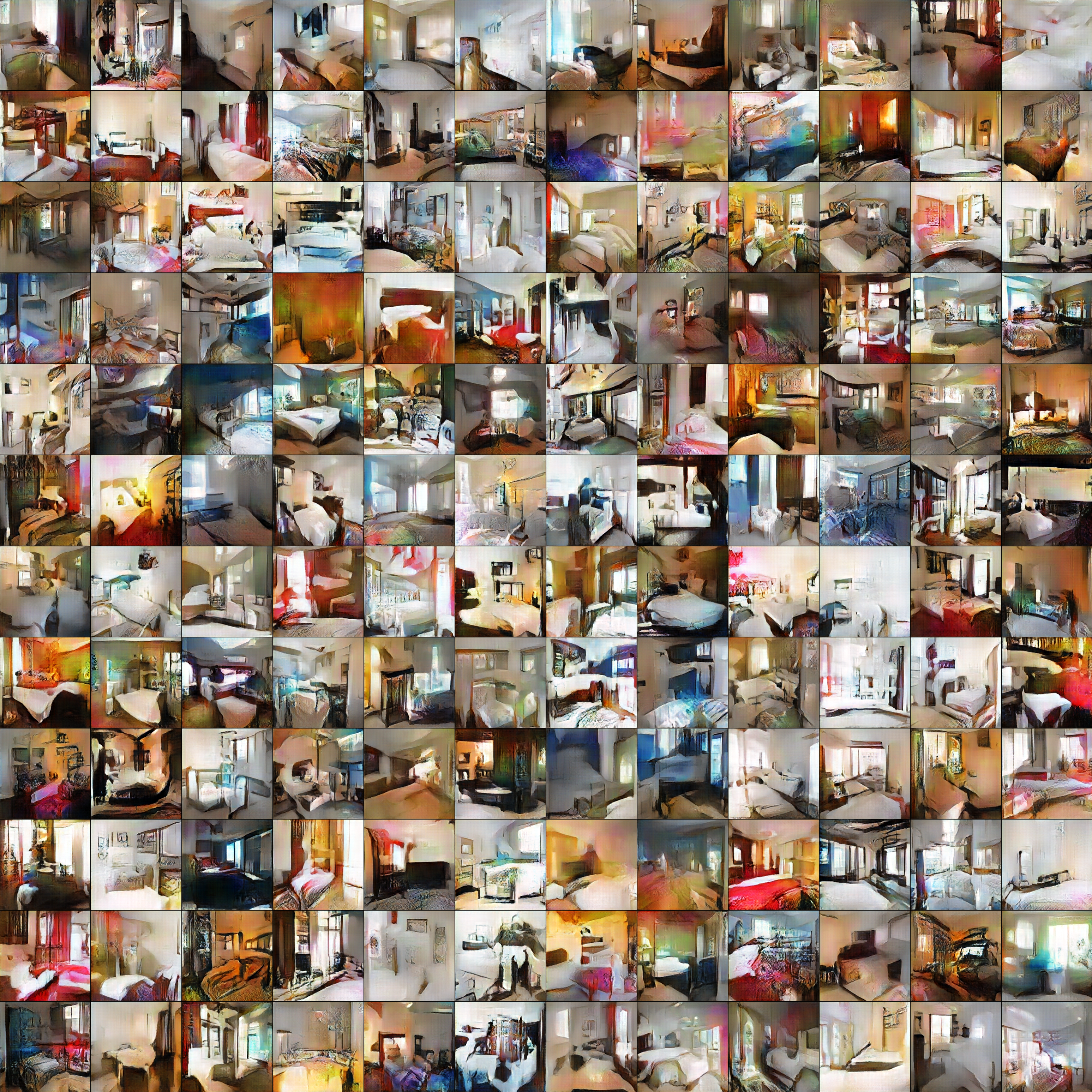}}
\caption{Generating bedroom images (256 $\times$ 256 pixels). The synthesized images are generated by the CoopNets algorithm that learns from LSUN dataset with 3033k training images.}
\label{fig:LSUN256}
\end{figure}

We first test the CoopNets on the LSUN bedrooms dataset containing 3,033k training images of 256 $\times$ 256 pixels. The network architectures are as follows. The descriptor network consists of 5 convolutional layers with numbers of channels $\{64, 128, 256, 512, 512\}$, filter sizes $\{5, 5, 5, 5, 3\}$, and sub-sampling factors $\{2, 2, 2, 2, 2\}$ at different layers (from bottom to top), and one fully connected layer with 10 filers. The generator network takes as input a 100-dimensional latent factor, and consists of 1 fully connected and 4 deconvolutional layers with numbers of channels $\{512, 256, 128, 64, 3\}$, kernels sizes $\{16, 5, 5, 5, 5\}$, and up-sampling factors $\{1, 2, 2, 2, 2\}$ at different layers (from top to bottom). Figure \ref{fig:LSUN256} displays the synthesized images.

\begin{figure}
\centering
\subfigure[training images]{
\includegraphics[width=.95\linewidth]{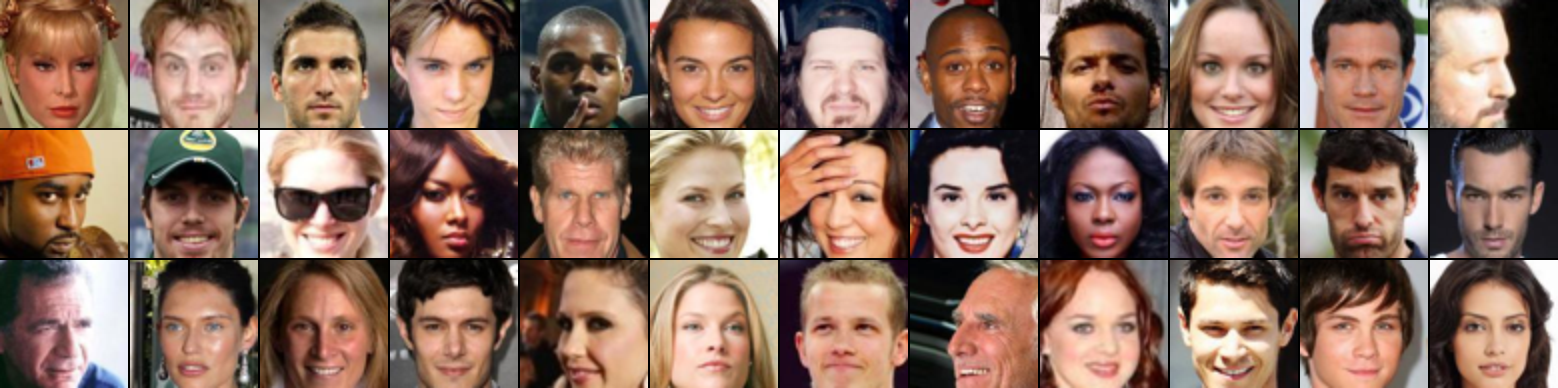}}
\subfigure[synthesized images]{
\includegraphics[width=.95\linewidth]{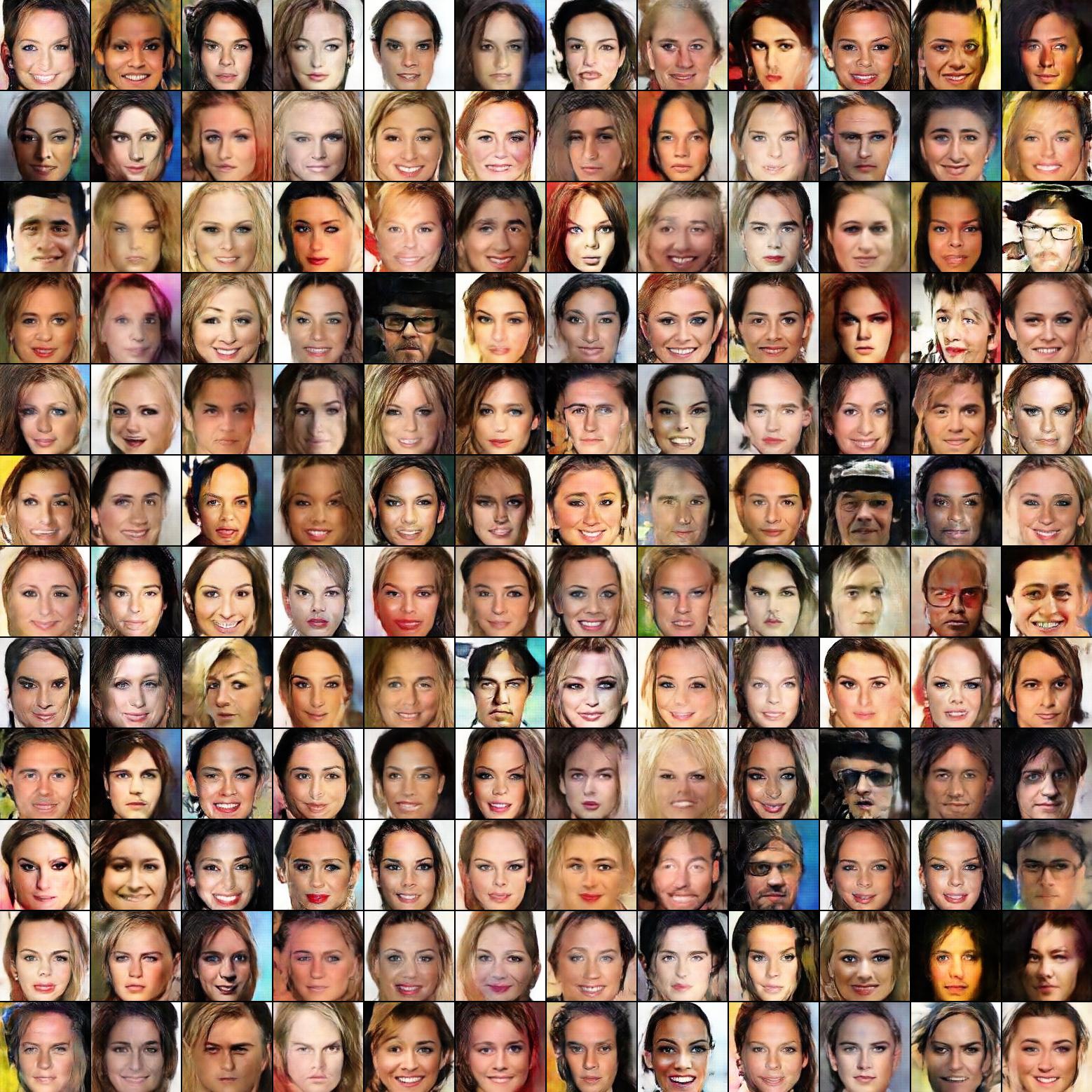}}
\caption{Generating human face images (128 $\times$ 128 pixels). The synthesized images are generated by the CoopNets algorithm that learns from CelebA dataset with 200k training images.}
\label{fig:celebA128}
\end{figure}

We then learn a model from the CelebA human faces dataset with 200k training images of 128 $\times$ 128 pixels. For this dataset, we adopt a descriptor network that has 3 convolutional layers with numbers of channels $\{64, 128, 256\}$, filter sizes $\{5, 3, 3\}$ and sub-sampling factors $\{2, 2, 1\}$ at different layers (from bottom to top), and one fully connected layer with 100 filers. Besides, we adopt a generator network that takes a 100-dimensional input and consists of one fully connected and 4 deconvolutional layers with numbers of channels $\{512, 256, 128, 64, 3\}$, kernels sizes $\{4, 5, 5, 5, 5\}$ and up-sampling factors $\{1, 2, 4, 2, 2\}$ at different layers respectively (from top to bottom). Figure \ref{fig:celebA128} shows synthesis results generated by the learned model. 

\begin{figure}
\centering
\subfigure[training images]{
\includegraphics[width=.95\linewidth]{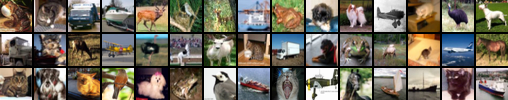}}
\subfigure[synthesized images]{
\includegraphics[width=.95\linewidth]{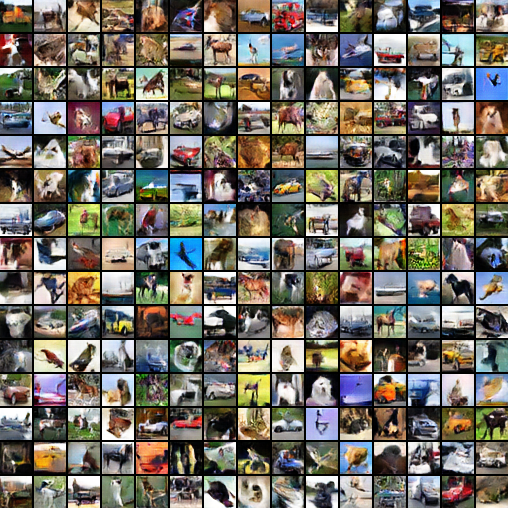}}
\caption{Generating object images (32 $\times$ 32 pixels). The synthesized images are generated by the CoopNets algorithm that learns from Cifar-10 dataset with 60,000 training images.}
\label{fig:cifar32}
\end{figure}

Figure \ref{fig:cifar32} displays synthesis results generated by the model learned from Cifar-10 objects dataset with 60k training images of 32 $\times$ 32 pixels. The descriptor network has 3 convolutional layers with numbers of channels $\{64, 128, 256\}$, filter sizes $\{5, 3, 3\}$ and sub-sampling factors  $\{2, 2, 2\}$, and one fully connected layer of 100 filters (from bottom to top). Starting from a 100-dimensional latent vector, the generator network has 1 fully connected and 3 deconvolutional layers with numbers of channels $\{256, 128, 64, 3\}$, kernels sizes $\{4, 5, 5, 5\}$ and up-sampling factors $\{1, 2, 2, 2\}$ (from top to bottom). The number of Langevin dynamics steps in each learning iteration is 10. The size of mini-batch is 300. The number of parallel chains is $\tilde{n}=400$. We disable the noise term in the Langevin revision dynamics in the second half of the learning algorithm for faster convergence. 

We evaluate the synthesis results by the Fr\'echet Inception Distance (FID) \cite{heusel2017gans}, which measures the dissimilarity between generated images and real ones. Table \ref{tab:FID} shows the performance of CoopNets, DCGAN, W-GAN, and VAE on LSUN bedrooms, CelebA and Cifar-10 datasets with respect to FID. The experiments show that the learned generator is good enough as a standalone model, though the learned descriptor produces even better results with Langevin revision. Both of them outperform the other baseline methods in terms of FID on these three benchmark datasets.

\begin{table}
\centering
\caption{The performance of CoopNets, DCGAN, W-GAN, and VAE on LSUN bedrooms, CelebA and Cifar-10 datasets with respect to the Fr\'echet Inception Distance (FID). }
\label{tab:FID}
\begin{tabular}{|c|r|r|r|}
\hline
 & LSUN & CelebA & Cifar-10 \\
\hline \hline
W-GAN \cite{arjovsky2017wasserstein}  & 67.72 & 52.54 & 48.40\\
DCGAN \cite{radford2015unsupervised}  & 70.40 & 21.40 & 37.70\\
VAE \cite{KingmaCoRR13}  & 243.47 & 50.53& 126.32\\ \hline
Generator in CoopNets (ours) & 64.30  & 16.98 & 35.25\\
Descriptor in CoopNets (ours) & \textbf{35.42}  & \textbf{16.65} & \textbf{33.61}\\
\hline
\end{tabular}
\end{table}

We further conduct a human perceptual study to compare the perceived realism of the synthesized examples generated by different generative models. More specifically, we randomly select 80 human subjects and ask them to rank the models, which includes CoopNets, DCGAN, W-GAN and VAE, according to the quality of their synthesized images. For each human subject, we present 16 synthesized images per model without telling the subject which model generates the images. The model order is randomized to ensure fair comparisons. Table \ref{tab:human_study} summarizes the average ranking results over human subjects on LSUN bedrooms, CelebA and Cifar-10 datasets. It can be seen that our model can generate more realistic images than the other baseline methods in this experiment.

\begin{table}
\centering
\caption{Human perceptual study for comparing synthesis qualities of different generative
models. The numbers are the average rankings.}
\label{tab:human_study}
\begin{tabular}{|c|r|r|r|}
\hline
 & LSUN & CelebA & Cifar-10 \\
\hline \hline
W-GAN \cite{arjovsky2017wasserstein}  & 2.400 & 3.800 & 2.171\\
DCGAN \cite{radford2015unsupervised}  & 1.913 & 2.463 & 2.395\\
VAE \cite{KingmaCoRR13}  & 3.988 & 2.300& 3.987\\
CoopNets (ours) & \textbf{1.700}  & \textbf{1.438} & \textbf{1.447}\\
\hline
\end{tabular}
\end{table}

\subsection{Experiment 6: Pattern completion} 

We conduct an experiment on learning from training images of human faces, and then testing the learned models on completing the occluded testing images. This can be considered an inpainting task. It can also be viewed as a form of content addressable memory or associative memory. The purpose of this experiment is to check whether the learned generator model can generalize to the testing data, i.e., whether the learned model overfits or underfits the training data. Underfitting can happen due to mode collapsing. We believe this is a powerful test of the generalizability of the learned generator model. 

 We adopt a 4-layer descriptor network, where the numbers of filters from bottom layer to top layer are 96, 128, 256, and 50. The filter size of each layer is $5\times5$, and the sub-sampling rate is 2. The final layer is a fully connected layer. The structure of the generator network is the same as in Section \ref{subsec:Exp2}. The training data are $10,000$ human faces randomly selected from CelebA dataset, and resized to 64 $\times$ 64 pixels.

\begin{figure}[h]
\centering
\includegraphics[width=.1\linewidth]{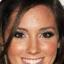}
\includegraphics[width=.1\linewidth]{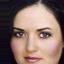}
\includegraphics[width=.1\linewidth]{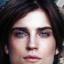}
\includegraphics[width=.1\linewidth]{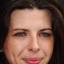}
\includegraphics[width=.1\linewidth]{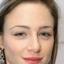}
\includegraphics[width=.1\linewidth]{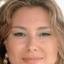}
\includegraphics[width=.1\linewidth]{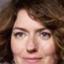}
\includegraphics[width=.1\linewidth]{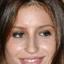}
\includegraphics[width=.1\linewidth]{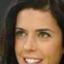} \\
\vspace{1mm}  
	\includegraphics[width=.1\linewidth]{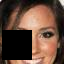}
	\includegraphics[width=.1\linewidth]{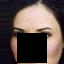}
	\includegraphics[width=.1\linewidth]{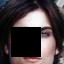}
	\includegraphics[width=.1\linewidth]{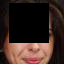}
	\includegraphics[width=.1\linewidth]{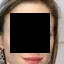}
	\includegraphics[width=.1\linewidth]{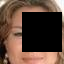}
	\includegraphics[width=.1\linewidth]{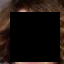}
	\includegraphics[width=.1\linewidth]{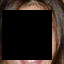}
	\includegraphics[width=.1\linewidth]{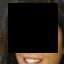} \\ \vspace{1mm}
	
		\includegraphics[width=.1\linewidth]{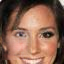}
	\includegraphics[width=.1\linewidth]{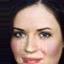}
	\includegraphics[width=.1\linewidth]{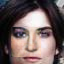}	
	\includegraphics[width=.1\linewidth]{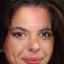}
	\includegraphics[width=.1\linewidth]{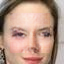}
	\includegraphics[width=.1\linewidth]{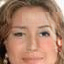}
	\includegraphics[width=.1\linewidth]{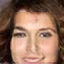}
	\includegraphics[width=.1\linewidth]{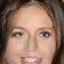}
	\includegraphics[width=.1\linewidth]{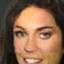} \\ 
(a) face \\ \vspace{1mm}

	\includegraphics[width=.1\linewidth]{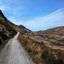}
	\includegraphics[width=.1\linewidth]{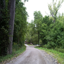}
	\includegraphics[width=.1\linewidth]{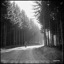}
	\includegraphics[width=.1\linewidth]{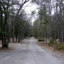}
	\includegraphics[width=.1\linewidth]{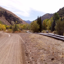}
		\includegraphics[width=.1\linewidth]{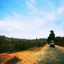}
	\includegraphics[width=.1\linewidth]{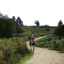}
	\includegraphics[width=.1\linewidth]{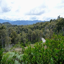}
	\includegraphics[width=.1\linewidth]{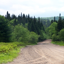}
\\ \vspace{1mm}
	
    \includegraphics[width=.1\linewidth]{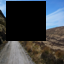}
    \includegraphics[width=.1\linewidth]{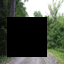}
	\includegraphics[width=.1\linewidth]{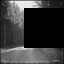}
	\includegraphics[width=.1\linewidth]{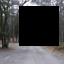}
	\includegraphics[width=.1\linewidth]{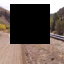}
		\includegraphics[width=.1\linewidth]{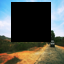}
	\includegraphics[width=.1\linewidth]{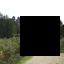}
	\includegraphics[width=.1\linewidth]{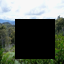}
	\includegraphics[width=.1\linewidth]{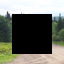}
\\	 \vspace{1mm}

	\includegraphics[width=.1\linewidth]{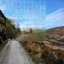}
	\includegraphics[width=.1\linewidth]{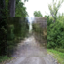}
	\includegraphics[width=.1\linewidth]{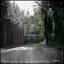}
	\includegraphics[width=.1\linewidth]{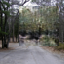}
	\includegraphics[width=.1\linewidth]{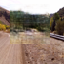}
		\includegraphics[width=.1\linewidth]{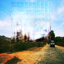}
	\includegraphics[width=.1\linewidth]{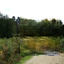}
	\includegraphics[width=.1\linewidth]{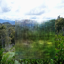}
	\includegraphics[width=.1\linewidth]{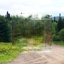}
\\ (b) forest road \\ \vspace{1mm}

	\includegraphics[width=.1\linewidth]{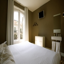}
	\includegraphics[width=.1\linewidth]{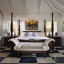}
	\includegraphics[width=.1\linewidth]{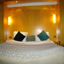}	
	\includegraphics[width=.1\linewidth]{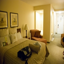}
	\includegraphics[width=.1\linewidth]{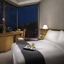}
		\includegraphics[width=.1\linewidth]{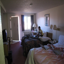}
	\includegraphics[width=.1\linewidth]{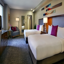}
	\includegraphics[width=.1\linewidth]{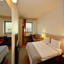}		
	\includegraphics[width=.1\linewidth]{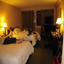}
\\ \vspace{1mm}

	\includegraphics[width=.1\linewidth]{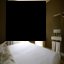}
	\includegraphics[width=.1\linewidth]{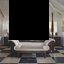}
	\includegraphics[width=.1\linewidth]{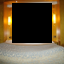}	
	\includegraphics[width=.1\linewidth]{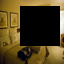}
	\includegraphics[width=.1\linewidth]{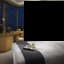}
		\includegraphics[width=.1\linewidth]{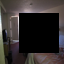}
	\includegraphics[width=.1\linewidth]{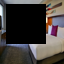}
	\includegraphics[width=.1\linewidth]{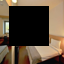}		
	\includegraphics[width=.1\linewidth]{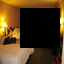}
\\ \vspace{1mm}

	\includegraphics[width=.1\linewidth]{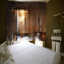}
	\includegraphics[width=.1\linewidth]{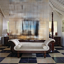}
	\includegraphics[width=.1\linewidth]{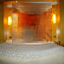}	
	\includegraphics[width=.1\linewidth]{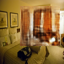}
	\includegraphics[width=.1\linewidth]{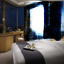}
		\includegraphics[width=.1\linewidth]{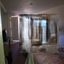}
	\includegraphics[width=.1\linewidth]{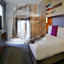}
	\includegraphics[width=.1\linewidth]{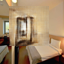}		
	\includegraphics[width=.1\linewidth]{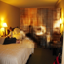} \\
	 (c) hotel room \\ 
		\caption{Pattern completion. First row: original images. Second row: occluded images. Third row: recovered images by the generator network learned via the CoopNets algorithm. (a) face. (b) forest road. (c) hotel room.}
	\label{fig:recovery}
\end{figure}

To quantitatively test whether we have learned a good generator network $g(X; \alpha)$ even though it has never seen the training images directly in the training stage, we apply it to the task of recovering the occluded pixels of testing images. For each occluded testing image $Y$, we use Step G1 of Algorithm G to infer the latent factors $X$. The only change is with respect to the term $\|Y - g(X; \alpha)\|^2$, where the sum of squares is over all the observed pixels of $Y$ in back-propagation computation. We run 1,000 Langevin steps, initializing $X$ from ${\rm N}(0, I_d)$. After inferring $X$, the completed image $g(X; \alpha)$ is automatically obtained.  We design 3 experiments, where we randomly place a $30 \times 30$, $40 \times 40$, or $50 \times 50$ mask on each $64\times64$ testing image. These 3 experiments are denoted by M30, M40, and M50 respectively (M for mask).

We report the recovery errors and compare our method with 7 different image inpainting methods as well as the DCGAN of \cite{radford2015unsupervised}. For DCGAN, we use the parameter setting in~\cite{radford2015unsupervised} and the number of learning iterations is 600. We use the same $10,000$ training images to learn DCGAN. After the model is learned, we keep the generator and use the same method as ours to infer the latent factors $X$, and recover the unobserved pixels. In the 7 inpainting methods, Methods MRF-$\ell_1$ and MRF-$\ell_2$ are based on Markov random field priors where the nearest neighbor potential terms are $\ell_1$ and $\ell_2$ differences respectively. Methods inter-1 to 5 are interpolation methods. Please refer to~\cite{inpainting_online} for details. 

Table \ref{table::recovery} displays the recovery errors of the 3 experiments, where the error is measured by per pixel difference (relative to the range of pixel values)  between the original image and the recovered image on the occluded region, averaged over 1,000 testing images. We also measure the error by the peak signal-to-noise ratio (PSNR). Figure \ref{fig:recovery} (a) displays some recovery results by our method. The first row shows the original images as the ground truth. The second row displays the testing images with occluded pixels. The third row displays the recovered images by the learned generator network. 

\begin{table*}[]
\centering
\caption{A comparison of recovery performances of different methods in 3 experiments}
\label{table::recovery}
\begin{tabular}{|c|c|cc|cc|ccccc|}
\hline
 & task & CoopNets & DCGAN & MRF-$\ell_1$ & MRF-$\ell_2$ & inter-1 & inter-2 & inter-3 & inter-4 & inter-5 \\ \hline \hline

 & M30 & ${\bf 0.115}$ &  0.211  & 0.132 & 0.134 & 0.120 & 0.120 & 0.265 & 0.120 & 0.120 \\ 
error & M40 & ${\bf 0.124}$ &  0.212  & 0.148 & 0.149 & 0.135 & 0.135 & 0.314 & 0.135 & 0.135 \\ 
 & M50 & ${\bf 0.136}$ &  0.214  & 0.178 & 0.179 & 0.170 & 0.166 & 0.353 & 0.164 & 0.164 \\  \hline %\hline
 & M30 & ${\bf 16.893}$ &  12.116  & 15.739 & 15.692 & 16.203 & 16.635 & 9.524 & 16.665 & 16.648 \\ 
PSNR & M40 & ${\bf 16.098}$ &11.984  & 14.834 & 14.785 & 15.065 & 15.644 & 8.178 & 15.698 & 15.688 \\ 
 & M50 & ${\bf 15.105}$ &  11.890  & 13.313 & 13.309 & 13.220 & 14.009 & 7.327 & 14.164 & 14.161 \\ \hline 
\end{tabular}
\end{table*}

We also apply the same approach to the forest road category and the hotel room category in MIT place205 dataset. See Figure \ref{fig:recovery} (b) and (c) for some qualitative results. The learned generator is quite imaginative in completing the occluded pixels.

\subsection{Experiment 7: Generating dynamic textures}

We can learn to generate video sequences by cooperative training of a spatial-temporal descriptor network \cite{xie2017synthesizing} and a spatial-temporal generator network. The spatial-temporal descriptor network consists of multiple layers of spatial-temporal filters that capture spatial-temporal features at various scales of the video sequences, while the spatial-temporal generator network maps the latent variables to the video sequences by multiple layers of spatial-temporal kernels (basis functions). We call the resulting model the spatial-temporal CoopNets (ST-CoopNets). 

We conduct experiments on generating dynamic textures \cite{doretto2003dynamic} with temporal stationarity. We learn a spatial-temporal CoopNets for each category from one training video. We collect the training video clips from DynTex++ dataset of \cite{ghanem2010maximum} and the Internet. Each training video clip is of size 128 pixels $\times$ 128 pixels $\times$ 64 frames.

\begin{figure*}
\begin{center}
%\rotatebox{90}{\hspace{4mm}{\footnotesize obs1}}
\includegraphics[height=.07\linewidth, width=.07\linewidth]{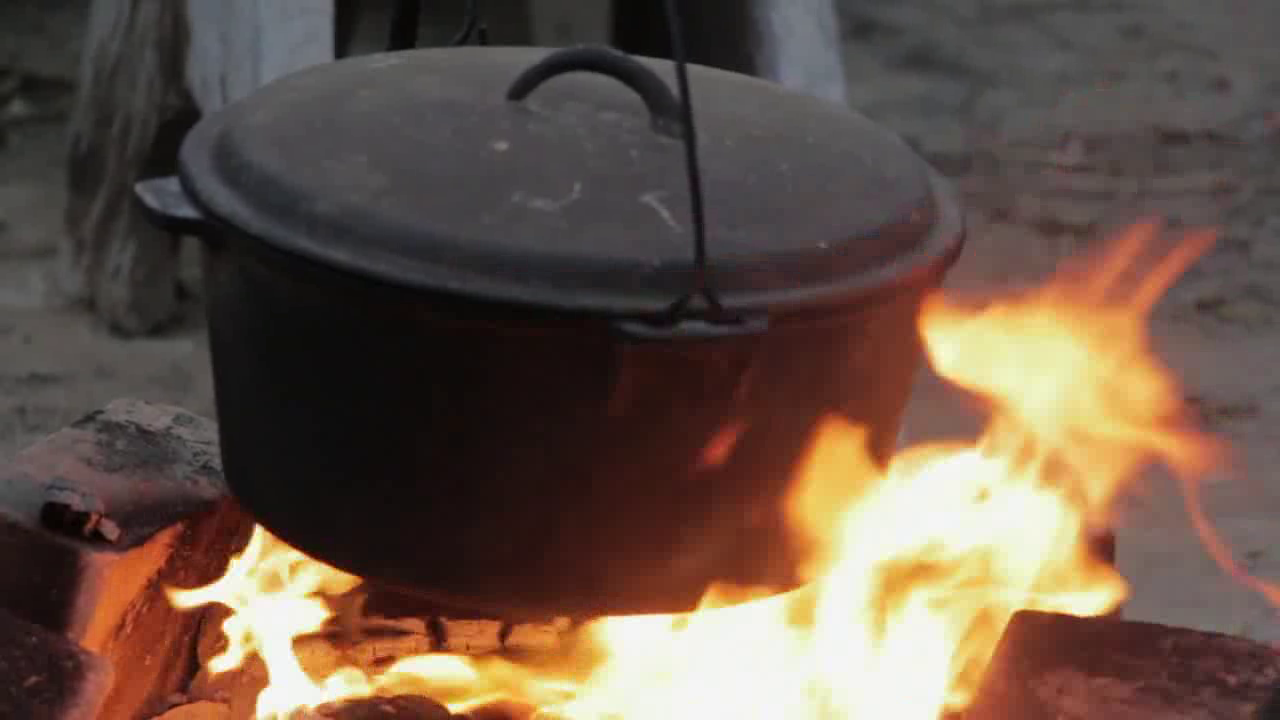}
\includegraphics[height=.07\linewidth, width=.07\linewidth]{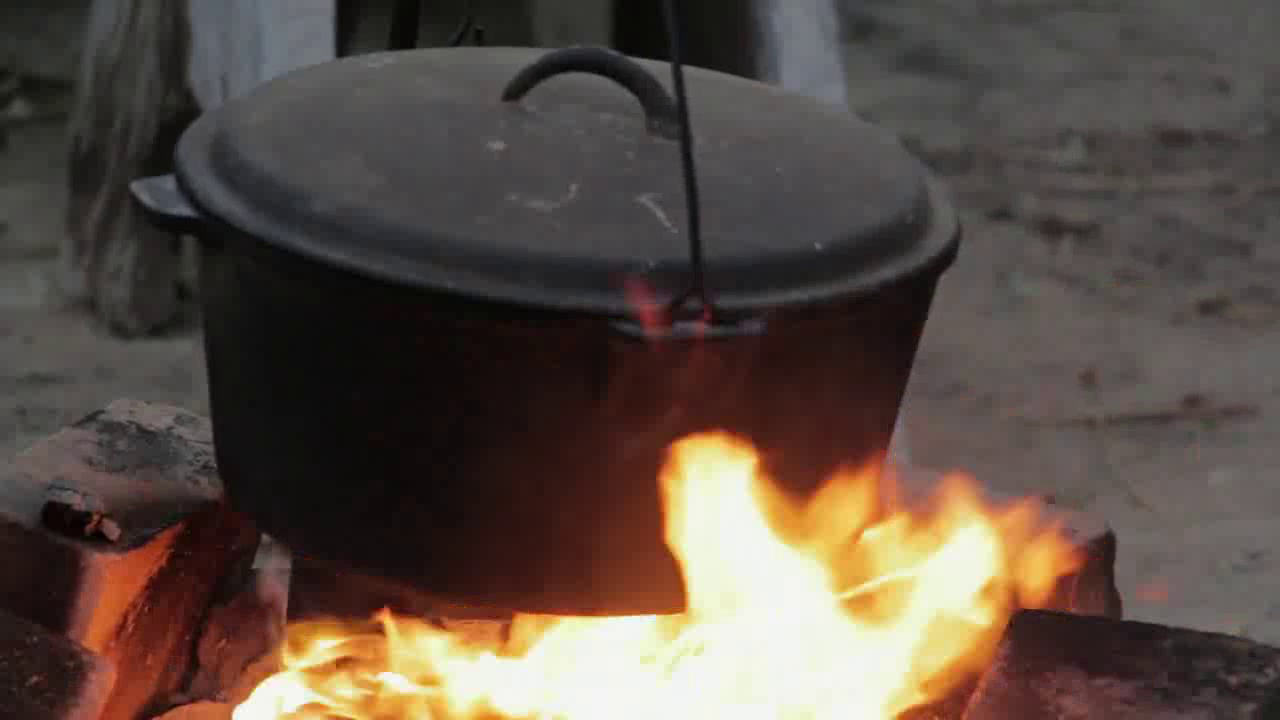}
\includegraphics[height=.07\linewidth, width=.07\linewidth]{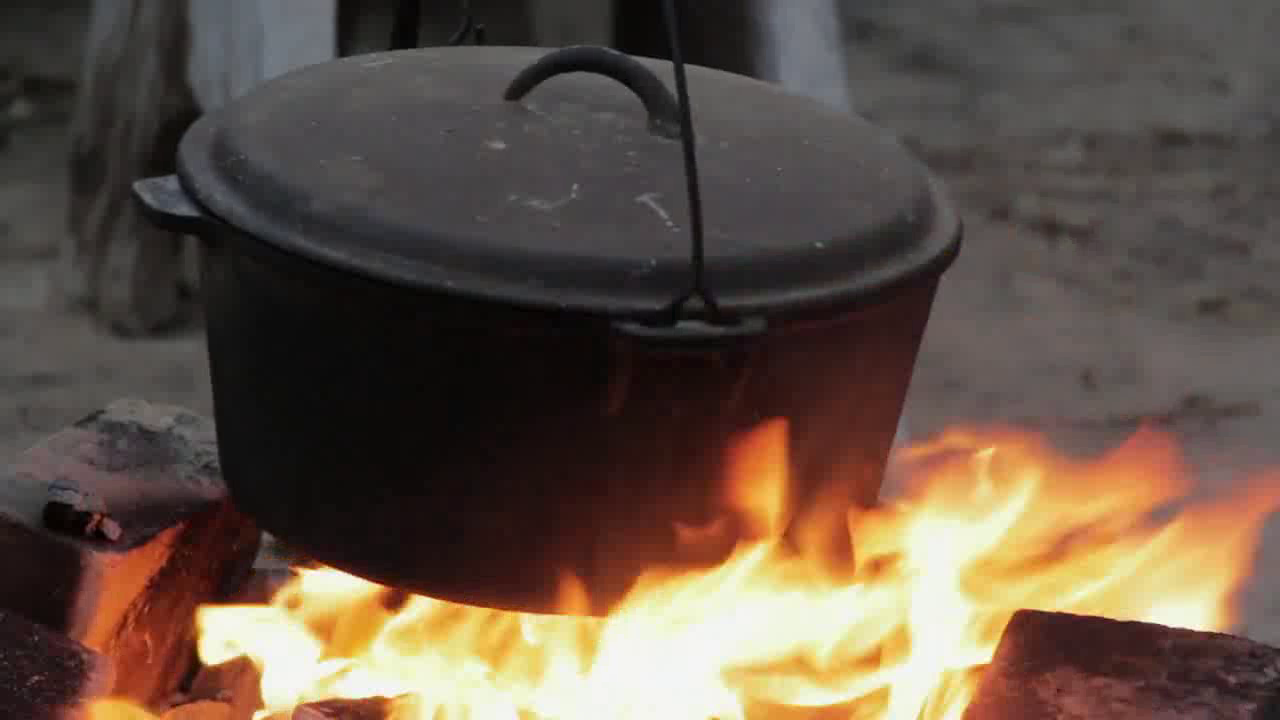}
\includegraphics[height=.07\linewidth, width=.07\linewidth]{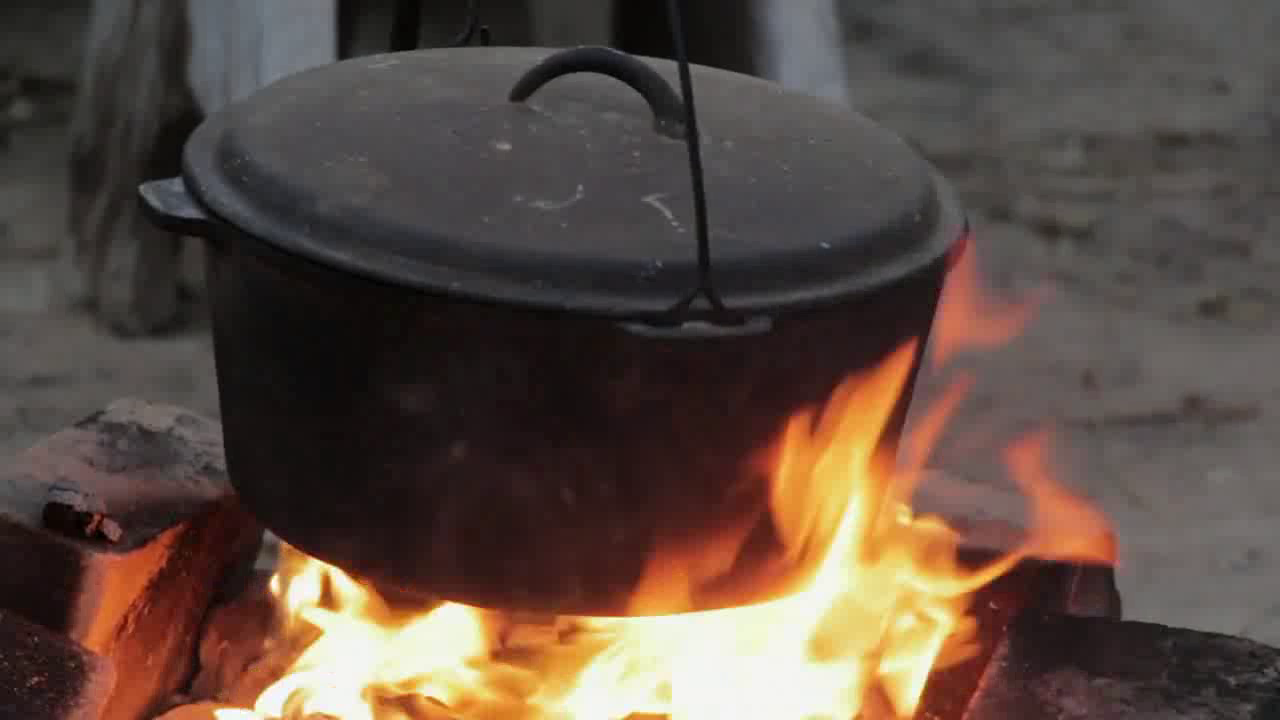}
\includegraphics[height=.07\linewidth, width=.07\linewidth]{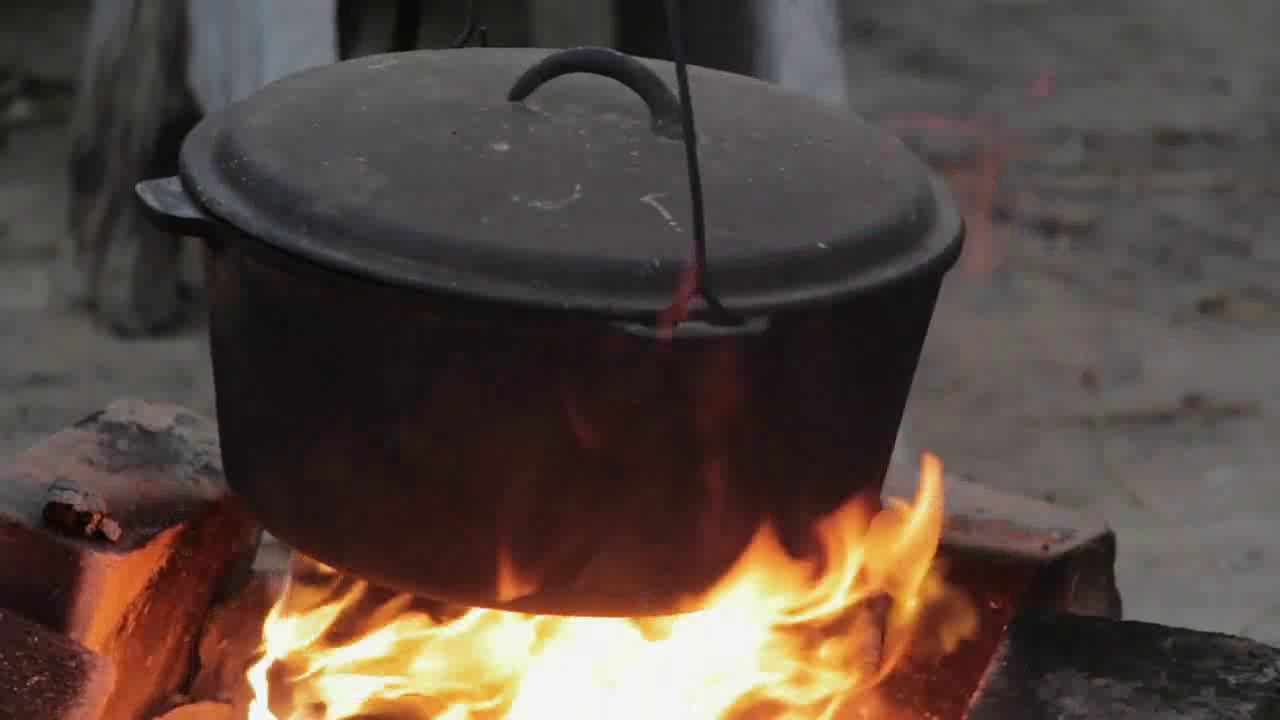}
\includegraphics[height=.07\linewidth, width=.07\linewidth]{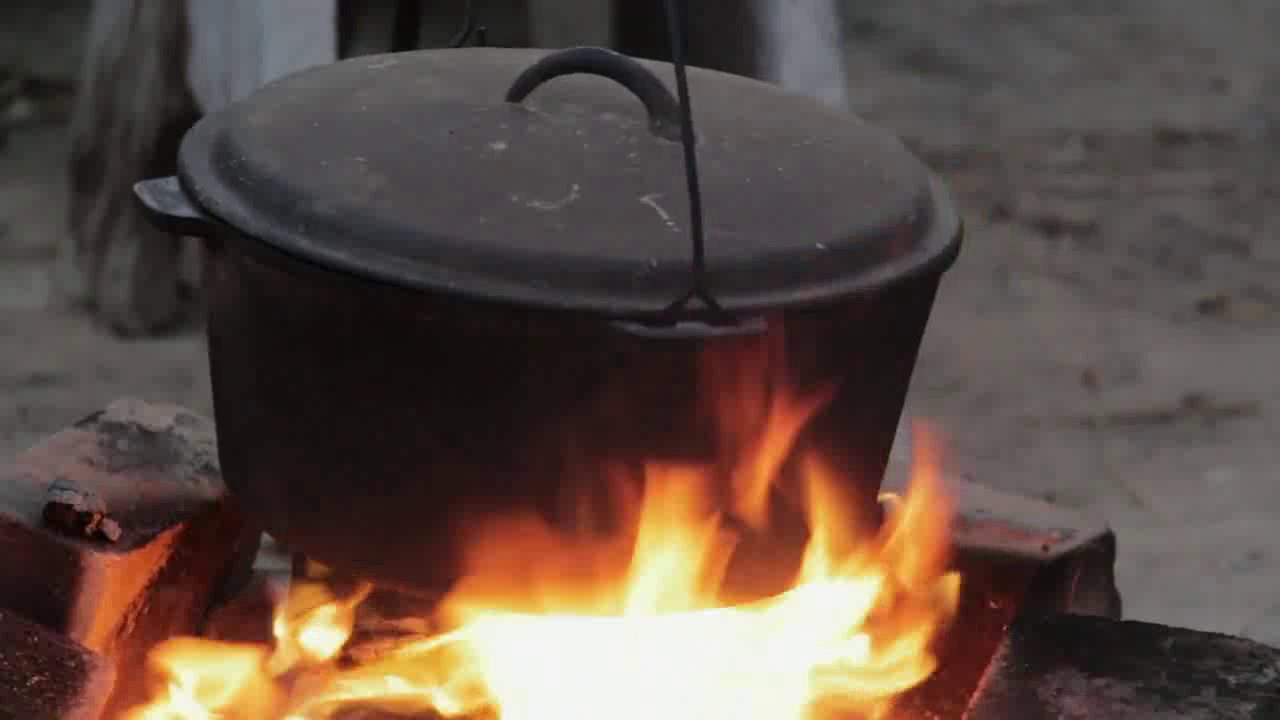}
\hspace{5mm}
%\rotatebox{90}{\hspace{4mm}{\footnotesize obs1}}
\includegraphics[height=.07\linewidth, width=.07\linewidth]{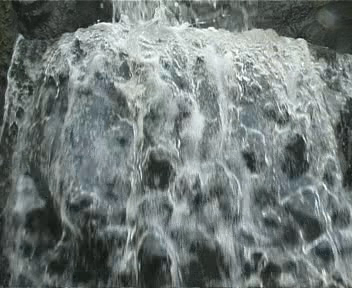}
\includegraphics[height=.07\linewidth, width=.07\linewidth]{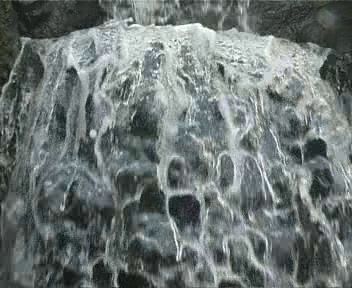}
\includegraphics[height=.07\linewidth, width=.07\linewidth]{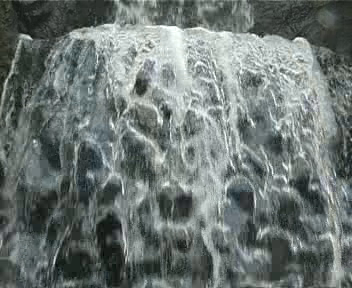}
\includegraphics[height=.07\linewidth, width=.07\linewidth]{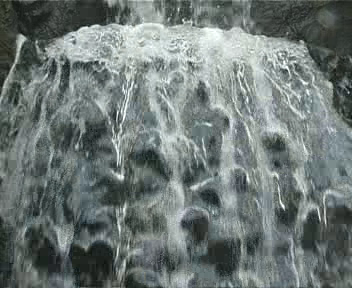}
\includegraphics[height=.07\linewidth, width=.07\linewidth]{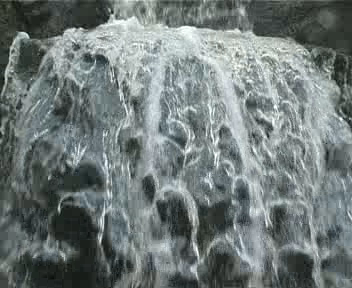} \includegraphics[height=.07\linewidth, width=.07\linewidth]{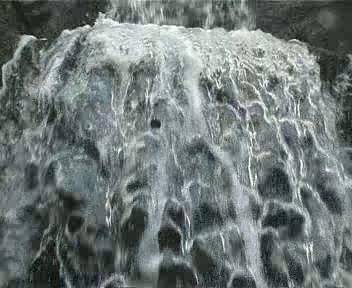}\\
\vspace{1mm}
%\rotatebox{90}{\hspace{4mm}{\footnotesize syn1}}
\includegraphics[height=.07\linewidth]{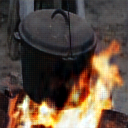}
\includegraphics[height=.07\linewidth]{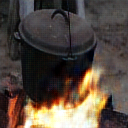}
\includegraphics[height=.07\linewidth]{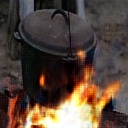}
\includegraphics[height=.07\linewidth]{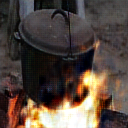}
\includegraphics[height=.07\linewidth]{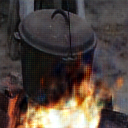}
\includegraphics[height=.07\linewidth]{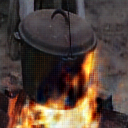}
\hspace{5mm}
%\rotatebox{90}{\hspace{4mm}{\footnotesize syn1}}
\includegraphics[height=.07\linewidth]{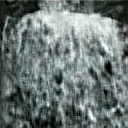}
\includegraphics[height=.07\linewidth]{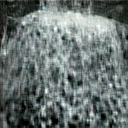}
\includegraphics[height=.07\linewidth]{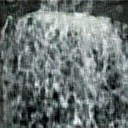}
\includegraphics[height=.07\linewidth]{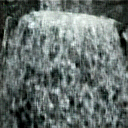}
\includegraphics[height=.07\linewidth]{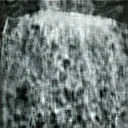}
\includegraphics[height=.07\linewidth]{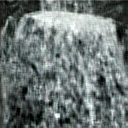}\\
\vspace{1mm}
%\rotatebox{90}{\hspace{4mm}{\footnotesize syn2}}
\includegraphics[height=.07\linewidth]{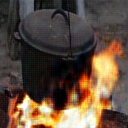}
\includegraphics[height=.07\linewidth]{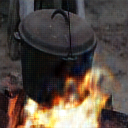}
\includegraphics[height=.07\linewidth]{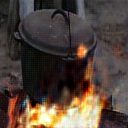}
\includegraphics[height=.07\linewidth]{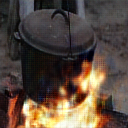}
\includegraphics[height=.07\linewidth]{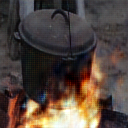}
\includegraphics[height=.07\linewidth]{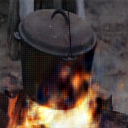}
\hspace{5mm}
%\rotatebox{90}{\hspace{4mm}{\footnotesize syn2}}
\includegraphics[height=.07\linewidth]{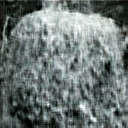}
\includegraphics[height=.07\linewidth]{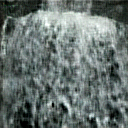}
\includegraphics[height=.07\linewidth]{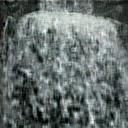}
\includegraphics[height=.07\linewidth]{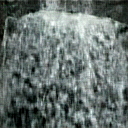}
\includegraphics[height=.07\linewidth]{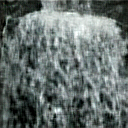}
\includegraphics[height=.07\linewidth]{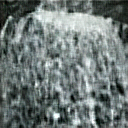}\\
\hspace{-15mm} (a) burning fire heating a pot  \hspace{55mm} (b) waterfall\\
\vspace{2mm}
%\rotatebox{90}{\hspace{4mm}{\footnotesize obs1}}
\includegraphics[height=.07\linewidth, width=.07\linewidth]{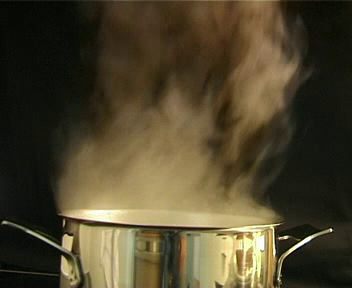}
\includegraphics[height=.07\linewidth, width=.07\linewidth]{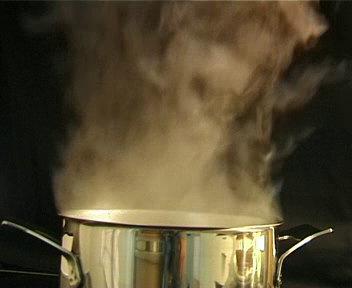}
\includegraphics[height=.07\linewidth, width=.07\linewidth]{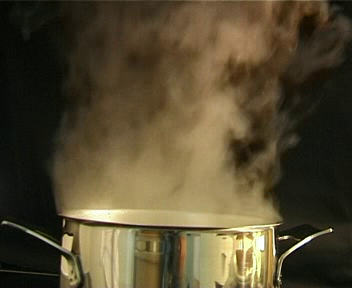}
\includegraphics[height=.07\linewidth, width=.07\linewidth]{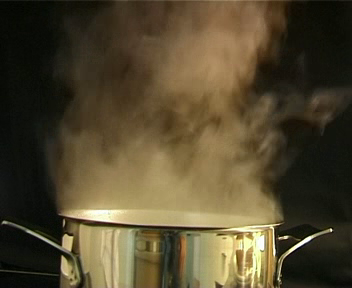}
\includegraphics[height=.07\linewidth, width=.07\linewidth]{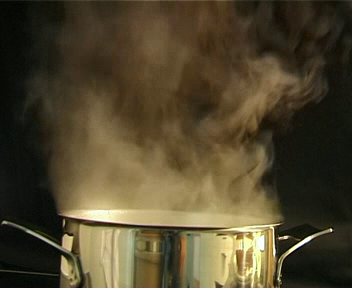}
\includegraphics[height=.07\linewidth, width=.07\linewidth]{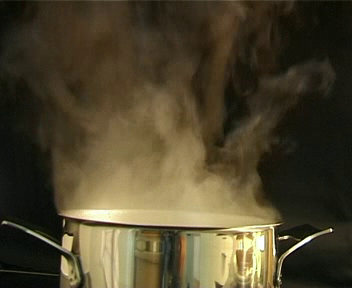}
\hspace{5mm}
%\rotatebox{90}{\hspace{4mm}{\footnotesize obs1}}
\includegraphics[height=.07\linewidth, width=.07\linewidth]{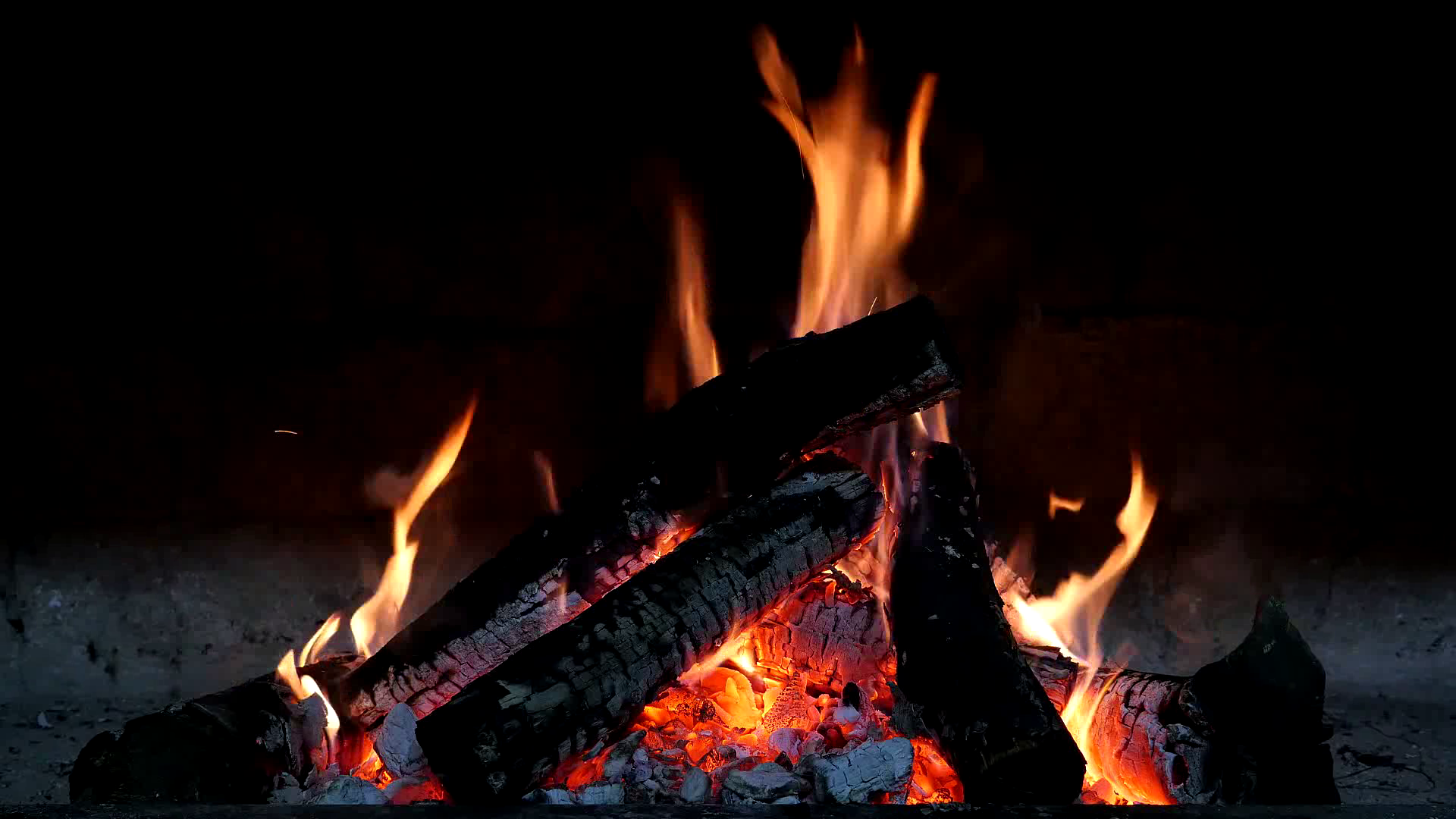}
\includegraphics[height=.07\linewidth, width=.07\linewidth]{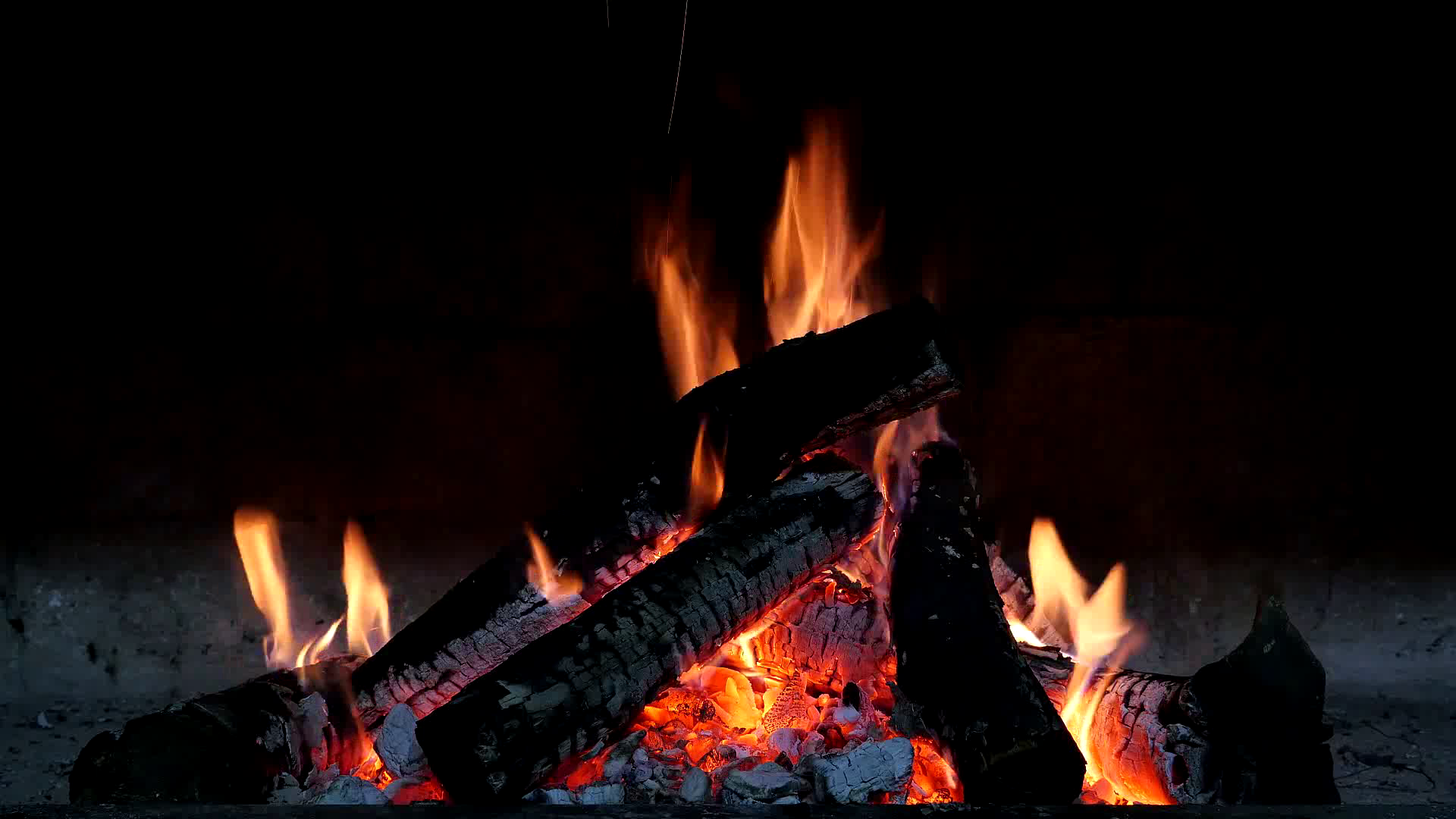}
\includegraphics[height=.07\linewidth, width=.07\linewidth]{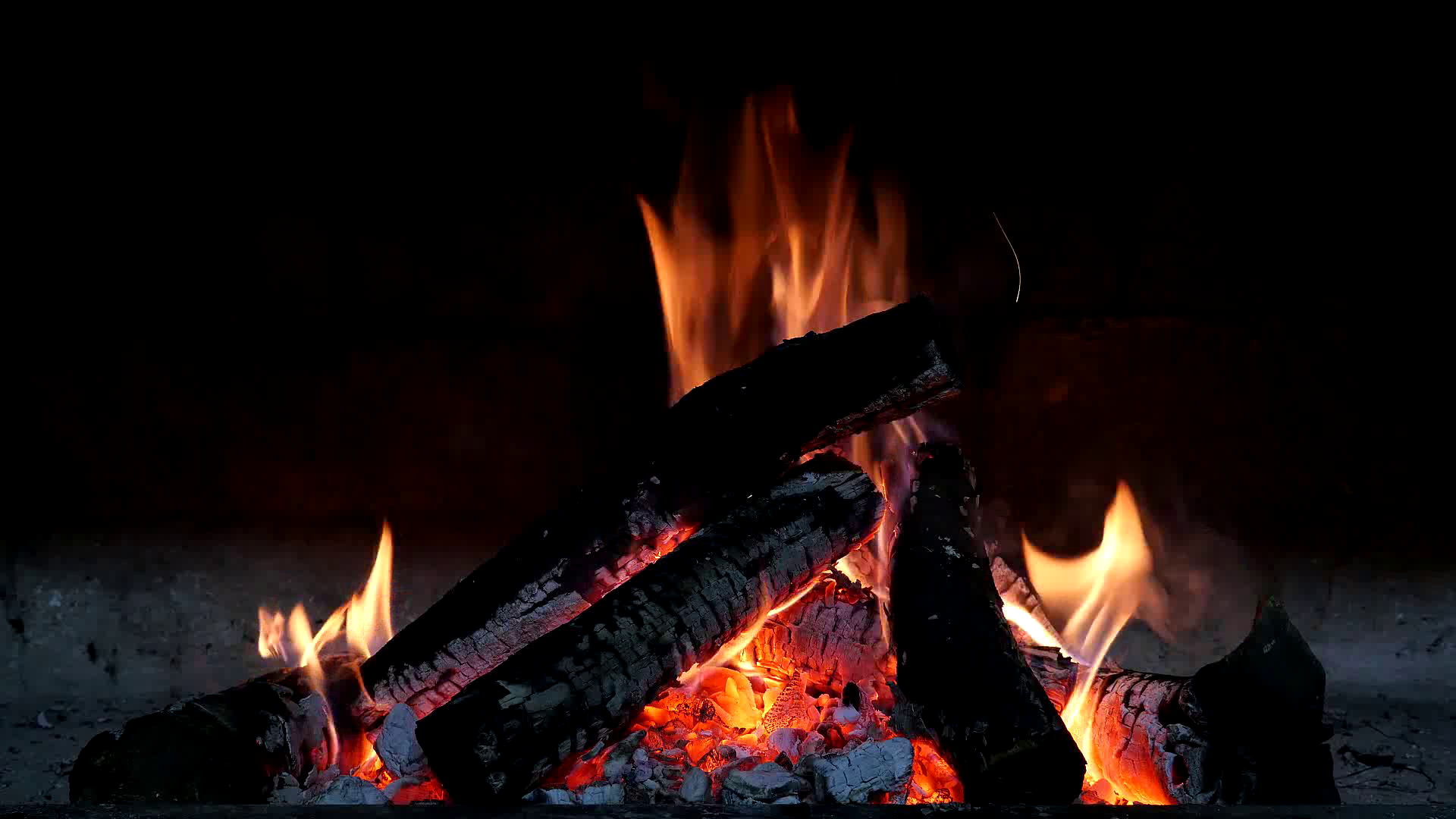}
\includegraphics[height=.07\linewidth, width=.07\linewidth]{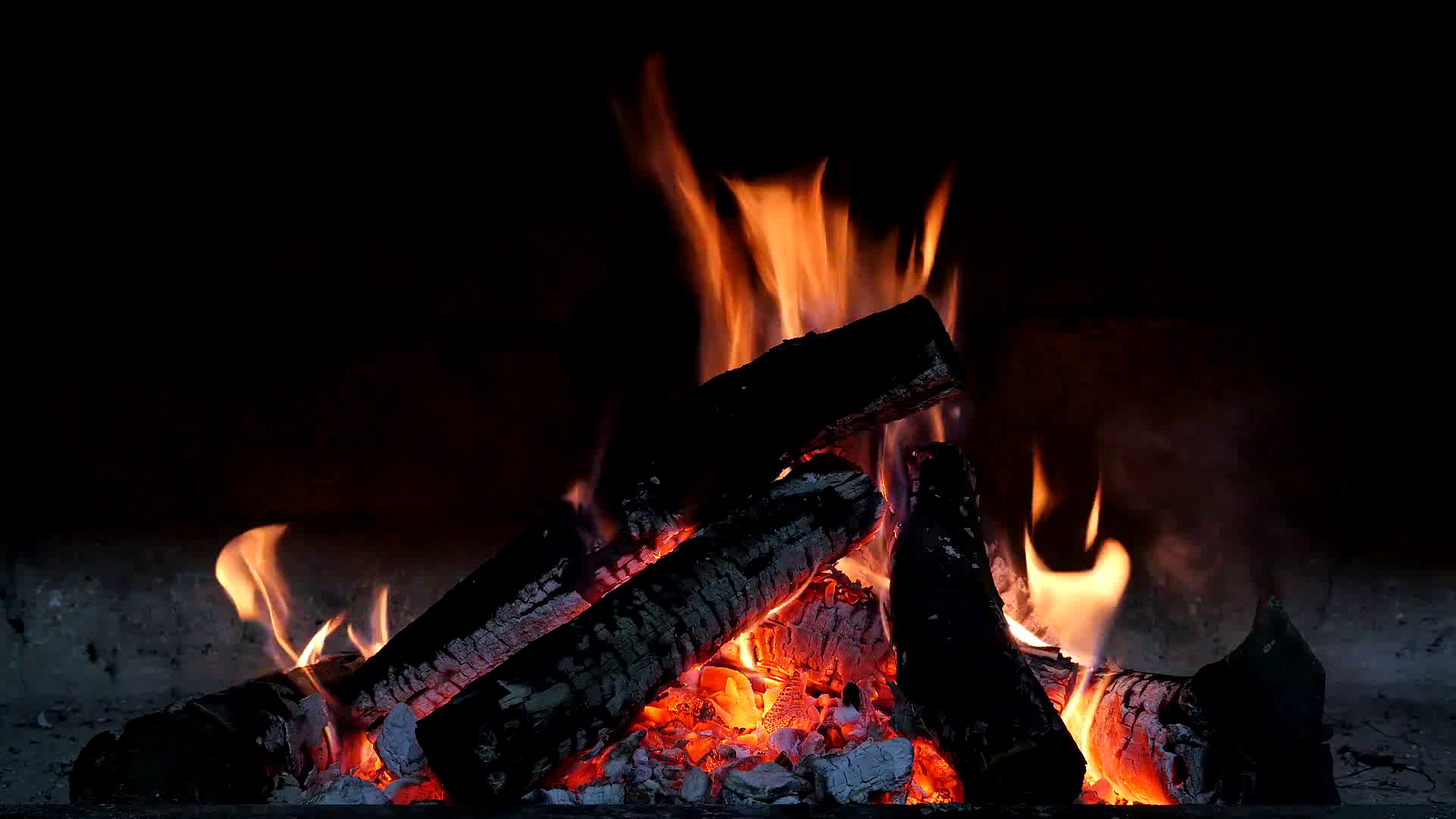}
\includegraphics[height=.07\linewidth, width=.07\linewidth]{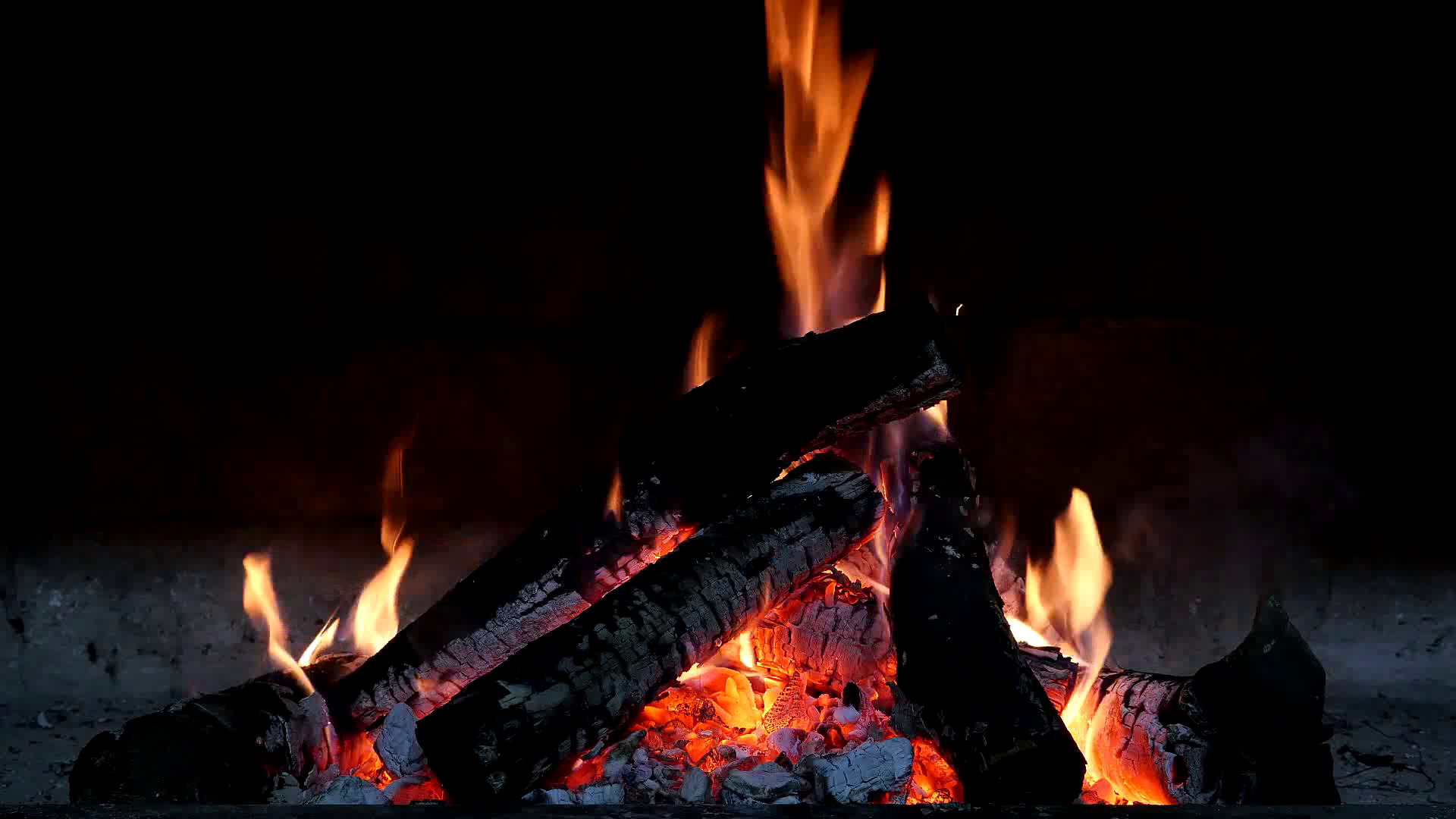} \includegraphics[height=.07\linewidth, width=.07\linewidth]{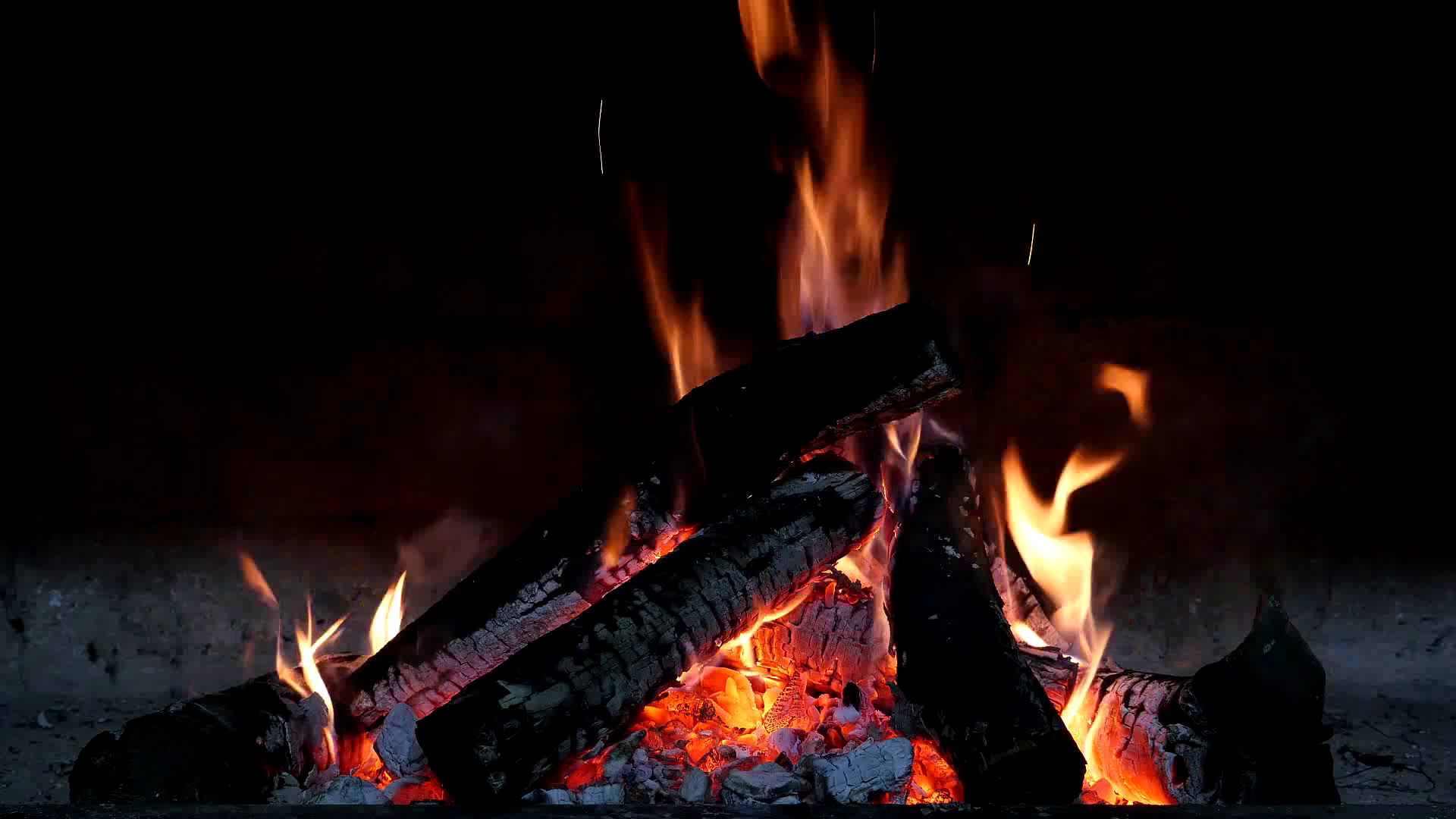}\\
\vspace{1mm}
%\rotatebox{90}{\hspace{4mm}{\footnotesize syn1}}
\includegraphics[height=.07\linewidth]{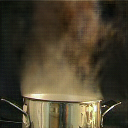}
\includegraphics[height=.07\linewidth]{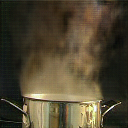}
\includegraphics[height=.07\linewidth]{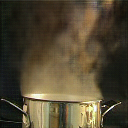}
\includegraphics[height=.07\linewidth]{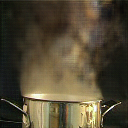}
\includegraphics[height=.07\linewidth]{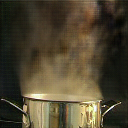}
\includegraphics[height=.07\linewidth]{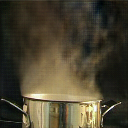}
\hspace{5mm}
%\rotatebox{90}{\hspace{4mm}{\footnotesize syn1}}
\includegraphics[height=.07\linewidth]{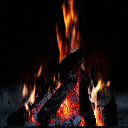}
\includegraphics[height=.07\linewidth]{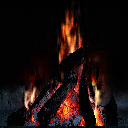}
\includegraphics[height=.07\linewidth]{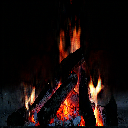}
\includegraphics[height=.07\linewidth]{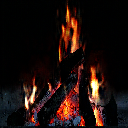}
\includegraphics[height=.07\linewidth]{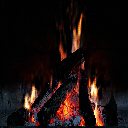}
\includegraphics[height=.07\linewidth]{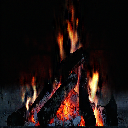}\\
\vspace{1mm}
%\rotatebox{90}{\hspace{4mm}{\footnotesize syn2}}
\includegraphics[height=.07\linewidth]{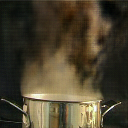}
\includegraphics[height=.07\linewidth]{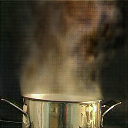}
\includegraphics[height=.07\linewidth]{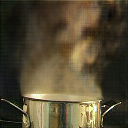}
\includegraphics[height=.07\linewidth]{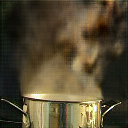}
\includegraphics[height=.07\linewidth]{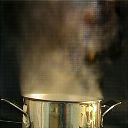}
\includegraphics[height=.07\linewidth]{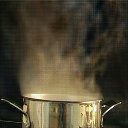}
\hspace{5mm}
%\rotatebox{90}{\hspace{4mm}{\footnotesize syn2}}
\includegraphics[height=.07\linewidth]{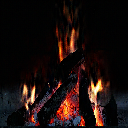}
\includegraphics[height=.07\linewidth]{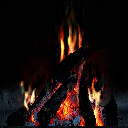}
\includegraphics[height=.07\linewidth]{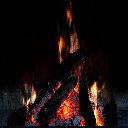}
\includegraphics[height=.07\linewidth]{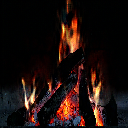}
\includegraphics[height=.07\linewidth]{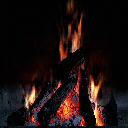}
\includegraphics[height=.07\linewidth]{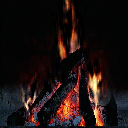}\\
(c) water vapor \hspace{60mm} (d) burning fireplace\\
\vspace{2mm} 
%\rotatebox{90}{\hspace{4mm}{\footnotesize obs1}}
\includegraphics[height=.07\linewidth, width=.07\linewidth]{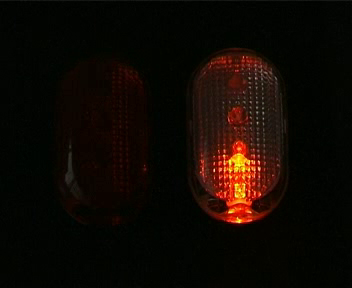}
\includegraphics[height=.07\linewidth, width=.07\linewidth]{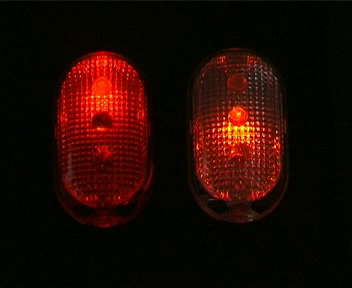}
\includegraphics[height=.07\linewidth, width=.07\linewidth]{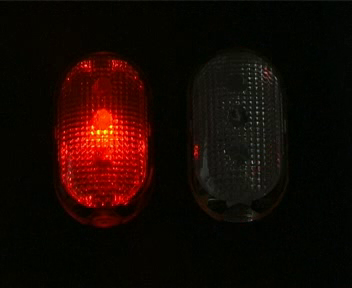}
\includegraphics[height=.07\linewidth, width=.07\linewidth]{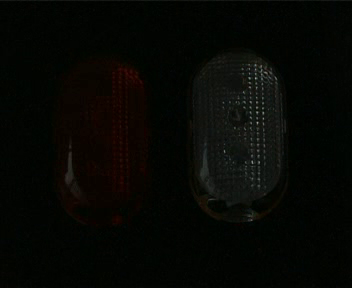}
\includegraphics[height=.07\linewidth, width=.07\linewidth]{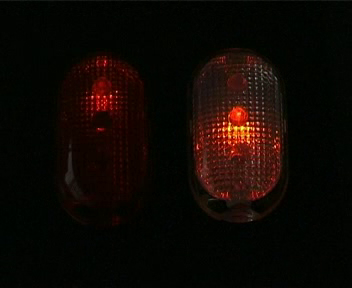}
\includegraphics[height=.07\linewidth, width=.07\linewidth]{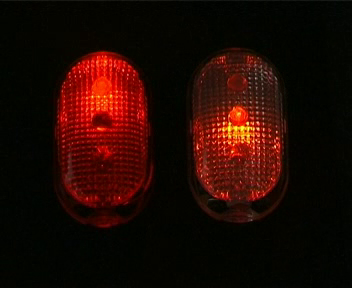} 
\hspace{5.1mm}
%\rotatebox{90}{\hspace{4mm}{\footnotesize obs1}} 
\includegraphics[height=.07\linewidth, width=.07\linewidth]{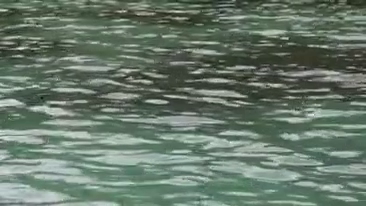}
\includegraphics[height=.07\linewidth, width=.07\linewidth]{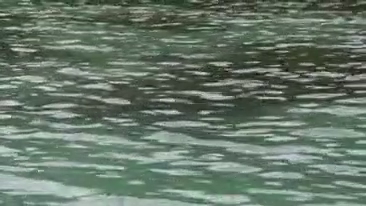}
\includegraphics[height=.07\linewidth, width=.07\linewidth]{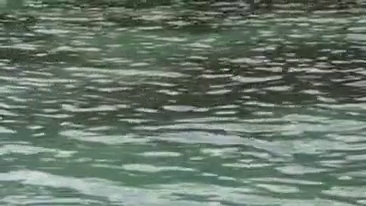}
\includegraphics[height=.07\linewidth, width=.07\linewidth]{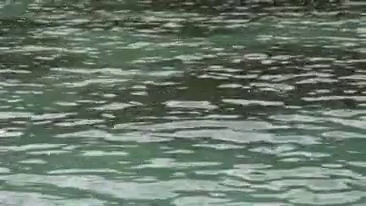}
\includegraphics[height=.07\linewidth, width=.07\linewidth]{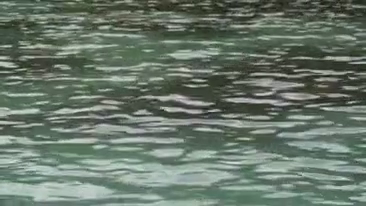}
\includegraphics[height=.07\linewidth, width=.07\linewidth]{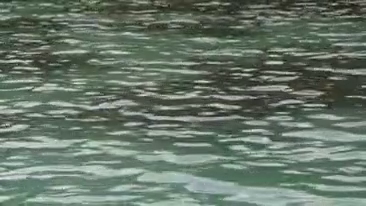}\\
\vspace{1mm}
%\rotatebox{90}{\hspace{4mm}{\footnotesize syn1}}
\includegraphics[height=.07\linewidth]{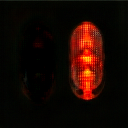}
\includegraphics[height=.07\linewidth]{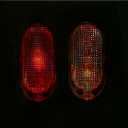}
\includegraphics[height=.07\linewidth]{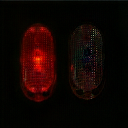}
\includegraphics[height=.07\linewidth]{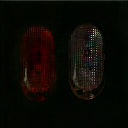}
\includegraphics[height=.07\linewidth]{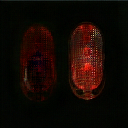}
\includegraphics[height=.07\linewidth]{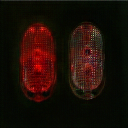} 
\hspace{5mm}
%\rotatebox{90}{\hspace{4mm}{\footnotesize syn1}}
\includegraphics[height=.07\linewidth]{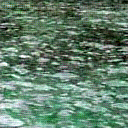}
\includegraphics[height=.07\linewidth]{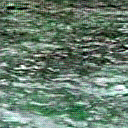}
\includegraphics[height=.07\linewidth]{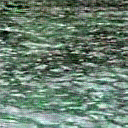}
\includegraphics[height=.07\linewidth]{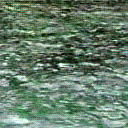}
\includegraphics[height=.07\linewidth]{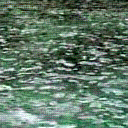}
\includegraphics[height=.07\linewidth]{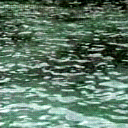}\\
\vspace{1mm}
%\rotatebox{90}{\hspace{4mm}{\footnotesize syn2}}
\includegraphics[height=.07\linewidth]{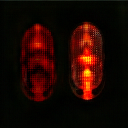}
\includegraphics[height=.07\linewidth]{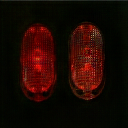}
\includegraphics[height=.07\linewidth]{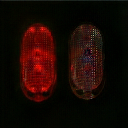}
\includegraphics[height=.07\linewidth]{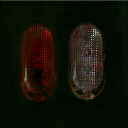}
\includegraphics[height=.07\linewidth]{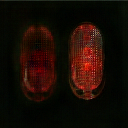}
\includegraphics[height=.07\linewidth]{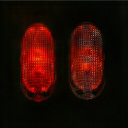}
\hspace{5mm}
%\rotatebox{90}{\hspace{4mm}{\footnotesize syn2}}
\includegraphics[height=.07\linewidth]{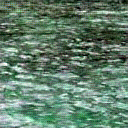}
\includegraphics[height=.07\linewidth]{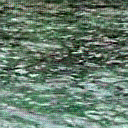}
\includegraphics[height=.07\linewidth]{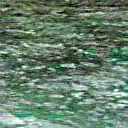}
\includegraphics[height=.07\linewidth]{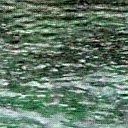}
\includegraphics[height=.07\linewidth]{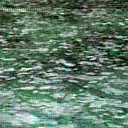}
\includegraphics[height=.07\linewidth]{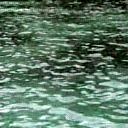}\\
\hspace{-10mm} (e) flashing lights  \hspace{65mm}(f) lake \\
\caption{Generating dynamic textures by the spatial-temporal CoopNets algorithm. For each category, the first row displays the frames of the
observed sequence, and the second and third rows display the corresponding frames of two synthesized sequences generated by the learning algorithm. The observed and synthesized videos are of size 128 pixels $\times$ 128 pixels $\times$ 64 frames. (a) burning fire heating a pot. (b) waterfall. (c) water vapor. (d) burning fireplace. (e) flashing lights. (f) lake.}
\label{fig:dt}
\end{center}
\end{figure*}

Since dynamic textures may have structured background that are not stationary in the spatial domain (e.g., burning fire heating a pot shown in Figure \ref{fig:dt} (a)), we adopt spatially fully connected and temporally convolutional layer in the bottom-up ConvNet structure of the descriptor. Specifically, we use a 3-layer descriptor network. The first layer has 120 5 pixels $\times$ 5 pixels $\times$ 5 frames filters with sub-sampling size of 2 pixels and frames. The second layer is a spatially fully connected layer, which contains 30 filters that are fully connected in the spatial domain but convolutional in the temporal domain. The temporal size of the filters is 5 with sub-sampling size of 2 and padding size of 2 in the temporal dimension. Due to the spatial full connectivity at the second layer, the spatial domain of the feature maps at the third layer is reduced to 1 $\times 1$. The third layer has 10 1 $\times$ 1 $\times$ 5 filters with sub-sampling size of 2 and padding size of 2 in the temporal dimension. We run $l_p=10$ Langevin revision steps in each iteration with the step size of 0.002. We set the standard deviation for reference distribution to $s=0.016$. The learning rate is 0.01.

The generator network maps a 5-channel $1 \times 1 \times 10$ latent factors to the video sequence by a 4-layer top-down ConvNet. Due to spatial non-stationarity and temporal stationarity of dynamics textures, we use spatially fully connected and temporally deconvolutional layer in the upper layers of the generator network. Specifically, the first layer uses $1 \times 1 \times 5$ kernels. The second layer is a spatially fully connected layer with kernels of size 5 in the temporal dimension. The third layer uses $5 \times 5 \times 5$ kernels with an up-sampling factor 2. The fourth layer uses $5 \times 5 \times 2$ kernels with an up-sampling size  $2 \times 2 \times 1$. The numbers of channels at different layers are 256, 128, 64, and 3 from top to bottom. Batch normalization and ReLU layers are used between consecutive layers and tanh is added at the bottom-layer. The standard deviation of the noise $\epsilon$ is set to  $\sigma=1$. The learning rate is 0.0002.

We use Adam optimizer to train the models. We run $\tilde{n}=2$ parallel chains. Figure \ref{fig:dt} displays the results. For each category, the first row shows 6 frames of the observed sequence, and the second and third rows show the corresponding frames of 2 synthesized sequences generated by the learning algorithm. We use the same set of parameters for all the categories without tuning. Our experiments show that the cooperative training of spatial-temporal descriptor and generator can synthesize realistic dynamic textures. 
 
To evaluate the quality of the synthesized examples, we compare our model with some baseline models for dynamic textures (e.g., LDS \cite{doretto2003dynamic}, FFT-LDS \cite{abraham2005dynamic}, MKGPDM \cite{zhu2016dynamic}, and HOSVD \cite{costantini2008higher}) in terms of PSNR and structural similarity measures (SSIM) on 6 dynamic texture videos. Each model learns from 64 image frames of each observed video, and then generates a 64-frame dynamic texture video for evaluation. The PSNR and SSIM are computed between the generated example and the observed example. Table \ref{tab:dt_comparison} shows the average performances of the models over the 6 videos. Our models are comparable or better than other baseline methods for dynamic textures.

\begin{table}[h]
\centering
\caption{A comparison of models for synthesizing dynamic textures}
\label{tab:dt_comparison}
\begin{tabular}{|c|c|c|}
\hline
   Model   & PSNR & SSIM \\
   \hline \hline
   LDS \cite{doretto2003dynamic}  & 19.148& 0.5939\\
   FFT-LDS \cite{abraham2005dynamic}  & 12.463 & 0.2898\\
   MKGPDM  \cite{zhu2016dynamic}  & 14.288& 0.3577\\   
   HOSVD  \cite{costantini2008higher}  & 18.392  & 0.4573\\   
   CoopNets (ours)  & \textbf{19.407}  & \textbf{0.5988}\\
   \hline
\end{tabular}
\end{table}

\section{Conclusion} 

This paper studies the fundamental problem of learning two important classes of deep generative models of images. By allowing the two models to cooperate with each other in a cooperative learning algorithm, our method can learn highly realistic generative models. We quantitatively evaluate our method on synthesis qualities of the generated images as well as the recovery errors of the reconstructed images. 

The most unique feature of our work is that the descriptor and the generator networks feed each other the synthesized data in the learning process. The generator feeds the descriptor the initial version of the synthesized data. The descriptor feedbacks the generator the revised version of the synthesized data. The generator then produces the reconstructed version of the synthesized data. While the descriptor learns from finite amount of observed data, the generator learns from virtually infinite amount of synthesized data. 

Another unique feature of our work is that the learning process interweaves the existing maximum likelihood learning algorithms for the two networks, so that  the two algorithms jumpstart each other's MCMC sampling. 

A third unique feature of our work is that the generator accumulates the MCMC transitions of the descriptor via MCMC teaching, and reproduces the MCMC transitions by direct ancestral sampling. In other words, the descriptor distills its MCMC algorithm into the generator. Powering the MCMC sampling of the descriptor model in \cite{LuZhuWu2016, XieLuICML} is a main motivation of this paper, with the bonus of turning the unsupervised learning of the generator    \cite{HanLu2016}  into supervised learning. 

Our cooperative training method can be generalized to a conditional version for learning the conditional distribution of a high-dimensional output given an input, which may also be high-dimensional. Examples include image generation given class label, text description, or abstract sketch, as well as recovering depth map from image. In the conditional cooperative learning method, the conditional descriptor learns the objective function or the conditional random field, and the conditional generator learns  to initialize the sampling or optimization algorithm for the objective function. 

Compared to reinforcement learning \cite{sutton1998introduction}, the descriptor network learns the cost function, the MCMC algorithm is like an optimal control algorithm, and the generator network is like a learned policy. Our method distills the optimal control algorithm into a policy. The optimal control is slow thinking while the policy is fast thinking. We shall investigate this connection in our future work. 

\section*{Project page} 

The code and results can be found at 
\url{http://www.stat.ucla.edu/~jxie/CoopNets/CoopNets.html}

\ifCLASSOPTIONcompsoc
  \section*{Acknowledgments}
\else
  \section*{Acknowledgment}
\fi

We thank Hansheng Jiang, Zilong Zheng, Erik Nijkamp, Tengyu Liu, Yaxuan Zhu, Zhaozhuo Xu and Xiaolin Fang for their assistance with coding and experiments. 

We gratefully acknowledge the support of NVIDIA Corporation with the donation of the Titan Xp GPU used for this research.

The work is supported by Hikvision gift fund, NSF DMS 1310391, DARPA SIMPLEX N66001-15-C-4035,  ONR MURI N00014-16-1-2007, and  DARPA ARO W911NF-16-1-0579.

\ifCLASSOPTIONcaptionsoff
  \newpage
\fi

\bibliographystyle{IEEEtran}
\bibliography{mybibfile}

\begin{thebibliography}{10}
\providecommand{\url}[1]{#1}
\csname url@samestyle\endcsname
\providecommand{\newblock}{\relax}
\providecommand{\bibinfo}[2]{#2}
\providecommand{\BIBentrySTDinterwordspacing}{\spaceskip=0pt\relax}
\providecommand{\BIBentryALTinterwordstretchfactor}{4}
\providecommand{\BIBentryALTinterwordspacing}{\spaceskip=\fontdimen2\font plus
\BIBentryALTinterwordstretchfactor\fontdimen3\font minus
  \fontdimen4\font\relax}
\providecommand{\BIBforeignlanguage}[2]{{%
\expandafter\ifx\csname l@#1\endcsname\relax
\typeout{** WARNING: IEEEtran.bst: No hyphenation pattern has been}%
\typeout{** loaded for the language `#1'. Using the pattern for}%
\typeout{** the default language instead.}%
\else
\language=\csname l@#1\endcsname
\fi
#2}}
\providecommand{\BIBdecl}{\relax}
\BIBdecl

\bibitem{lecun1998gradient}
Y.~LeCun, L.~Bottou, Y.~Bengio, and P.~Haffner, ``Gradient-based learning
  applied to document recognition,'' \emph{Proceedings of the IEEE}, vol.~86,
  no.~11, pp. 2278--2324, 1998.

\bibitem{krizhevsky2012imagenet}
A.~Krizhevsky, I.~Sutskever, and G.~E. Hinton, ``Imagenet classification with
  deep convolutional neural networks,'' in \emph{Advances in Neural Information
  Processing Systems}, 2012, pp. 1097--1105.

\bibitem{Lecun2006}
Y.~LeCun, S.~Chopra, R.~Hadsell, M.~Ranzato, and F.~J. Huang, ``A tutorial on
  energy-based learning,'' in \emph{Predicting Structured Data}.\hskip 1em plus
  0.5em minus 0.4em\relax MIT Press, 2006.

\bibitem{Hinton2002a}
G.~E. Hinton, ``Training products of experts by minimizing contrastive
  divergence.'' \emph{Neural Computation}, vol.~14, no.~8, pp. 1771--1800,
  2002.

\bibitem{zhu1997minimax}
S.-C. Zhu, Y.~N. Wu, and D.~Mumford, ``Minimax entropy principle and its
  application to texture modeling,'' \emph{Neural Computation}, vol.~9, no.~8,
  pp. 1627--1660, 1997.

\bibitem{roth2005fields}
S.~Roth and M.~J. Black, ``Fields of experts: A framework for learning image
  priors,'' in \emph{IEEE Conference on Computer Vision and Pattern
  Recognition}, vol.~2, 2005, pp. 860--867.

\bibitem{rubin1982algorithms}
D.~B. Rubin and D.~T. Thayer, ``{EM} algorithms for {ML} factor analysis,''
  \emph{Psychometrika}, vol.~47, no.~1, pp. 69--76, 1982.

\bibitem{zhu2003statistical}
S.-C. Zhu, ``Statistical modeling and conceptualization of visual patterns,''
  \emph{IEEE Transactions on Pattern Analysis and Machine Intelligence},
  vol.~25, no.~6, pp. 691--712, 2003.

\bibitem{guo2003modeling}
C.-E. Guo, S.-C. Zhu, and Y.~N. Wu, ``Modeling visual patterns by integrating
  descriptive and generative methods,'' \emph{International Journal of Computer
  Vision}, vol.~53, no.~1, pp. 5--29, 2003.

\bibitem{teh2003energy}
Y.-W. Teh, M.~Welling, S.~Osindero, and G.~E. Hinton, ``Energy-based models for
  sparse overcomplete representations,'' \emph{Journal of Machine Learning
  Research}, vol.~4, pp. 1235--1260, 2003.

\bibitem{wu2004information}
Y.~N. Wu, S.-C. Zhu, and C.-E. Guo, ``From information scaling of natural
  images to regimes of statistical models,'' \emph{Quarterly of Applied
  Mathematics}, vol.~66, pp. 81--122, 2008.

\bibitem{Ng2011}
J.~Ngiam, Z.~Chen, P.~W. Koh, and A.~Y. Ng, ``Learning deep energy models,'' in
  \emph{International Conference on Machine Learning}, 2011, pp. 1105--1112.

\bibitem{Dai2015ICLR}
J.~Dai, Y.~Lu, and Y.~N. Wu, ``Generative modeling of convolutional neural
  networks,'' in \emph{International Conference on Learning Representations},
  2015.

\bibitem{LuZhuWu2016}
Y.~Lu, S.-C. Zhu, and Y.~N. Wu, ``Learning {FRAME} models using {CNN}
  filters,'' in \emph{Thirtieth AAAI Conference on Artificial Intelligence},
  2016.

\bibitem{XieLuICML}
J.~Xie, Y.~Lu, S.-C. Zhu, and Y.~N. Wu, ``A theory of generative {ConvNet},''
  in \emph{International Conference on Machine Learning}, 2016.

\bibitem{Tu2017}
J.~Lazarow, L.~Jin, and Z.~Tu, ``Introspective generative modeling: Decide
  discriminatively,'' \emph{arXiv preprint arXiv:1704.07820}, 2017.

\bibitem{zeiler2013visualizing}
M.~D. Zeiler and R.~Fergus, ``Visualizing and understanding convolutional
  neural networks,'' in \emph{European Conference on Computer Vision}, 2014,
  pp. 818--833.

\bibitem{Alexey2015}
A.~Dosovitskiy, J.~Tobias~Springenberg, and T.~Brox, ``Learning to generate
  chairs with convolutional neural networks,'' in \emph{IEEE Conference on
  Computer Vision and Pattern Recognition}, 2015, pp. 1538--1546.

\bibitem{goodfellow2014generative}
I.~Goodfellow, J.~Pouget-Abadie, M.~Mirza, B.~Xu, D.~Warde-Farley, S.~Ozair,
  A.~Courville, and Y.~Bengio, ``Generative adversarial nets,'' in
  \emph{Advances in Neural Information Processing Systems}, 2014, pp.
  2672--2680.

\bibitem{denton2015deep}
E.~L. Denton, S.~Chintala, R.~Fergus \emph{et~al.}, ``Deep generative image
  models using a laplacian pyramid of adversarial networks,'' in \emph{Advances
  in Neural Information Processing Systems}, 2015, pp. 1486--1494.

\bibitem{radford2015unsupervised}
A.~Radford, L.~Metz, and S.~Chintala, ``Unsupervised representation learning
  with deep convolutional generative adversarial networks,'' \emph{arXiv
  preprint arXiv:1511.06434}, 2015.

\bibitem{neal2011mcmc}
R.~M. Neal, ``{MCMC} using hamiltonian dynamics,'' \emph{Handbook of Markov
  Chain Monte Carlo}, vol.~2, 2011.

\bibitem{girolami2011riemann}
M.~Girolami and B.~Calderhead, ``Riemann manifold langevin and hamiltonian
  monte carlo methods,'' \emph{Journal of the Royal Statistical Society: Series
  B (Statistical Methodology)}, vol.~73, no.~2, pp. 123--214, 2011.

\bibitem{zhu1998grade}
S.~C. Zhu and D.~Mumford, ``Grade: Gibbs reaction and diffusion equations,'' in
  \emph{International Conference on Computer Vision}, 1998, pp. 847--854.

\bibitem{HanLu2016}
T.~Han, Y.~Lu, S.-C. Zhu, and Y.~N. Wu, ``Alternating back-propagation for
  generator network,'' in \emph{31st AAAI Conference on Artificial
  Intelligence}, 2017.

\bibitem{dempster1977maximum}
A.~P. Dempster, N.~M. Laird, and D.~B. Rubin, ``Maximum likelihood from
  incomplete data via the em algorithm,'' \emph{Journal of the royal
  statistical society. Series B (methodological)}, pp. 1--38, 1977.

\bibitem{salakhutdinov2009deep}
R.~Salakhutdinov and G.~E. Hinton, ``Deep boltzmann machines,'' in
  \emph{AISTATS}, 2009.

\bibitem{Hinton06}
G.~E. Hinton, S.~Osindero, and Y.-W. Teh, ``A fast learning algorithm for deep
  belief nets,'' \emph{Neural Computation}, vol.~18, pp. 1527--1554, 2006.

\bibitem{Bengio2016}
T.~Kim and Y.~Bengio, ``Deep directed generative models with energy-based
  probability estimation,'' \emph{arXiv preprint arXiv:1606.03439}, 2016.

\bibitem{KingmaCoRR13}
D.~P. Kingma and M.~Welling, ``Auto-encoding variational bayes,'' in
  \emph{International Conference on Learning Representations}, 2014.

\bibitem{RezendeICML2014}
D.~J. Rezende, S.~Mohamed, and D.~Wierstra, ``Stochastic backpropagation and
  approximate inference in deep generative models,'' in \emph{International
  Conference on Machine Learning}, 2014, pp. 1278--1286.

\bibitem{MnihGregor2014}
A.~Mnih and K.~Gregor, ``Neural variational inference and learning in belief
  networks,'' in \emph{International Conference on Machine Learning}, 2014.

\bibitem{hinton2015distilling}
G.~Hinton, O.~Vinyals, and J.~Dean, ``Distilling the knowledge in a neural
  network,'' \emph{arXiv preprint arXiv:1503.02531}, 2015.

\bibitem{tu2007learning}
Z.~Tu, ``Learning generative models via discriminative approaches,'' in
  \emph{IEEE Conference on Computer Vision and Pattern Recognition}, 2007, pp.
  1--8.

\bibitem{lazarow2017introspective}
J.~Lazarow, L.~Jin, and Z.~Tu, ``Introspective neural networks for generative
  modeling,'' in \emph{IEEE Conference on Computer Vision and Pattern
  Recognition}, 2017, pp. 2774--2783.

\bibitem{jin2017introspective}
L.~Jin, J.~Lazarow, and Z.~Tu, ``Introspective classification with
  convolutional nets,'' in \emph{Advances in Neural Information Processing
  Systems}, 2017, pp. 823--833.

\bibitem{lee2017wasserstein}
K.~Lee, W.~Xu, F.~Fan, and Z.~Tu, ``Wasserstein introspective neural
  networks,'' \emph{arXiv preprint arXiv:1711.08875}, 2017.

\bibitem{oord2016conditional}
A.~van~den Oord, N.~Kalchbrenner, L.~Espeholt, O.~Vinyals, A.~Graves
  \emph{et~al.}, ``Conditional image generation with pixelcnn decoders,'' in
  \emph{Advances in Neural Information Processing Systems}, 2016, pp.
  4790--4798.

\bibitem{xie2018CoopNets}
J.~Xie, Y.~Lu, R.~Gao, and Y.~N. Wu, ``Cooperative learning of energy-based
  model and latent variable model via {MCMC} teaching,'' in \emph{Thirtieth
  AAAI Conference on Artificial Intelligence}, 2018.

\bibitem{xie2018learning}
J.~Xie, Z.~Zheng, R.~Gao, W.~Wang, S.-C. Zhu, and Y.~N. Wu, ``Learning
  descriptor networks for {3D} shape synthesis and analysis,'' in \emph{IEEE
  Conference on Computer Vision and Pattern Recognition}, 2018, pp. 8629--8638.

\bibitem{xie2017synthesizing}
J.~Xie, S.-C. Zhu, and Y.~N. Wu, ``Synthesizing dynamic patterns by
  spatial-temporal generative convnet,'' in \emph{IEEE Conference on Computer
  Vision and Pattern Recognition}, 2017, pp. 7093--7101.

\bibitem{robbins1951stochastic}
H.~Robbins and S.~Monro, ``A stochastic approximation method,'' \emph{The
  Annals of Mathematical Statistics}, pp. 400--407, 1951.

\bibitem{younes1999convergence}
L.~Younes, ``On the convergence of markovian stochastic algorithms with rapidly
  decreasing ergodicity rates,'' \emph{Stochastics: An International Journal of
  Probability and Stochastic Processes}, vol.~65, no. 3-4, pp. 177--228, 1999.

\bibitem{tieleman2008training}
T.~Tieleman, ``Training restricted boltzmann machines using approximations to
  the likelihood gradient,'' in \emph{International Conference on Machine
  Learning}, 2008, pp. 1064--1071.

\bibitem{ziebart2008maximum}
B.~D. Ziebart, A.~L. Maas, J.~A. Bagnell, and A.~K. Dey, ``Maximum entropy
  inverse reinforcement learning.'' in \emph{Twenty-Third AAAI Conference on
  Artificial Intelligence}, vol.~8, 2008, pp. 1433--1438.

\bibitem{abbeel2004apprenticeship}
P.~Abbeel and A.~Y. Ng, ``Apprenticeship learning via inverse reinforcement
  learning,'' in \emph{Proceedings of the twenty-first international conference
  on Machine learning}.\hskip 1em plus 0.5em minus 0.4em\relax ACM, 2004, p.~1.

\bibitem{hopfield1982neural}
J.~J. Hopfield, ``Neural networks and physical systems with emergent collective
  computational abilities,'' \emph{Proceedings of the National Academy of
  Sciences}, vol.~79, no.~8, pp. 2554--2558, 1982.

\bibitem{seung1998learning}
H.~S. Seung, ``Learning continuous attractors in recurrent networks,'' in
  \emph{Advances in Neural Information Processing Systems}, 1998, pp. 654--660.

\bibitem{grenander2007pattern}
U.~Grenander and M.~I. Miller, \emph{Pattern theory: from representation to
  inference}.\hskip 1em plus 0.5em minus 0.4em\relax Oxford University Press,
  2007.

\bibitem{cover2012elements}
T.~M. Cover and J.~A. Thomas, \emph{Elements of Information Theory}.\hskip 1em
  plus 0.5em minus 0.4em\relax John Wiley \& Sons, 2012.

\bibitem{ioffe2015batch}
S.~Ioffe and C.~Szegedy, ``Batch normalization: Accelerating deep network
  training by reducing internal covariate shift,'' \emph{arXiv preprint
  arXiv:1502.03167}, 2015.

\bibitem{deng2009imagenet}
J.~Deng, W.~Dong, R.~Socher, L.-J. Li, K.~Li, and L.~Fei-Fei, ``Imagenet: A
  large-scale hierarchical image database,'' in \emph{IEEE Conference on
  Computer Vision and Pattern Recognition}, 2009, pp. 248--255.

\bibitem{kingma2014adam}
D.~P. Kingma and J.~Ba, ``Adam: A method for stochastic optimization,''
  \emph{arXiv preprint arXiv:1412.6980}, 2014.

\bibitem{szegedy2016rethinking}
C.~Szegedy, V.~Vanhoucke, S.~Ioffe, J.~Shlens, and Z.~Wojna, ``Rethinking the
  inception architecture for computer vision,'' in \emph{IEEE Conference on
  Computer Vision and Pattern Recognition}, 2016, pp. 2818--2826.

\bibitem{wang2004image}
Z.~Wang, A.~C. Bovik, H.~R. Sheikh, and E.~P. Simoncelli, ``Image quality
  assessment: from error visibility to structural similarity,'' \emph{IEEE
  Transactions on Image Processing}, vol.~13, no.~4, pp. 600--612, 2004.

\bibitem{zhou2014learning}
B.~Zhou, A.~Lapedriza, J.~Xiao, A.~Torralba, and A.~Oliva, ``Learning deep
  features for scene recognition using places database,'' in \emph{Advances in
  Neural Information Processing Systems}, 2014, pp. 487--495.

\bibitem{salimans2016improved}
T.~Salimans, I.~Goodfellow, W.~Zaremba, V.~Cheung, A.~Radford, and X.~Chen,
  ``Improved techniques for training gans,'' in \emph{Advances in Neural
  Information Processing Systems}, 2016, pp. 2226--2234.

\bibitem{zhao2016energy}
J.~Zhao, M.~Mathieu, and Y.~LeCun, ``Energy-based generative adversarial
  network,'' \emph{International Conference for Learning Representations},
  2017.

\bibitem{arjovsky2017wasserstein}
M.~Arjovsky, S.~Chintala, and L.~Bottou, ``Wasserstein generative adversarial
  networks,'' \emph{International Conference on Machine Learning}, 2017.

\bibitem{chen2016infogan}
X.~Chen, Y.~Duan, R.~Houthooft, J.~Schulman, I.~Sutskever, and P.~Abbeel,
  ``Infogan: Interpretable representation learning by information maximizing
  generative adversarial nets,'' in \emph{Advances in Neural Information
  Processing Systems}, 2016, pp. 2172--2180.

\bibitem{breuleux2011quickly}
O.~Breuleux, Y.~Bengio, and P.~Vincent, ``Quickly generating representative
  samples from an rbm-derived process,'' \emph{Neural Computation}, vol.~23,
  no.~8, pp. 2058--2073, 2011.

\bibitem{bengio2014deep}
Y.~Bengio, E.~Laufer, G.~Alain, and J.~Yosinski, ``Deep generative stochastic
  networks trainable by backprop,'' in \emph{International Conference on
  Machine Learning}, 2014, pp. 226--234.

\bibitem{bengio2013representation}
Y.~Bengio, A.~Courville, and P.~Vincent, ``Representation learning: A review
  and new perspectives,'' \emph{IEEE Transactions on Pattern Analysis and
  Machine Intelligence}, vol.~35, no.~8, pp. 1798--1828, 2013.

\bibitem{yu15lsun}
F.~Yu, Y.~Zhang, S.~Song, A.~Seff, and J.~Xiao, ``{LSUN}: Construction of a
  large-scale image dataset using deep learning with humans in the loop,''
  \emph{arXiv preprint arXiv:1506.03365}, 2015.

\bibitem{liu2015deep}
Z.~Liu, P.~Luo, X.~Wang, and X.~Tang, ``Deep learning face attributes in the
  wild,'' in \emph{IEEE International Conference on Computer Vision}, 2015, pp.
  3730--3738.

\bibitem{krizhevsky2009learning}
A.~Krizhevsky, ``Learning multiple layers of features from tiny images,''
  University of Toronto, Tech. Rep., 2009.

\bibitem{heusel2017gans}
M.~Heusel, H.~Ramsauer, T.~Unterthiner, B.~Nessler, and S.~Hochreiter, ``Gans
  trained by a two time-scale update rule converge to a local nash
  equilibrium,'' in \emph{Advances in Neural Information Processing Systems},
  2017, pp. 6626--6637.

\bibitem{inpainting_online}
\BIBentryALTinterwordspacing
J.~D'Errico. (2004) Interpolation inpainting. [Online]. Available:
  \url{https://www.mathworks.com/matlabcentral/fileexchange/4551-inpaint-nans}
\BIBentrySTDinterwordspacing

\bibitem{doretto2003dynamic}
G.~Doretto, A.~Chiuso, Y.~N. Wu, and S.~Soatto, ``Dynamic textures,''
  \emph{International Journal of Computer Vision}, vol.~51, no.~2, pp. 91--109,
  2003.

\bibitem{ghanem2010maximum}
B.~Ghanem and N.~Ahuja, ``Maximum margin distance learning for dynamic texture
  recognition,'' in \emph{European Conference on Computer Vision}, 2010, pp.
  223--236.

\bibitem{abraham2005dynamic}
B.~Abraham, O.~I. Camps, and M.~Sznaier, ``Dynamic texture with fourier
  descriptors,'' in \emph{Proceedings of the 4th International Workshop on
  Texture Analysis and Synthesis}, 2005, pp. 53--58.

\bibitem{zhu2016dynamic}
Z.~Zhu, X.~You, S.~Yu, J.~Zou, and H.~Zhao, ``Dynamic texture modeling and
  synthesis using multi-kernel gaussian process dynamic model,'' \emph{Signal
  Processing}, vol. 124, pp. 63--71, 2016.

\bibitem{costantini2008higher}
R.~Costantini, L.~Sbaiz, and S.~Susstrunk, ``Higher order svd analysis for
  dynamic texture synthesis,'' \emph{IEEE Transactions on Image Processing},
  vol.~17, no.~1, pp. 42--52, 2008.

\bibitem{sutton1998introduction}
R.~S. Sutton and A.~G. Barto, \emph{Introduction to Reinforcement
  Learning}.\hskip 1em plus 0.5em minus 0.4em\relax MIT press Cambridge, 1998,
  vol. 135.

\end{thebibliography}

\end{document}